# Bootstrapping Developmental AIs

From Simple Competences to Intelligent, Human-Compatible AIs

Mark Stefik and Robert Price[1]

Version April 3, 2024

Developmental AI creates embodied AIs that develop human-like abilities. The AIs start with innate competences and learn more by interacting with the world including people. Developmental AIs have been demonstrated, but their abilities so far do not surpass those of pre-toddler children.

In contrast, mainstream approaches have led to impressive feats and commercially valuable AI systems. The approaches include deep learning and generative AI (e.g., large language models) and manually constructed symbolic modeling. However, manually constructed AIs tend to be brittle even in circumscribed domains. Generative AIs are helpful on average, but they can make strange mistakes and not notice them. Not learning from their experience in the world, they can lack common sense and social alignment.

This position paper lays out prospects, gaps, and challenges for a bootstrapping approach to developmental AI that follows a bio-inspired trajectory. The approach creates experiential foundation models for human-compatible AIs. A virtuous multidisciplinary research cycle has led to developmental AIs with capabilities for multimodal perception, object recognition, and manipulation. Computational models for hierarchical planning, abstraction discovery, curiosity, and language acquisition exist but need to be adapted to an embodied learning approach. The remaining gaps include nonverbal communication, speech, reading, and writing. These competences enable people to acquire socially developed competences.

Aspirationally, developmental AIs would learn, share what they learn, and collaborate to achieve high standards. They would learn to communicate, establish common ground, read critically, consider the provenance of information, test hypotheses, and collaborate. The approach would make the training of AIs more democratic.

[1] Intelligent Systems Lab, PARC, part of SRI International, Palo Alto, California, 94304

# Contents

# 1. Introduction

Impressive research results, systems, and infrastructure have become widely available based on generative approaches, large language models (LLMs). Foundation models are trained at massive scale over multimodal data that can include text, code, and images (e.g., Brown et al, 2020; Huang et al, 2022; Zeng et al., 2022)). Augmented by in-context learning (e.g., Liu et al., 2023; Ahn et al., 2022) and manually engineered models, these models provide a starting point to train other AIs.

Mainstream AIs increasingly operate in situations where they take actions that can affect people adversely. For example, noticing a ball rolling down a sidewalk, self-driving cars should slow down and be alert for children, adults, and pets that might chase the ball into the street. They should never drive into people or drag them down the street.

The current surge of interest in LLMs feels like a land rush. Three articles on this point were published in the *New York Times* in May and June 2023. Their writers sought to understand and communicate the new power, importance, and risks of AI. One article reported on how tech entrepreneurs were returning to San Francisco *en masse* from remote work to participate in the latest AI boom (Griffith, 2023). Another reported on an attorney who faced judicial sanctions because the legal materials that he submitted for a court case included opinions and citations that an AI made up (Weiser, 2023). Meanwhile, the valuation of the graphics and AI chip company NVIDIA soared (Warner, 2023).

In summary, with vast data and computational resources, deep learning methods have demonstrated successes where manual approaches of earlier AI technology could not.

But are current approaches to machine learning good enough?

Consider Fei-Fei Li's bronze horse story in the sidebar. Automatically writing photo captions was a tour de force. The demonstration built on methods and photo and language resources developed over several years. Over time the photo resource grew to include millions of photos and the language resource grew to encompass dictionaries and WordNet (Deng et al., 2009). This line of research developed crucial insights about the power of "big

> ### *Lessons from a Bronze Horse*
>
> Stanford University professor Fei-Fei Li recounts an experience from her research lab about an AI system that creates captions for photographs (Li, 2018; 2024; Karpathy & Li, 2015). Her student Andrej Karpathy was demonstrating his latest results to her.
>
> "I hurried in to see the latest. … On the screen was a photo of a teenager and a skateboard, both in midair, against a backdrop of blue sky and distant bushes. In a tiny command-line window beneath the image, a sentence was printed out.
>
>  'A person on a skateboard.'
>
> I was smiling before I even realized it. Andrej let the moment linger for a second, then pressed a key, Another image appeared, this one depicting a messy construction site with two workers in orange vets pouring cement. It was followed after a second or two by another sentence.
>
>  'Construction workers working on the curbside.'
>
> He hit the key again. Another image, another caption. Then another, and another, and another. The quantity and variety of the scenes made it clear that these scenes weren't just being dug up from a training corpus somewhere. The model was writing them. …
>
> He clicked again, and a new image came up, snapped by a tourist in a rustic Spanish plaza …
>
>  'A man riding a horse down a street next to a building.'
>
> … we both laughed at the near-perfect description and its sole, crucial omission: that the man and the horse were made of bronze." (Li, 2024) data" and triggered widespread enthusiasm for deep learning approaches (Bengio, et al, 2021; Li, 2023).

Reflecting on the bronze horse example yields additional insights. Why do people laugh when they see its caption: "man on a horse?" At root, human experiences of animals and models of them are experiential, multi-sensory, and 3D. People routinely see human and animal sculptures in public spaces. As children they play with toy farm animals. They may ride on ponies and on the wooden horses of merry-go-rounds. People know that a real horse would not stand still on a pedestal in a public square.

Although such machine learning techniques harvest vast data from the internet, training with such data is not the same as gathering data continuously in the real world based on context and curiosity, using that data to guide future real world actions, and tuning what has been learned depending on outcomes.

In his 1990 AAAI presidential address, Daniel Bobrow colorfully cautioned about the limitations of manually-constructed symbolic AI systems:

> "Newell's analysis postulated a single agent, … disconnected from the world, with neither sensors nor effectors, and more importantly with no connection to other … intelligent agents. … intelligent but deaf, blind, and paraplegic …" (Bobrow, 1991).

Bobrow's colorful description does not discuss machine learning or the importance of learning by interacting with the world. Nonetheless, it begins to explain why the automatic captioning system and today's mainstream AI approaches can make strange mistakes. In Bobrow's characterization, the then current AIs were like pedantic children who had book knowledge but lacked real world experience.

The side bar compares three generations of AI technologies at a high level. It casts AI progress on a trajectory from manual construction, to learning from vast (disembodied) data, to learning continuously from direct experiences and from socially developed information that was recorded or written.

> **Three AI Approaches and their Limitations**
>
> Current AI approaches are not individually adequate to create robust AIs or AI collaborators.
>
> (1) *Traditional* symbolic AI systems require extensive hand-engineering of knowledge and interaction rules to support narrow domains. They lack robustness when they face unanticipated situations and can be updated only by teams of experts.
>
> (2) *Generative* AI systems built on large foundation models draw on vast amounts of socially developed information on the web, but they learn in a decontextualized way. They do not know the source of the information, whether it is trustworthy, when different elements can safely be combined, or whether it is consistent with common sense. Foundation models are trained on datasets scraped from the web by masking out part of the input and having the network reconstruct the missing part. This approach explicitly trains models to produce text that looks plausible. The result, however, is assistants that are statistically good but occasionally and strangely catastrophic in their predictions and lack of understanding. They are unable to update their world models during interactions except through context learning.
>
> (3) *Developmental* AI systems develop competences like human children, benefit from real-world grounding, and acquire elements of commonsense. However, this approach has not yet scaled to adult-level team and task competencies. When development stops at the *communication gaps*, the AI agents do not acquire socially developed competences.

AI systems are now being developed for real world applications where their actions have human consequences. Developers race to augment applications and turn them loose. The AIs' decisions go unvetted and unchecked. Lacking a comprehensive approach to testing and validation, it is tempting to create test cases and then to patch the systems to fix whatever bugs and deficiencies are found. *This testing and validation process is like a game of whack-a-mole.* You never know what problems will pop up in other situations, what the training misses, or what side effects a patch will cause.

Focusing on easy automation diverts attention from a difficult and substantially unaddressed challenge. The challenge is to accurately fuse information from multiple information sources that have diverse provenance, reliability, and applicability.[2] For some entrepreneurs and others, this caution is not welcome. It can lead to delays to create stronger approaches for system validation and testing.

When an AI incorporates information without testing and validating it, the incorporation creates a security attack surface for misinformation and cyber-attacks. For example, it can enable bad actors to poison and bias the information used for learning. Current practices fine tune foundation models for better performance in specific applications and use cases. However, even modest changes can unintentionally reduce the

### A Sampling of Recent Rethinking of AI Approaches

- (Landay et al., 2023) James Landay, Fei-Fei Li, and Eric Horvitz discuss challenges and possible futures for generative AIs and embodied AIs.
- (Cangelosi et al., 2022, 2015, 2010) and (Asada, 2024) review multidisciplinary foundations and the state of the art of developmental robotics.
- (LeCun, 2022) proposes an architecture and training for autonomous AIs to learn percepts and action plans at multiple levels of abstraction. It incorporates intrinsic motivation, predictive world models, and hierarchical embeddings.
- (Spelke, 2022) reviews and analyzes multidisciplinary research about what babies know and how they learn.
- (Dubova, 2022) proposes grounded components for linguistic intelligence to move AI from a point where it is now, failing to generalize to new data, failing to capture common sense, and failing in communication settings.
- (Zador, et al., 2022) Overview of research in neuroscience providing guidance and examples on requirements and properties needed for flexible and embodied intelligence beyond mainstream AI.
- (Bonmasani et al., 2021) offer analyses and cautions about risks in the broad and uncritical reuse of foundation models to create AIs.
- (Hawkins, 2021) proposes that a computational model of the neocortex with reference frames, continuous learning, and voting of multiple models is better for flexible and general intelligence than current NNs.
- (Hinton et al., 2021, 2018) propose "capsules" to represent part-whole relationships and distinct object properties in artificial neural networks and improve their flexibility for general AI.
- (Ullman and. Tenenbaum, 2020) review prospects for a unifying Bayesian computational framework for understanding how children build and use cognitive models of the world.
- (Levine, 2019) explores ways that robots could collectively pool their knowledge to share what they learn.
- (Marcus and Davis, 2019) review current mainstream AI methods for creating AI systems and how these methods fail to lead to reliable AIs with deep understanding.
- (Mao et al., 2019) propose multimodal approaches and neurosymbolic representations for AIs to learn visual concepts and language.
- (Russell, 2019) reviews the state of the art in AI and analyzes requirements for making AIs more human compatible and the risks created by not doing that.
- (Pearl, 2019) advocates an approach for representing causal models of the world, arguing that machine learning approaches fail in their promise because they do not capture or enable causal reasoning.
- (Smith, 2019) proposes object centric representations for understanding the physical dynamics of 3D scenes.
- (Oudeyer et al., 2017 a-b, 2016, 2006) Grounded in linguistic research and developmental robotics, these papers review intrinsic motivation systems, learning about human development from baby robots, and learning for autonomous robots.

security of foundation models (Henderson et al., 2024). Like thoughtful investigators, future AIs should treat external information resources as provisional, subject to skepticism and testing. They should employ critical analysis and skepticism, and hone robust world models. Embodied approaches enable AIs to experiment with the world and refine their world models.

Skilled people use personal experience, knowledge about the provenance of knowledge, and high-level causal reasoning to evaluate the information that they use. Their skills for doing this come from many social experiences.

The relevant human experiences start in the toddler years and include playground interactions and early speech. Children acquire social skills when they interact with others such as during play. They hear and tell jokes. They tease and are teased. They learn that different people have different goals. They learn that not everything that others say is true. They meet people who have different world views, goals, and beliefs. As they learn about people, they hone their analytic and skeptical skills. They develop social, skeptical, and investigative skills that help them to decide what people, information and sources to trust.

---

[2] Subhabrata Dutta and Tanmoy Chakraborty published an opinion piece "Thus Spake ChatGPT" in the *Communications of the ACM* that clearly illustrates such problems with GPT-3 using a series of examples of its generated responses to detailed questions of science. As they characterize the issues, the chat program generated "polished, confident [but incorrect] textual responses to queries [that] lack factual consistency." As they explain this "popular majority versus authentic minority" problem, "a language model learns from the majority of examples while rigorous scientific truth is mostly a minority." ChatGPT's inability to cite sources makes the testing of accuracy much more difficult for readers.

There is a history of AI researchers stepping back to rethink the goals, successes, and challenges of AI approaches. A sidebar gives a sampling of leading researchers and their proposals for pivots in approaches to AI. Although they recommend different toolboxes or conceptual frameworks (e.g., reinforcement learning, differentiable models, multimodal foundation models, neuro-symbolic models, causal models, cognitive models, or probabilistic models), there is much overlap in the recognized limitations of mainstream AI approaches. Many of them call out limitations such as the lack of grounding, lack of context awareness, lack of common sense, and lack of understanding about the provenance, trustworthiness, and relevance of the information that they combine and use.

Historically, engineering and scientific approaches divide a large area of knowledge into subareas that are studied independently. Babies do otherwise. Why does their simultaneous multi-pronged bootstrapping approach for developing intelligence work better than one-competence-at-a-time reductionism?

As Linda Smith says in her foreword to *Developmental Robotics* (Cangelosi & Schlesinger, 2015):

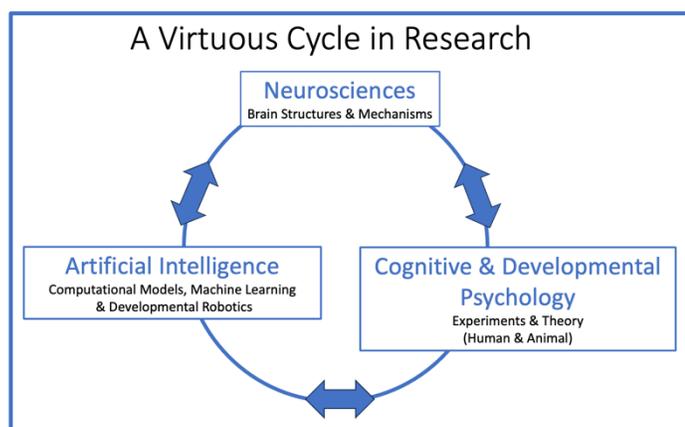

*Figure 1. A virtuous research cycle.*

> "The dominant method of science is analysis and simplification. This was clearly articulated by Descartes in 1628: In studying any phenomenon, simplify it to its essential components dissecting away everything else. This approach is motivated by the belief that complicated systems will be best understood at the lowest possible level. By reducing explanations to the smallest possible entities, the hope is that we will find entities that are simple enough to fully analyze and explain. The spectacular success of this methodology in modern science is undeniable. *Unfortunately, it has not given us an understanding of how systems made up of simple elements can operate with sufficient complexity to be autonomous agents*. [emphasis added]" (Smith, 2015)

Smith's observation poses a challenge for AI to do better than reductionism. What competences are required for developmental AIs that learn what they need? What do they do and why are they needed? How do they collectively and dynamically enable human-compatible intelligent agents?

This paper draws on decades of work by researchers operating in the multidisciplinary virtuous research cycle in Figure 1.[3] It lays out prospects, gaps, and challenges of a bootstrapping developmental approach and proposes a roadmap for developmental AI research.

---

[3] This virtuous research cycle is often described as including neuroscience, developmental and cognitive psychology, and artificial intelligence. Later in this position paper, we add research in education to the research community of this virtuous cycle for creating experiential training experiences and testing developmental AIs.

# 2 Towards AI Systems that can Learn like Children

Remarkably, 400,000 intelligent agents are created every day. Within their first two years, most of these agents can perceive the world, model how it works, take effective actions, reason using goals and plans, understand and speak a community's natural language, and begin to collaborate. How do babies do it?

## 2.1 What Young Children Know and Can Do

Before children can walk, they learn to recognize objects. They learn to move their hands. They learn to push themselves up, to sit, to roll around, to crawl, and to reach a standing position. When they are standing and move one leg forward, they notice that it stops them from falling. Advancing on a *trajectory* of interdependent skills, children learn to walk.

Just as there is a trajectory of skills for people learning to walk, there is a trajectory of skills for learning to communicate, to read, to do everyday tasks, to collaborate, and to carry out expert professional work. These trajectories are effective, reliable, and largely overlooked in mainstream AI.

In contrast, mainstream AI methods attempt to create AIs with expert and *adult-level* capabilities without building up to those levels from more primitive levels.

Consider the fictional home scenario in Figure 2 and photographs from a similar research scenario.

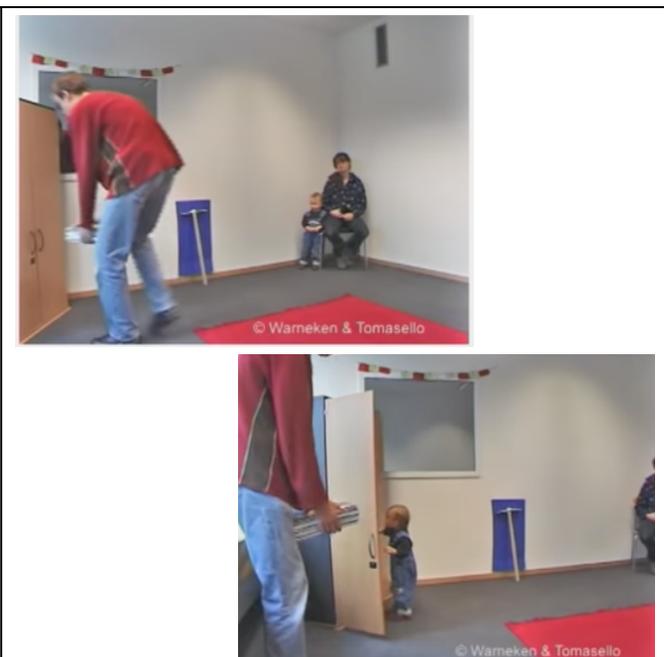

**Opening a Door for Dad**

Macer is a toddler. He is on the floor building a tower of blocks. He sees his dad (Nick) tape a ribbon to the last wrapped package on the table. Nick puts the ribbon, scotch tape, and wrapping paper in a box.

Gathering up the wrapped packages, Nick walks to the hall closet. He plans to hide the presents in the closet so that Macer's mom will not see them before her birthday celebration that evening.

Standing by the closet door, Nick struggles to hold all of the packages with one arm. He looks at Macer.

Macer gets up from playing with his blocks on the floor. He runs to his dad. Macer opens the closet door and holds it open.

*Figure 2. Fictional scenario of a collaborating toddler and frames from a research video from Warneken & Tomasello (referenced in Tenenbaum, 2019).*

For parents and other people familiar with small children, the scenario is not surprising. Toddlers often join their parents in activities such as putting toys away and helping to prepare food. For parents, such toddler learning is natural and a relief. From the perspective of AI research, however, the scenario is challenging *because the performing AI needs to know so much*.

The scenario begins with a toddler building a tower of blocks. To replicate this behavior an AI would need to know about the interactions of blocks and gravity, and have competences for seeing, grasping, and stacking blocks.

The next part of the scenario involves understanding Nick's goals. For example, the AI needs to recognize that Nick has a goal of putting packages in the closet.

The sidebar lists some competences that Macer demonstrates in the scenario. These include inferring

> **Example Competences in the *Opening a Door for Dad* Scenario**
>
> - There are other agents in the world that do things and have plans.
> - Nick is an agent. His body is like Macer's body.
> - Nick wants to put the presents in the closet.
> - Nick has a multi-step plan based on several goals and constraints.
>   - Nick does not want to drop the packages.
>   - He needs to use both hands to hold the packages.
>   - To open the closet door, Nick would need to use one of his hands.
>   - If Nick frees a hand to open the closet door, some packages will fall.
> - If Nick does not get help, his plan will fail.
> - Macer could open the closet door for Nick.

that Nick's hands are full and that he would probably drop the packages if he used a hand to open the closet door. An AI would need to understand what Nick intends to do, how he is likely to do it, and that he will likely fail in his plan. The AI needs to recognize what Nick needs and then offer to help.

The toddler's performance seems easy because adults can do this without difficulty and have seen kids doing it. The example shows how ordinary life can require knowing many things.

Twenty four months earlier, Macer was a newborn. At that time, he could barely see or move. He had never seen blocks or thought about stacking them. Using his primitive vision, he could not distinguish between Nick carrying boxes and shadows on the wall.

In 2000, the National Research Council and Institute of Medicine published *From Neurons to Neighborhoods* (Shonkoff & Phillips, 2000). The report begins by reflecting on what children learn in their first years of life:

> "From the moment of conception to the initial, tentative step into a kindergarten classroom, early childhood development takes place at a rate that exceeds any other stage of life. The capacity to learn and absorb is simply astonishing in these first years of life." (Shonkoff & Phillips, 2000)

Toddlers learn a lot in their first two years. Understanding exactly what they know is not simple, as Elizabeth Spelke discusses in *What Babies Know* (Spelke, 2022). At two years, toddlers have very limited abilities to speak and communicate. How can we know what they are thinking and learning?

A few years after the National Research Council report, child development psychologists Celia Brownell and Claire Kopp reflected on this scientific challenge:

> "Many disagreements about the toddler period revolve around two related issues, the nature of competence and skill at a given age and how to characterize developmental change more broadly. One core problem is that we must try to infer what lies in the mind of a small, inscrutable being who looks, and even sometimes behaves, very much like one of us. But at the same time this being may also act very much like one of our pets, and not very much like an older human child at all. The intellectual temptation to 'adultomorphize' the toddler is tantalizing and often hard to resist. … Unfortunately, few methodologies and procedures can adjudicate such interpretive disagreements in preverbal children, and much good scholarship is wasted on argument without the necessary clarifying evidence." (Brownell & Kopp, 2007)

Brownell and Kopp's comparison of toddler and pet cognition is apt. Many of the childhood phenomena about intelligence and learning occur in animals as well. Christopher Völter and Ludwig Huber reported on findings (Völter & Huber, 2017) from eye tracker experiments with dogs as subjects, using methods like those used in research studies of human children. Based on pupil dilation experiments, they found that dogs like toddlers have implicit expectations about contact causality. Restated, dogs and children learn initial physics models about the properties and interactions of objects in much the same way. Without a physics model, a dog could not catch a frisbee or avoid obstacles. The experiments reveal that like children, dogs develop models over time as they interact with the world. Follow on experiments could study differences in how dogs develop world models and how children do. The order of competence development in primates is also very close to that of humans, resulting in "mosaic organizational heterochrony" (e.g., Langer, 2001), where differences in timing can result in substantial differences in the cognitive development of primates.

The next section describes how instrumentation, experimental design, and computational modeling have advanced significantly in the last decade, providing insights and enabling better modeling of "the mind of the small inscrutable being."

### *2.2 Advances in Instrumentation and Modeling*

Advances in instrumentation enable more sophisticated and fine-grained studies of brain activity including changes in behavior during cognitive development and learning. Over the last two decades, there have been many advances in instrumentation including eye tracking (e.g., Eckstein et al., 2017; Smith et al., 2019) and brain scanning technologies (e.g., fMRI, EEG, and MEG) (e.g., Pollatou et al., 2022; Kim et al., 2021; Hauk & Weiss, 2020; Raschle et al., 2012; Knickmeyer et al., 2008). Neuroimaging experiments with fMRI are also conducted for understanding brain activity when people teleoperate robots (e.g., Menon et al., 2017).[4] A sidebar describes current scanning technologies.

---

[4] This paper by Samir Menon, Amaury Soviche, Jananan Mithrakumar, Alok Subbarao, and Oussama Khatib illustrates an overlap of methodology and experimentation for the related purposes of creating teleoperated robots and creating autonomous AI robots. By using haptic interfaces with fMRI, the experiment collects data for both sensing and control. In this example, the researchers demonstrated the feasibility of collecting neuroimaging data about high-fidelity five-axis force perception and motor control. This research supports constructing computational models of human neural response.

By combining the use of scanning technologies with experiments that set up situations, researchers can probe moment-to-moment mental activities. Violation of Expectation (VOE) experiments log precisely what an infant sees, measure when its pupils dilate, and determine where an infant directs its focus. VOE is now used to probe a wide range of child learning events in controlled experiments (Jackson & Sirois, 2022; Stahl & Feigenson, 2017).

Consider Jean Piaget's experiments in the 1930s when he studied peek-a-boo, and he developed his constructivist theory about how children construct models of the world (Piaget, 1954). In a typical version of the game, a mom hides her face behind a blanket. She then drops the blanket, pops back into the baby's view, and says "Peek-a-boo!" Piaget reported that children learn about *object permanence* – the concept that objects can continue to exist even when they move out of sight. Mom may disappear behind something. But she's not necessarily gone. She can reappear in a predictable way.

> **Brain Scanning Technologies**
>
> Tomography is a general term for technologies that image a cross section. CT refers to computer tomography, where a computer controls the scan, and collects and organizes the image data.
> - **EEGs** (electrocephalograms) and **ERPS** (event related potentials) measure changes in electrical potentials of electrodes. ERPs are noninvasive and require that the subject wears a net.
> - **MEGs** measure magnetic field changes. The subject wears a helmet. The technique requires a magnetically sealed room.
> - **fMRI** (functional magnetic resonance imaging) tracks brain activity as a function of blood flow. It has better spatial resolution than ERP. The subject lies in the tube of an MRI machine. fMRI measures small changes in blood flow during brain activity. Cognitive experiments using fMRI have a subject performing a set cognitive activity while blood flow is measured in parts of the brain. When an area of the brain is active, blood flow increases.
> - **NIRS** (near-infrared spectroscopy) measures transmission of light through the brain tissue as affected by hemoglobin concentration changes.
> - **PET** (positron emission tomography) scan can show the metabolic or biochemical function of tissues and organs. It uses a radioactive tracer drug to show metabolic activities.

Using VOE experiments today, researchers can probe fine grained transitions in children's behavior, such as what children at a particular age know about contact physics (how objects interact when they bump each other), and the movement and stability of objects, and how they explore *object persistence*. VOE experiments enable them to track limitations and advances in children's cognitive development. Looking ahead, VOE experiments and analogs of brain instrumentation provide examples and inspiration for experiments and approaches that can be adapted for benchmarking the cognitive development of AI systems.

Besides instrumentation, computational models of cognition are developed in cognitive psychology and AI. Such models have been developed for many years (e.g., Wiener, 1948; Selfridge, 1959; Feigenbaum & Feldman, 1963; Newell & Simon, 1972; Rumelhart et al., 1986; Tenenbaum, 2020). Today's models include functional models ranging from neural levels to high level cognition.

Computational models are a natural point of collaboration between child computational psychologists and AI researchers. As Michael Frank, a developmental psychology professor at Stanford observed:

> "We try to look … from both the perspective of infants and infant development and the perspective of artificial agents – AIs and robots that are trying to learn and explore.
>
> As a developmental psychologist what I focus on is characterizing children's behaviors and trying to build models that explain them. That has a really interesting synergy with the engineering in artificial intelligence. …
>
> So, we're really trying to translate between the insights of behavior in developmental science all the way through engineering and robotics. We have found that the observation of children's behavior feeds into AI. [Then] AI comes back and helps inform the tools that we use to analyze children's behaviors. Those kinds of virtuous cycles are something that I never anticipated in this line of work." (Frank, 2022)

In summary, the last two decades of research in the virtuous cycle have yielded increased understanding and computer models for learning how the world works, learning how to perform actions, and learning how to interact with and collaborate with people. Looking ahead, how might human-compatible AIs that learn from experience and learn from people better address future needs? How might a bootstrapping developmental approach be extended to incorporate AIs that have super-human perception, powerful computation, and large information resources? How could such AIs better address society's future needs?

# 3 What Kinds of Computations are Competences?

A *competence* is the ability to do something well. For example, in preparing a meal a chef needs to crack eggs, create attractive presentations of food, make desserts, find or create interesting recipes, and so on. Having a competence implies that an agent has the knowledge and the physical means to exercise it.

A *model* is a runnable physical, mathematical, or logical representation that expresses knowledge and relations in a domain. A computational model can provide detailed spatial, temporal, and behavior predictions about a cognitive competence. In AIs, computational models are typically programs used to implement competences such as for interpreting perceptual data, reasoning, or acting.

### *3.1 Innate, Self-Developed, and Socially Developed Competences*

Future AI systems for diverse applications and environments will need diverse competences. For example, AI robots will be used for working in factories, in package delivery, and in homes. They will be used under water, in zero gravity, and in extreme and dangerous environments. Robots for manipulating objects under water or in zero or low gravity need different competences from robots in earth normal situations.

There are three processes for acquiring competences corresponding to innate competences, self-developed competences, and socially developed competences.[5]

*Innate competences* are present from an individual's beginning. They correspond to knowledge acquired during the lives of ancestors. Natural innate competences develop via evolution.[6] Innate competences are coded in genes. Many animals in the wild face intense survival pressures. Within hours of being born, the precocial offspring of grazing species including deer, horses, goats, and giraffes can walk to avoid predators. Snake hatchlings can feed themselves shortly after they hatch. In contrast, human infants are relatively helpless, but they can learn to thrive in a variety of environments.

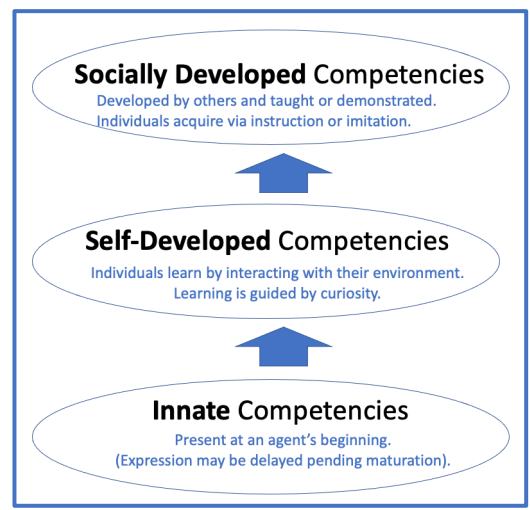

*Figure 3. Each level enables and supports the next higher level of competences.*

---

[5] Innate competences are present from at an individual's beginning as distinct from competences that are learned later. For the purposes of this paper, it is useful to distinguish the three kinds of competence according to the three methods of acquisition.

[6] Based on readily available review articles (e.g., Schopf, 2007), the earth formed about 4.5 billion years ago and single celled life appeared about 3.7 billion years ago. Plants and photosynthesis appeared a billion years later. Mammals appeared about 210-250 million years ago. Estimates of when the first humans appeared vary from about 130,000 years ago to 2.5 million years ago. These numbers are subject to definitions, interpretation of data, and ongoing research. Agriculture was invented about 20,000 years ago and writing 500 years ago.

> **The Richness of the "Child as Scientist" Metaphor**
>
> Children and adults learn about the world by interacting with it as described by Allison Gopnik et al. and others. Infant learning is largely about perception and motor control, or "sensorimotor" learning as Piaget called it. Cognitive mechanisms for pattern recognition and prediction power this learning but additional learning abilities are involved for building models of the world. The trajectory of advancing capabilities includes generalizing, forming abstractions, and curiosity. Scientists, other people, and members of other species draw on *socially developed* knowledge of others. This requires communication skills which themselves need to be learned.

*Self-developed competences* are acquired by individuals as they learn from their experiences. They represent knowledge learned as they interact in their current contexts and build models of the world including other individuals. Species where individuals learn by interacting with the environment develop new competences many orders of magnitude more rapidly than evolution can.

Children start with some competences and acquire new ones. They grow up, learn local languages, develop social relationships, and collaborate. They are born into different cultures including nomadic and tribal cultures. They live in deserts, on boats, in agricultural settings, and in cities. They grow up in families with different child-rearing practices.

*Socially developed competences* are developed and held by other individuals and groups. They can be taught, learned by imitation, or learned by reading and interacting with media.

The three levels in **Error! Reference source not found.** correspond to different processes for acquiring competences.

In *The Scientist in the Crib,* Alison Gopnik and her colleagues credit three factors for driving the rapid development of children in their first years of life: innate knowledge, superior learning ability, and human caretakers shaped by evolution to teach them effectively. These factors correspond to the three levels in **Error! Reference source not found.**. As the authors put it:

> "The question, as always, is how do [kids] do it? The answer … is that they are born knowing a great deal, they learn more, and we are designed to teach them." (Gopnik et al., 1999.

Future AI agents will need competences at all three levels.

- Without *socially developed* competences AIs could not learn from others or take advice. They would need to discover everything on their own.
- Without *self-developed* competences, they could not change their behavior based on learning from their experiences.
- Without *innate* competences, they could not function at all.

Akin to bootstrapping[7] concepts for computer systems, developmental psychology studies how primitive competences develop to more robust ones. Developmental psychologists such as Jean Piaget (Piaget, 1954) characterized children's cognitive development in terms of them building models of the world. Susan Carey and others (Carey, 2004; Carey 2009) studied how higher-level mathematical concepts such as integer, number, and Abelian group are developed, represented, and understood by children bootstrapping from innate competences. Michael Tomasello (Tomasello, 2009; Tomasello, 2019) and others studied how social cognition arises from the competences of individuals, how it leads to socially constructed behaviors and representations, and how it provides competitive advantages for groups and their members.

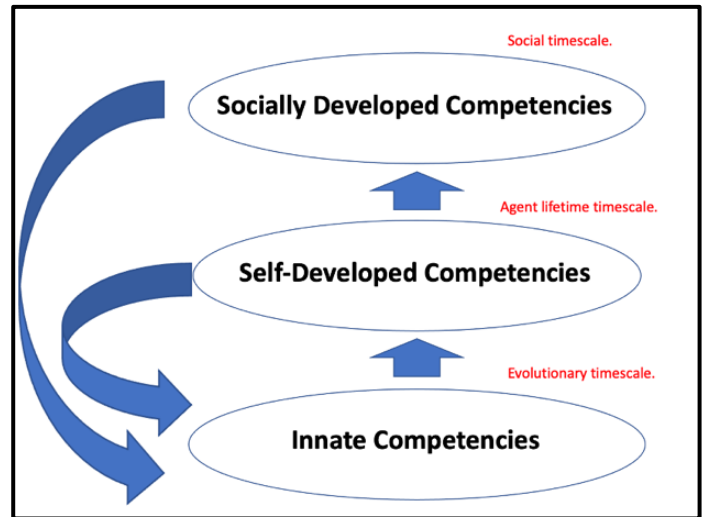

*Figure 4. Copying memory of multiple kinds of competences.*

Figure 4 illustrates that self-learned competences can potentially be propagated to new computational agents by copying their representations. Such direct copying presents an opportunity for mass producing AIs and is akin to the idea of foundation classes as discussed later. It would enable the creation of AIs with complex competences *without the need for every new individual to have the early learning experiences*.[8] By analogy with the terminology of foundation models, an aspirational product of this methodology is to create *experiential foundation models*, that is, computational models that carry knowledge learned through an AI's embodied experiences of interacting with the world and people.[9]

Compared to other species, humans have greater abilities to learn extensively from other individuals. Social learnings are distilled and propagated to others via teaching, imitation, and publication. Passing along socially developed competences avoids requiring individuals to discover, learn and choose what to investigate on their own.

---

[7] The term "bootstrapping" comes from the expression "pull oneself up by one's bootstraps." In computers "bootstrapping" or "booting" refers to the loading of initial software into memory after turning on power. A simple loader (built in or entered manually) reads in a larger program. The bootstrapping program loads larger and more complex code in stages including an operating system and applications as needed.

[8] A caution about overusing the copying approach is that a population of AIs can be more robust to security vulnerabilities when there is diversity in agent competences and training. The collective robustness of a population of AIs can be improved in copying approaches by including a wide base of different individuals.

[9] A review and development of formal theories of experiential learning is beyond the scope of this paper. Interested readers may find Thomas Howard Morris's review of experiential learning theory helpful (Morris, 2019). He suggests that cognitive concepts such as embodiment are needed to put this theory on a sound theoretical footing.

Socially developed competences have *internal representations* for modeling and *external representations* for communication. Examples of external representations include spoken language, written language, gestures, and observable behaviors of others both live and recorded. Individuals perceive external representations and create internal representations for interpretation, learning, integration, and use. Social competences are generated, tested, published and taught by teachers and social institutions. The competency representations can be the product of group activities over multiple lifetimes. Depending on their physical properties (e.g., books, movies, and digital services), the external representations of social competences can outlast individuals and benefit multiple generations.

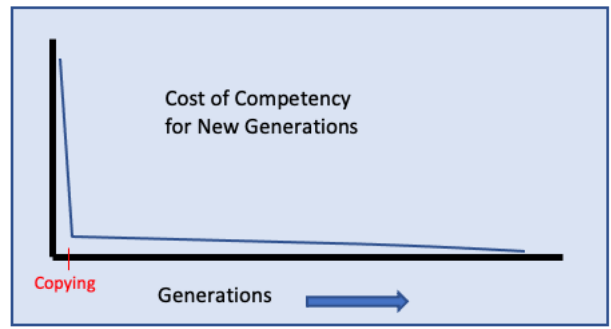

*Figure 5. Notional incremental cost of acquiring a competency for a new generation.*

Figure 5 suggests how copying can amortize development costs.[10] Each layer of competences bootstraps the development of competences at the next higher layer.[11]

In summary, human cognitive abilities are acquired by three pathways. Humans are born with *innate competences* developed over evolution. They learn *self-developed competences* by interacting and experimenting with the world. *Socially developed competences* are learned by some people and then taught to others or imitated by them.

---

[10] Another potential way to reduce the cost of raising robots is to use "robot nannies" and "robot teachers" that are expertly trained and collaborate with diverse human teams. This approach could also provide a source of diversity in AI collaborators.

[11] In this way, a copy of an AI's memory is a foundation for the copied AI. The terminology for such copying and its relation to different kinds of foundations has not settled down. This research is at a leading edge of AI and robotics and there is much exploration and variation in the technology and terminology (e.g., see Morris et. al, 2024; Majumdar et al., 2024).

### *3.2 Core Competences*

In 2018, Rodney Brooks proposed four "super intelligence" goals for AI research:

> "The object recognition capabilities of a two-year-old.
> The language understanding capabilities of a four-year-old.
> The manual dexterity of a six-year-old.
> The social understanding of an eight-year-old." (Brooks, 2018)

These goals trace back to his "Intelligence without Representation" paper (Brooks, 1991) that advocated creating AIs that replicate human-level intelligence by learning via interactions with the external world. A key question for bootstrapping competences is "What does an agent start with?" What innate abilities are present or latent when human babies are born?

Research by Elizabeth Spelke and others describes how infants begin life with multiple innate competences (Spelke, 2022; Spelke & Kinzler, 2007; Carey, 2009). Although Spelke refers to these competences as "core foundations" for intelligence.[12] The book *What Babies Know* Spelke reviews open questions about competences for vision, objects, places, numbers, forms, agents, social cognition, and language. All humans have these competences. As they mature, people extend early versions of these competences to adult-level competences. Core competences are innate and similar to those found in other mammals and other species.

For example, one of Spelke's core foundations is Objects.[13]

> "Objects are the primary things that we perceive, act on, categorize, name, count and track over time. Like the perception of surfaces in depth, the experience of unitary, bounded, solid objects arises immediately and effortlessly. … We experience objects as standing in front of a background that continues behind them, even in pictures … and as complete, solid bodies, although their backs are hidden." (Spelke, 2022)

The human Objects foundation is complex. Objects can have parts. Objects retain many properties when they move. For example, simple solid objects typically do not perceptively change shape when they move. The Objects concept governs how we understand things in the world. Object competences are connected to the impulse to dodge or duck when a flying object is coming our way. The Objects foundation is core, but it is not tiny or fixed. As people have more experiences, they add to what they know about objects.

---

[12] The term "foundation models" is popularly used in the context of generative machine learning (Bonmasani et al., 2021; Bonmasani & Liang, 2021). Large foundation models are created for natural language and computer vision applications by computations over large datasets. Bonmasani and his colleagues at Stanford's Human-centered Artificial Intelligence (HAI) program describe advantages and issues with this approach. In the context of bootstrapping as described in this paper, we propose the term *experiential foundation models* to refer to computational models of competences created by embodied AIs via their interactions with the world including people.

[13] An "Object concept" was often located at the top of a class hierarchy in early object-oriented and knowledge representation languages. These representations and the theoretical issues of early simulation, programming, and knowledge representation research were different from and simpler than Spelke's core Object foundations. In contrast to the theoretical frameworks of embedded systems, the foundations of early AI symbolic representations were based on logic but not perception or communication.

| Domain | Description |
| --- | --- |
| Objects | supports reasoning about objects and the laws of physics |
| Agents | supports reasoning about agents that act autonomously to pursue goals |
| Places | supports navigation and spatial reasoning around an environment |
| Number | supports reasoning about quantity and how many things are present |
| Geometry | supports representation of shapes and their affordances |
| Social World | supports reasoning about Theory of Mind and social interactions |

*Table 1. Core Knowledge Concepts used in DARPA's Machine Common Sense (MCS) Program.*

Core foundational competences have also been used in defining benchmarks and milestones for programmatic investigations of common sense. Table 1 shows the areas of core knowledge that were proposed for DARPA's Machine Common Sense (MCS) program (DARPA, 2018; Gunning, 2018).

The innate competences for bootstrapping AIs can differ from the core foundational competences of humans. Goals for AI research include bootstrapping AIs to high-level competences, characterizing competences in detail using analytic tools for understanding computational models, and creating training experiences involving situations and sensory data potentially including potentially senses beyond those of natural organisms.

The following scenarios illustrate interactions of core competences in two everyday situations.

> **Encounter in a Kitchen**
>
> Jack and Jill carry their glasses to the kitchen sink. They arrive at about the same time. Jack gestures for Jill to fill her glass first. Jill reaches for the faucet and turns on the water. The hot water splashes on her hand. She jumps back and drops her glass in the sink. Jack adjusts the water faucet for colder temperatures, picks up Jill's glass, and hands it to her. Jill fills her glass. She smiles at Jack, thanks him, fills her glass, and returns to the party.
>
> **Encounter at a Store**
>
> Jill and Jack are each carrying packages as they approach an exit door of a department store. Jack is struggling with his packages. Jill holds the door open and gestures for Jack to go through first. As Jack steps through the doorway his shoe catches on a grating. He stumbles and drops one of his packages. Jill leans against the door to hold it open. She picks up Jack's package and hands it to him. Jack smiles and thanks her. They continue out to the sidewalk.

Jack and Jill know about the properties of hot and cold water and gravity's effects on packages and glasses. In the MCS categories, this knowledge is in an Objects model. They also know about kitchens and stores (Places model), activities, plans, getting a drink at a sink, exiting a store, and repairing a failed plan (Agents model). Knowledge about collaboration between people would be in a Social World model.

In summary, adult-level intelligence depends on multiple competences. Adult-level ensembles of sub-competences develop over time.

### *3.3 Multi-Model Information Fusion*

Joshua Tenenbaum and his colleagues raised fundamental questions in their investigations of people and their mental models of the world (Tenenbaum, et al., 2011):

> "In coming to understand the world—in learning concepts, acquiring language, and grasping causal relations—our minds make inferences that appear to go far beyond the data available. How do we do it? … How do our minds get so much from so little? … If the mind goes beyond the data given, another source of information must make up the difference." (Tenenbaum, et al., 2011)

### *3.3.1 A Blindfolded Treasure Hunt*

As a thought experiment for investigating these questions, consider the following scenario:

> **A Blindfolded Treasure Hunt**
>
> The birthday party was Jill's idea. Jack was in the entrance hall of the 1900's house that Jill shared with her girlfriends. He was a frequent dinner guest. Jill slipped a blindfold on him and whispered, "The party begins with a treasure hunt. Here is a bag with some things for you."
>
> Jack reached into Jill's bag. He felt a rubber band, a ball of yarn, and a cold metal object. He ran his finger along the metal object. It had a shaft and a metal loop at one end. The other end was bumpy. It was an antique style cabinet key. Perhaps he would need the key to open a cabinet door to reach the treasure. Jack remembered that there was old cabinet in Jill's dining room.
>
> Jack heard faint laughter and sounds from a fireplace. He heard the voice of Jill's friend, Jackie. Jackie had just returned from a skiing trip with a sprained ankle. She would not be able to walk very far.
>
> "Do come in," urged Jill "I am supposed to spin you around. Mind the rug."

Figure 6 illustrates Jack (drawn as a wooden figure) stepping into Jill's living room. Jack is blindfolded so he cannot see. From his previous visits Jack remembers the layout of the living room and dining room.

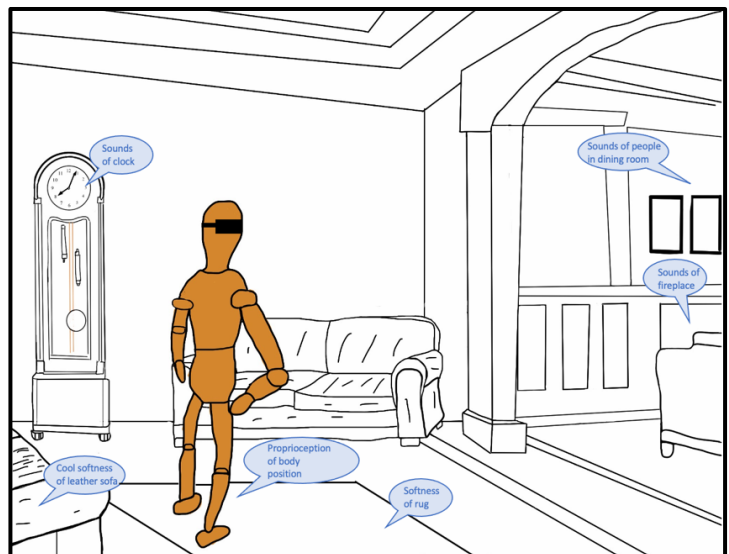

*Figure 6. In the blindfolded treasure hunt scenario, Jack (drawn here as a wooden figure) combines information from several senses and his memory to understand his current surroundings and make a plan.*

Like other people, Jack has body sense, that is, proprioception.[14] This sense arises from sensors on muscles as a person moves. It enables Jack to sense his movements and actions. Combining what he remembers from previous visits with information from his senses, Jack estimates his location, orientation and posture in the setting. As Jack steps into the living room, his left hand brushes a cool soft surface at a height above his knee. It is an arm rest of the leather sofa. He realizes that the sofa has been moved since his last visit. He feels a familiar softness of the floor as he steps onto a rug. He hears a ticking grandfather clock on his left. From the next room on his right, he hears giggling. Jack concludes that the birthday party guests are gathering in the dining room. He recognizes Jackie's voice and remembers that Jackie injured her ankle recently. Jackie would have difficulty if the treasure hunt involved walking on stairs to another location. Probably the treasure hunt will be in the dining room. Jack hypothesizes that the old cabinet key that he felt in the bag might open the built-in cabinet in the dining room.

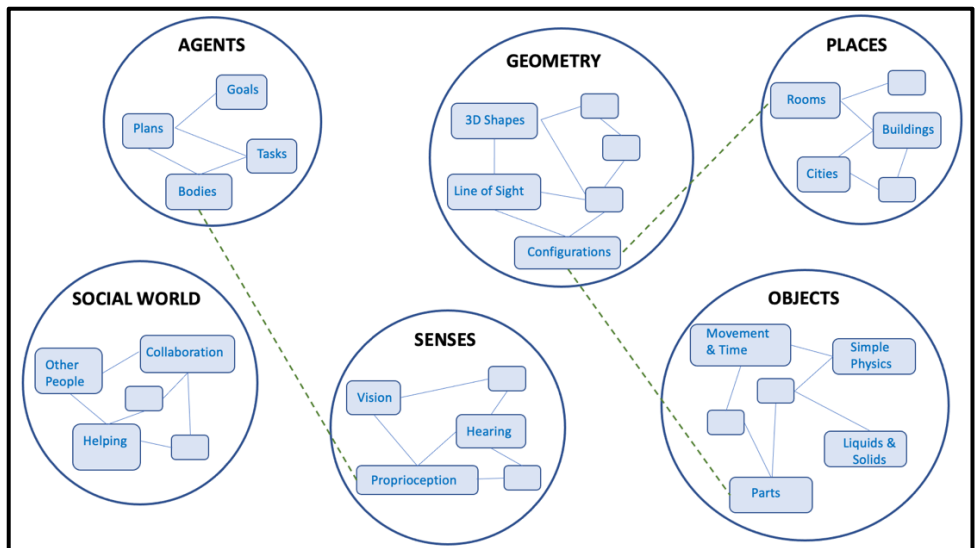

*Figure 7. The connections within models are dense and connections between models are sparse.*

Figure 7 shows notional groups of models for agents, kinds of planning situations, kinds of shapes, related groups of physics phenomena, stages of visual processing, and so on. The dashed lines suggest sparse interactions among the large ensemble models.

In the treasure hunt scenario, Jack imagines where the hidden treasure might be and plans how he could get it. He combines several sources of information, including:

- Touch information about the shape of a metal object in Jill's bag.
- Memory of the shape and feel of old cabinet keys.
- Memory of a built-in cabinet in the dining room of Jill's 1900s home.
- Audible sensing of sounds from the fireplace and party goers in the dining room.
- The searching activities of treasure hunts.

Jack's information comes from multiple senses and foundation models. He predicts what will happen if he takes different actions. He combines information from hearing and proprioception with his memory of the layout of the living room and dining room from (Places) and with information about goals and actions in treasure hunting (Agents). His classification of the key in Jill's bag is informed by his knowledge of the kinds of activities that are typical in a treasure hunt.

---

[14] Proprioception and kinesthesia are closely related terms. Proprioception describes awareness of posture and movement and also body position and awareness of nearby objects. Kinesthesia refers to a sense of joint position and movement. Our use of the term proprioception is intended to mean a body's sense of its configuration and movement (including kinesthesia) together with perception of its position in an environment.

### 3.3.2 The Dynamics of Multi-Model Information Fusion

*Multi-model* information fusion refers to the combining and fusing of information across multiple models. The special case of fusing information from multiple senses is called *multimodal* perception. Propagating information and constraints across multiple models reduces ambiguity in information fusion.

To convey a sense of the dynamics of multi-model information fusion, we return to the Blindfolded Treasure Hunt scenario to illustrate how Jack's understanding grows incrementally. At the beginning of the scenario, Jill tells Jack about the game, slips a blindfold on him, and hands him a bag of items. Moment by moment, Jack decides what to do in order to find and acquire the treasure. He draws on multiple loosely connected mental models (about Places, Objects, Agents, and so on) to answer questions:

- What should he do next to find and acquire the treasure?
- Where is he now? (How is he positioned in Jill's house?)
- What is the treasure?
- Where is it?

As Jack gets more information, he makes progress in answering these questions. He recognizes that he is in Jill's house. But where? He uses his perceptions to narrow the possibilities. He remembers that he stepped through the front door. This fact rules out being in most places in the house. He must be in the living room. But what direction is he facing? Jack hears sounds from a fireplace. He triangulates the location of the fireplace in the adjoining dining room. He recognizes the voices of people and identifies Jackie.

Without conscious effort, information for answering questions in one line of thought also narrows the possible answers for other lines of thought. Jack recalls that Jackie recently hurt her ankle. She cannot walk far. She will not be able to walk around the house. In the context of a treasure hunt, Jack realizes that Jackie is probably seated near the treasure. This suggests that the treasure is in the dining room. To recap, information from the sound of Jackie's voice combines with information about her state of health which combines with rules of treasure hunts which narrows possibilities for where Jack is and for where the treasure probably is.

For another example, the contents of the "hint bag" included an antique cabinet key. Unless someone put the key in the bag as a deliberate distraction, it is probably related to the location of the treasure. Jack remembers that there is an antique built-in cabinet in the dining room. When information reduces the set of possible answers for questions in one model, it can reduce the set of possible answers to questions in connected models. In computational terms, this amounts to propagating constraints from one model to another.

Neuroscience research provides insights into mechanisms that can support information fusion. Stein and Meredith review neural, perceptual, and behavioral evidence about how brains fuse multimodal sensory information (Stein and Meredith, 1993). They suggest that mammalian brain systems for information fusion are based on an ancient scheme of sensorimotor organization that predates modern species. Interactions among the senses occur at single cells and involve multiple areas of the brain. As they describe this in the context of childhood development,

> "Once one begins dealing with older infants and especially adults … there is little disagreement that *intersensory integration* is a fundamental characteristic of normal perception [emphasis added]." (Stein and Meredith, 1993)

Computational approaches for combining multi-model information are well-studied in Bayesian approaches to belief networks (Pearl, 2014) which uses network representations and probability distributions to model the fusion of information and beliefs. Ernst and Bülthoff present a Bayesian information processing view about how people can generate unambiguous interpretations of the world from streams of sensory information. As they put it,

> "To perceive the external environment our brain uses multiple sources of sensory information derived from several different modalities, including vision, touch and audition. All these different sources of information have to be efficiently merged to form a coherent and robust percept. … humans combine information following two general strategies: The first is to maximize information delivered from the different sensory modalities ('sensory combination'). The second strategy is to *reduce the variance in the sensory estimate* to increase its reliability ('sensory integration')." (Ernst and Bülthoff, 2004)

Combining information from multiple models is a long-established AI practice in hierarchical problem solving and planning (e.g., Manheim, 1966; Simon, 1969; Sacerdoti, 1974; Stefik, 1981a).[15] Hierarchical planning uses abstraction spaces over nearly independent subproblems and orchestrates the passing of information across abstract plans. This concept is ubiquitous in blackboard systems (Erman et al., 1980; Nii et al., 1982; Nii, 1986) where signal processing systems interconnect with symbol-based systems to analyze data over multiple levels of abstraction. It was explored in early research on spreading activation in semantic networks (Quillian, 1968).

"Spreading activation" is a widely used concept for modeling psychological phenomena in terms of activity on a network of interconnected nodes (e.g., (Anderson, 1983)). It refers to:

> "[the] … class of algorithms that propagate numerical values (activation levels) in a network for the purpose of selecting the nodes that are most closely related to the source of the activation." (Shrager et al., 1987)

In the following we draw on analyses and experimentation with computational models of spreading activation to gain insights into the dynamics of information fusion in multi-model approaches to cognition.

Across models of spreading activation, there are many variations in what information is in a node, what information (e.g., "strength") is in a link, and how the different network model parameters govern the process. For example, John Anderson's theory of human memory has a network model with few parameters. As he describes it,

> "[In] the ACT theory of factual memory, information is encoded in an all-or-none manner into cognitive units and the strength of these units increases with practice and decays with delay." (Anderson, 1983).

For computational experiments about network effects and spreading activation, Jeff Shrager, Tad Hoff, and Bernardo Huberman used a simple network model with three parameters:

> "Spreading activation networks typically consist of a set of potentially active nodes representing various states or items, variously interconnected by weighted links, and a local

---

[15] Marvin Manheim described hierarchical design for highway systems proceeding progressively from high-level design decisions to lower-level decisions of construction plans (Manheim, 1966). Herbert Simon (Simon, 1969, 1996) characterized this approach in terms of nearly independent subproblems. He described networks of dense connections within each of the nearly independent problems and a comparatively sparse network of constraints connecting them. Earl Sacerdoti formalized a version of the approach for robot planning problems (Sacerdoti, 1974). He used a logic-based representation of hierarchical abstraction spaces and emphasized least commitment planning rather than heuristic planning with backtracking. Stefik formalized a hierarchical object-oriented framework for nearly independent problems based on object representations for domain variables and constraints between values of variables (Stefik, 1981a).

relaxation rate at which the activity of an isolated node decays to zero. The dynamic behavior of these networks is controlled by three parameters. The first, specifying their topology, is the average number of links per node, $\mu$. The second, $\alpha$, is a positive number that describes the relative amount of activity that flows from a node to its neighbors per unit time. The third parameter is the relaxation rate $\gamma$, which can have a value between 0 and 1. In a typical application, some nodes (the sources) are activated by external inputs and these in turn cause others to become active with varying intensities." (Shrager et al., 1987)

In a related paper, Bernardo Huberman and Tadd Hogg start from results in statistical mechanics, to analyze spreading activation over random graphs (Huberman and Hogg, 1987). They studied search in large networks, considering research on semantic nets, spreading activation, and heuristic search in problem spaces. They predicted that AI systems and cognitive models will undergo sudden phase transitions when their connectivity increases beyond a critical value. This phenomenon is analogous to phase transitions in nature. Such phase transitions are predicted in highly-connected neural networks and are expected to appear in the behavior of dedicated AIs as they scale up to (say) millions of processors.

> "Statistical mechanics, based on the law of large numbers, has taught us that many universal and generic features of large systems can be quantitatively understood as approximations to the *average behavior* of infinite systems. Although such infinite models can be difficult to solve in detail, their overall qualitative features can be determined with surprising degrees of accuracy." (Huberman and Hogg, 1987)

Huberman and Hogg showed that the phenomena of computational phase transitions is analogous to physical phase transitions and is largely independent of local details. The term is familiar for describing how matter undergoes dramatic changes in its qualitative properties when certain parameters pass through specific values. For example, water turns to ice when its temperature drops to 0º C. They note that information systems can also have sudden and dramatic changes analogous to phase transitions in matter. In these situations where the number of degrees of freedom specified by constraints and the number of interesting parameters is much fewer than the actual number of freedoms in the systems.

> "[An] interesting phase transition is ... displayed by *percolation processes*. They can be easily visualized by considering a network of channels connecting sites and water flow between them. Given an initial concentration of fluid in a given site and a pressure gradient, a typical problem consists in determining how many sites, on the average, get wet after a long time. This percolation problem, which has been extensively studied in contexts ranging from the spread of an epidemic to the conductivity of electrical networks … has been shown to possess a phase transition independent of the detailed geometry of the system.
>
> When the *average number of paths connecting the sites* is below a certain value, the probability of finding a wet site at an arbitrarily large distance from the source is vanishingly small. This means that only a finite number of sites are connected to the source. As the number of paths increases, however, there is a singularity in the probability that distant sites get wet, and above a critical value the connected cluster becomes infinite. …" (Huberman and Hogg, 1987)

Jack's understanding of the game situation in the blindfolded treasure hunt scenario "crystallizes" because there are enough connections between his multiple models of the situation. He makes progress towards finding the treasure as possible values of situation variables in the models lock in.

Figure 8 shows how the average size of clusters of nodes increases as a function of the average branching factor in random graphs. A singularity and phase transition takes place at v=1. In the first regime where v<1, the cluster size is finite and increases with v. In the second regime where v>1, the cluster size is infinite.

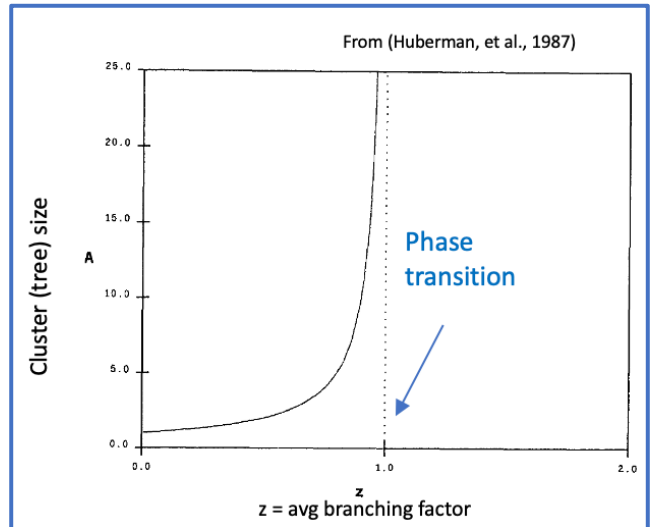

*Figure 8. The average size of node clusters in large random graphs as a function of the average branching factor. Adapted from (Huberman et al., 1987).*

In a related paper, Jeff Shrager, Tadd Hogg, and Bernardo Huberman created computational networks to test for singularity behavior experimentally (Shrager et al., 1987). Varying the network topology and activation parameters, they verified the phenomena predicted by the analysis.

How do the analytic and experimental findings about spreading activation relate to multi-model information fusion?

To understand the connection between spreading activation and information fusion, it may be helpful to recall Herbert Simon's description of "generation" and "testing" (or pruning) as dual processes in search, problem solving, and cognition. A similar distinction can be made between *spreading activation* and *spreading deactivation*. Spreading activation corresponds to generation and is about including variables and possible values in a search process. Testing or "deactivation" accounts for "ruling out" possible variable values across multiple models.

Models with large numbers of nodes and connections give detailed descriptions of a situation. In the Blindfolded Treasure Hunt, the models provide information answering the questions posed earlier. Here are questions related to adding possibilities in generation.

- What should Jack do next to find and acquire the treasure?
- Where is he now? (How is he positioned in Jill's house?)
- What is the treasure?
- Where is it?

Figure 8 shows how cluster size increases with the average number of links per node, $\mu$. This corresponds to the amount of detailed information that is under consideration for answering the questions. Restated, with the greater detail available when multiple models are combined, the answers that Jack needs are more likely to be resolved.

Testing or "deactivation" accounts for "ruling out" possible variable values across multiple models. Here are parallel questions related to removal of possibilities and "deactivation."

- What should Jack *not* do next?
- Where is Jack *not* located now?
- What is *not* the treasure?
- Where is the treasure *not* located?

Since all of the models refer to the same situation, derived limitations are additive. What is ruled out in one model can be ruled out in the others. This is the "spreading deactivation" part of the generate and test process.

Standard algorithms for spreading activation unify generating and testing under the concept of utility – favoring the options having the highest value. This unification of generating and testing brings into focus a required property about information fusion. Pseudo equation (1) says simply that the utility of the information elements increases when the information is fused.[16] Consider two variables A and B that appear in different models.

(1) $V(A + B) \supseteq V(A) + V(B)$           (Value of information additivity)

Given two elements of information A and B, the utility of the fused information is greater than or equal to the sum of their individual utilities. In the notation of equation (1), the function V maps an element of information to its utility. The notation for the "+" operator is used for both fusing the information elements and for adding their utilities.

Reflecting on the opening question from Tenenbaum et al., what is the "other information that makes the difference" when people appear to get so much from so little information?[17] The "other information" is what flows along the network that connects the multiple models. It has been hiding in plain sight.

---

[16] The combining of information here refers to the union of information in support of reasoning. More broadly in machine learning systems, representations can be "combined" and *generalized* not only in support of an immediate decision but also in support of broader and related future decisions. Statistical and experimental considerations for proposing and testing generalizations are beyond the scope of this section.

[17] In *How to Grow a Mind: Statistics, Structure, and Abstraction* (Tenenbaum et al., 2011), Tenenbaum and his colleagues asked several penetrating and related questions such as: "How does the mind get so much from so little? How does abstract knowledge guide learning and reasoning from sparse data? What forms does abstract knowledge take across different domains and tasks? How is abstract knowledge itself acquired?" The second part of the title and the last three questions show how the authors take a *structured representation* perspective in answering the first question. Abstract knowledge is encoded in a probabilistic generative model. The structure of that model guides inference in a probabilistic hypothesis space. A conclusion can "crystalize" from such processes when it has a relatively high probability.

# 4. A Trajectory for Developmental Bootstrapping

Why do people actively seek to understand the world?

Robert White raised this question in *Psychological Review* (White, 1959) as a challenge to existing scientific theories of motivation. The dominant theories at the time were Hull's theory of drive reduction and Freud's instinct theory, which White saw as very similar. Citing growing discontent with those theories of motivation, he suggested that a revolution was taking shape in research on animal behavior and in psychoanalytic psychology. He saw increasing interest in behaviors that the existing theories failed to explain.[18]

The unexplained behaviors included a tendency to explore the environment, a tendency to manipulate the environment, activities directed to learn about the environment, and others. He proposed that learning organisms possess a "competency drive" to have effective interactions with the environment.

> "In organisms capable of but little learning, this capacity might be considered an innate attribute, but in the mammals and especially man, with their highly plastic nervous systems, fitness to interact with the environment is slowly attained through prolonged feats of learning. We need a different kind of motivational idea to account fully for the fact that man and the higher mammals *develop a competence* in dealing with the environment which they certainly do not have at birth and certainly do not arrive at simply through maturation." (White, 1959)

Ongoing research on learning and curiosity of children and self-directed AI systems share White's perspective. Figure 9 shows a notional high-level model for an agent interacting with the environment and acquiring competences. Perceptions, physical actions, and mental actions take place in parallel. The pattern "Observe-Act (Repeat)" is a loop where actions can be physical or mental. Larger patterns are loops of experimentation in the environment saying broadly "Do something, see what happens, and learn from it. (Repeat)." [19]

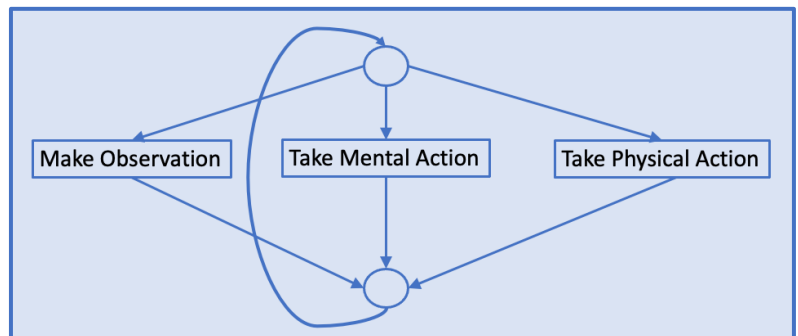

*Figure 9. Notional high-level "Observe-Act" cycle for an embodied agent.*

---

[18] Three years after White's paper, Thomas Kuhn popularized the term "paradigm shift" to describe the situation when the foundations of a scientific field shift from older orthodox theories to a new theory that better explains cases not adequately covered by the old theory (Kuhn, 1959). In retrospect, White's paper was published on the cusp of such a paradigm shift.

[19] Broadly summarizing patterns in the physical organization of the human nervous system and its cognitive processes, Jeff Hawkins suggests that higher levels of cognition are supported by similar patterns in the cortical regions and neocortex as much earlier and simpler low-level patterns of sensorimotor control (Hawkins, 2021). Roughly, the older parts of the brain have an Observe-Act loop that does not include much learning. Evolution has led to more recent brain layers with Observe-Predict-Learn" cycles that "advise" older parts of the brain and enable more nuanced control of actions.

Restated, an intelligent organism does not just sit there and do nothing. Lack of action in the wild would result in going hungry or being eaten. An organism would learn little from observing nothing and doing nothing By observing, moving, and experimenting, it can learn about the world and expand its ability to interact with it.

The foundational competences required for creating embodied AIs include perception, motor skills, discovering abstractions, planning, modeling other agents, social awareness, and communication in natural language. The competences as described in the following sections are similar to core foundational competences as described by Elizabeth Spelke (Spelke, 2022). Analogous competences arise in other species. Our thesis is that having analogous innate competences and learning processes will enable AIs to develop analogous competences for intelligence and human compatibility.

### *4.1 Bootstrapping Perception, Understanding, and Manipulation of Objects*

In his early studies of childhood development, Jean Piaget (1936, 1954) investigated how children develop mental models of the world including their own bodies, senses, and motor control. Piaget's observations and findings were limited by the instrumentation and cognitive theories of his time, but his sensorimotor framework of child development continues to inspire and inform research. He observed how children fused information from multiple senses as they moved about, learned how the world works, and manipulated objects. By the time people become adults, they have rich competences for perceiving and understanding the world, including models of its 3D structure.

#### *4.1.1 Research on Sensorimotor Skills*

Decades after Piaget, Elizabeth Spelke (Spelke, 1990) studied infant's object perception in their first months. She found that infants from two to three months recognize objects by dividing their visual field (represented as a perceptual array) into units that move as connected wholes, move separately from one another, maintain their size and shape as they move, and change each other's motion only when they touch each other.

> "All the experiments provided evidence that young infants perceive object boundaries by detecting surface motions and surface arrangements. Infants perceived two objects as separate units when one object moved relative to the other object, even when the objects touched throughout the motion … Infants also perceived two stationary objects as separate units when the objects were spatially separated on any dimension, including separation in depth. … Our research suggests that the processes by which humans apprehend objects occur relatively late in visual analysis, after the recovery of [this] information for three-dimensional surface arrangements and motions." (Spelke, 1990)

Spelke's findings characterized how newborns recognize objects in their visual field. When objects move behind each other, early infant vision systems model a limited sense of depth. Young infants create simple 3D models of space and imprecise notions of distance. As their bodies and brains develop, children's initial primitive visual systems mature into more capable ones. Kopp summarizes this development of infant vision in *Baby Steps:*

> "At birth, [a newborn's visual system] is far from fully functional because of the interrelated immaturity of the eyes (e.g., the retina), the visual portion of the brain, and the ability to perceive certain stimuli (e.g., a very narrow band of white and black stripes). … newborns differ appreciably from older babies, toddlers, or adults in the quality and quantity of what they take in through their senses." (Kopp, 2013)

In their book *The Merging of the Senses*, Barry Stein and M. Alex Meredith reviewed research findings from neural, perceptual, and behavior studies of humans and animals (Stein and Meredith, 1993). They focused on the signals and structures in human and other mammalian brains. They found that activations from different senses affect the same multisensory neurons. Those neurons are organized in overlapping and *integrated sensorimotor maps*. As they summarize it,

> "A stimulus that is seen, heard, or felt can activate the same neurons in the superior colliculus and these, in turn, can produce a coordinated series of premotor signals that ultimately initiate overt behavior." (Stein and Meredith, 1993)

The brain areas of monkeys and people include maps of body parts such as a hand or a foot. Somatotopy is the point-for-point correspondence of an area of the body to a specific point in the central nervous system. Information from the senses is received in the somatosensory cortex. The map areas for sensing and motor control are interconnected. Restated, the *brain combines data from multiple senses* and feeds the information into cells corresponding to the same body locations. These cells also connect to corresponding nerves for controlling deliberate motion.

Consider the following scenario of a young child starting to explore the world.

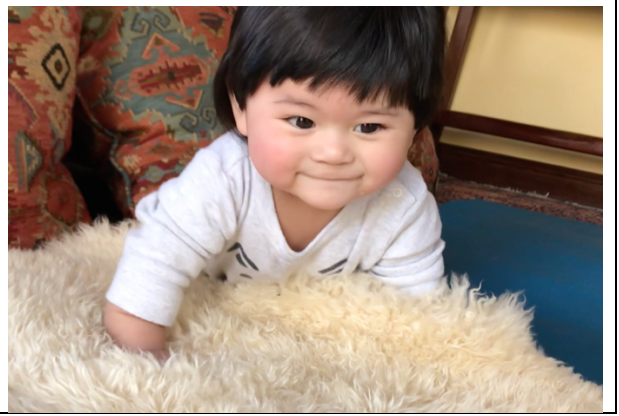

**Learning to Crawl**

Macer's grandma sets him on a sheepskin. He is surrounded by pillows, a sofa, a table, and other things.

Macer holds up his head, forms his hands into fists, kicks, and bounces his feet. He watches things move in the world. Grandma beckons and talks to him. "Come here, little Macer," she coos. "That's a good boy!"

As Macer moves, he learns relations between what he feels, what moves in his visual field, and what happens as he moves. His nervous system predicts what he will feel or see next as he takes action. A little at a time, Macer builds a model about his body and the world.

The scenario describes Macer learning to crawl and explore the world. He begins to understand his own body, the surrounding objects, and his motor capabilities.

A hypothesis is that the *innate prewiring in the sensorimotor maps of an infant's neurons guides the infant's attentio*n as it notices and learns the arrangement and motions of parts of its body. As described by Stein and Meredith, the cells in the sensorimotor map that (1) connect to the sensors in the fingers also (2) connect to the neurons in the motor map that (3) connect to the muscles in the fingers. These connections in neuron wiring result in nearly simultaneous signaling by sensors for external touch, internal proprioceptive sensing, and controlled moving of fingers. When an infant sees its hands move, the simultaneous signaling may guide the fusing of visual information into its body model and world model. Restated, the prewiring guides learning. It guides the infant in learning how to move and how to fuse information across multiple models of the world.

Human adults routinely identify objects using vision and touch. They use the senses both separately and simultaneously. Until recently, little was known about how they develop the ability to *combine* information from multiple modalities. Giulia Purpura and colleagues tracked the course of children's performance on object recognition with visual information alone, haptic (touch) information alone, and combined visual and touch information (Purpura et al., 2017). Starting at four months children can combine vision and touch senses to recognize objects. Combining multimodal sense perceptions improves cognition and performance. Their ability to combine perceptions gets better over time starting in their first year.

*4.1.2 AI Systems that Perceive, Understand, and Manipulate Objects*

Children's ability to reach for objects and investigate them is preceded by other developments. Newborn infants flail their arms in uncontrolled manners. Occasionally an arm will bump into a toy or some other object and cause it to move. This uncontrolled movement is called *motor babbling*.

Over a few months toddlers learn to reach towards target objects. When something brushes the palm of an infant's hand, its fingers close involuntarily in a grasping gesture known as the palmar reflex. Combining these reflexes, an infant's hand can close on a small object when a moving hand touches it. The combined reflexes demonstrate how a roaming hand can grasp and move an object. The child's perception of these events bootstraps further developments in reaching, grasping, and increased exploration of the environment. Over time, the early reflex actions fade and are superseded by actions performed with skill and intention. Kopp summarizes the transition for children between four and seven months of age.

> "Improvements in the baby's visual perception make many of these cognitive and social developments possible. A particularly important advance is that visual acuity now moves more closely to adult levels. Now babies can actually see quite clearly, and they find many things in their surroundings they want to examine more closely. Better sight helps guide reach so that, in fact, a baby can take hold of and investigate objects that are of interest." (Kopp, 2013)

Inspired by findings about childhood development processes in the context of bootstrapping AIs, Jonathan Juett and Benjamin Kuipers created a prototype AI that learns like children to move its arm, reach towards objects, and grasp them (Juett and Kuipers, 2019). The authors created a computational learning model and robot arm that has robotic analogs of children's innate reflexes and uses them to bootstrap the AI's learning process for grasping and moving objects. Like children, the AI starts by observing the action of its arm as caused by (programmed) innate reflexes for flailing and closing its hand when it touches something. In short, the AI learns like a child to grasp and move objects.

In another example Luis Pilato and colleagues created an AI that learned a diverse set of physics concepts (Pilato et al., 2022). They used a simulation world, a dataset of curated videos of objects participating in physics events, a deep learning approach, and a VOE approach for probing the extent and generality of the concepts that the AI learned. The VOE test concepts included object permanence – that objects do not suddenly disappear. Another test concept was "solidity" – how objects cause other objects to move when they bump into them. Another test was "continuity" – that objects do not suddenly "teleport" from one location to another. Their PLATO AI system has a perceptual subsystem that learns concepts from the videos as a set of object codes encoding conditions and actions. Its prediction subsystem predicts what will happen next in a video. When there is a difference between what the AI predicts and what actually happens next, the system logs a VOE event and records the differences between what was expected and what was seen. The curated videos function as training sequences and the VOE experiments probe the generality of what the AI has learned. Related AI research in building physics models from visual data includes (Ebert et al., 2019; Smith et al., 2019).

Returning to multimodal perception, David Miralles and his colleagues demonstrated an artificial cognitive system that combines touch and vision on an object recognition task (Miralles et al., 2022). Combining vision and touch improved its performance and enabled it to adjust for degradations of its vision system. Restated, their machine learning system performed better when it was trained on the combination of vision and touch data. Related AI research on multimodal competences includes (Falco et al., 2020; Lin et al., 2019; Sundaram et al., 2019).

In an invited talk[20] to AAAI-23, Sami Haddadin described advances in modeling, sensors, modular appendages, learning, and robotics that enable robots to have more robust models of their bodies and tactile senses. His research at the Technical University of Munich includes contact sensors for "robot skin," strain sensing proprioceptive sensors for joints, proprioceptive models of sensors and body parts, new robotic fingertips, tactile reflexes, improved modular robot part design and improved modular parts including hands and arms. The advances couple research on building robots with research on learning proprioceptive models of their bodies and their body's interactions with the environment.

The embodied AI research mimics human development, learns multiple competences at once, starting from an egocentric perspective. As Jiafei Duan and colleagues said in their research survey[21] of embodied AI,

> "There has been an emerging paradigm shift from the era of "internet AI" to "embodied AI", where AI algorithms and agents … learn through interactions with their environments from an egocentric perception similar to humans." (Duan et al., 2022)

The examples in this section focus on the first few months when children begin to develop multimodal models for perceiving, understanding, and manipulating objects in their environments. By eighteen months children have richer models of how objects interact. They learn that objects can have external moving parts and also internal mechanisms. As Alison Gopnik and her colleagues write:

> "Children continue to learn about causal relations among objects throughout their toddler years. Before they are three, children are already giving appropriate explanations about what caused what. They say things like 'The bench wiggles because these [bolts] are loose' or 'The nail broke because it got bent.' By three or four, they can make quite explicit predictions about how simple mechanical systems will work. For instance, you can show them a sort of Rube Goldberg apparatus of pipes and tubes through which a ball rolls. Three-year-olds can predict that the ball will have to travel a certain distance before it can bump into another ball and make the machine move." (Gopnik, 1999)

In summary, neuroscience and developmental psychology provide important insights about how perception and control are managed and learned in living systems. In human and animal development, innate sensorimotor structures in the brain guide learning about how to perceive the body and the environment and how to control the body in the environment.

Developmental AI agents perceive their environments and move in them. They fuse multimodal streams of perceptual data and reinforce what they learn across multiple models as they learn about how the world works. They perceive what they are touching, the position of their body, and the world around them. They build mental models of the world, including the 3D arrangement of objects and the naïve physics of how things fall. They use the mental models to predict what will happen next. When the predictions differ from what happens, the differences are noted and used to inform agent learning. Researchers use VOE probes to assess the correctness and generality of the developing AI models.

Advances in the bootstrapping of perceiving, understanding, and manipulating objects prompts the following hypothesis:

---

[20] A video showing some of this research is available (Haddadin, 2022).

[21] Duan and colleagues also review the capabilities, limitations, and challenges for training AIs using current simulation and virtual reality (VR) systems. Limitations of these systems affect the "sim-to-real" transition, that is, how well the AIs can function in the real world when their training is in simulated worlds.

- *Sensorimotor discovery hypothesis.* Sensorimotor discovery is where AI agents learn about their embodiments and their environments. Training involves observing and correlating sensory percepts of sight, sound, touch, proprioception, and motor (muscle) control. The AIs build predictive models of how percepts change when the agents are moved and when they exercise motor actions. In the cognitive architecture, the learning system is separate from lower level perception and motor control system. Learning is unconscious, incorporates high-level multi-dimensional data, does not do motor control, and informs the motor execution system with learned possible action sequences.

The next section explores challenges and advances in modeling and bootstrapping cognitive competences that support goals, multi-step actions, and abstraction discovery to support higher levels of cognition for more complex perception and action.

## *4.2 Bootstrapping Goals, Multi-step Actions, and Abstraction Discovery*

The following "Filling a Cup with Water" scenario is representative of what a three- or four-year-old human child can do.[22]

> **Filling a Cup with Water**
>
> Macer sat at the kitchen table, playing with a toy car. His mom was at the stove preparing lunch. She turned to him and asked, "Do you want some water with lunch?"
>
> Macer nodded. He had watched his parents get water from a spigot in the refrigerator on previous occasions. He picked up the plastic cup at his place at the table and carried it to the refrigerator. Opening the refrigerator door, he glanced at the water spigot on the left side. He stretched out his arm but could not quite reach the oval gray button above the spigot.
>
> Macer looked around the kitchen. A blue footstool was on the floor next to the refrigerator. Macer put his cup on the footstool and pushed it to a spot below the refrigerator door. Stepping up on the stool, he reached again for the water spigot. He pushed its gray button with the fingers of one hand. With his other hand he held his cup steady below the spigot.

Macer's actions make sense to us because we know what Macer wants to do and we are familiar with kitchens. We infer that Macer picks up the cup and carries it to the refrigerator because he plans to fill his cup with water from its spigot. When Macer cannot quite reach the spigot and then looks around, we infer that he is looking for something to stand on. Our understanding of the scenario requires that we know both how the world works (e.g., pushing a button on the spigot turns on the water) and how Macer's step-by-step actions can meet his water drinking goal (e.g., moving the blue footstool so that he can step on it to reach the water spigot).

The term *teleology* refers to theories and approaches that explain the behavior of agents according to their goals and purposes.[23] Models of mental state are studied in psychology, sociobiology, and related fields. For example, Michael Tomasello and others compare the ways that children and animals such as great apes behave socially, cooperate, and have goals (e.g., Tomasello, 2019; Tomasello, 2009; Carey, 2004; Carey, 2009). This research looks back over 5 to 6 million years and traces the physical, social, and cultural changes of early humans. Among other things, during this time period there were brain and behavioral changes in humans, lengthening of the time of early child development, changing the culture of parenting, and increasing the use of tools.

---

[22] The scenario is based on observations of a four year old in a family kitchen.

[23] Related terms include "folk psychology" and "theory of mind" (ToM). J.W. Astington et al. defines the area of research as follows: "Theory-of-mind research investigates children's understanding of people as mental beings, who have beliefs, desires, emotions, and intentions, and whose actions and interactions can be interpreted and explained by taking account of these mental states." (Astington et al., 2008) Alison Gopnik and Andrew Meltzoff explore variations on these ideas in their book *Words, Thoughts and Theories* where they provide historical context and use the term "Theory Theory" to refer to a family of theories about attributing mental state. (Gopnik & Meltzoff, 1997).

What goals do agents have? Do they have hierarchical goals and multi-part activities? These questions have been asked about children at different ages, about primates, about human organizations, about human teams, and about colonies of insects. These questions now arise in developmental bootstrapping of AI systems.

*4.2.1 Early AI Systems with Goals, Multi-Step Actions, and Abstractions*

Explicit representation of agent goals has long been part of AI and cognitive psychology. Early AI research pioneered symbolic cognitive models and representing problems in a state space and search framework. It established terminology and technical approaches for modeling *actions, prerequisites, goals, states, plans, beliefs, and abstractions.*[24] It modeled problem solving methods and studied their computational complexity.

For example, means-ends analysis is an early problem solving framework where an agent has a current state and a target or goal state. The agent considers the difference between its current state and a goal state. The "difference" can be cast, for example, as a distance to be reduced, obstacles to be removed, a path to be found, or properties to be changed. A solution is a sequence of agent actions that reduces the difference and enables the agent to reach the goal state. In this way, early AI systems addressed problem solving situations similar to the cup filling and block tower building scenarios. In early systems the algorithm or model for going from differences to actions was designed manually.

Early symbolic AI systems did not see or move real physical objects. They did not fill water cups in the real world. Models of human-like problem solving were developed after taking and analyzing verbal protocols of experts solving a problem. AI systems were tuned by expert curation of sets of rules and other representations. This symbolic AI research modeled cognition and tasks at a high level. Overall, it involved manually developing models of expertise following information gleaned from verbal protocols.

A cognitive limitation of early AI problem-solving models (including both planning systems and RL) was that they did not reason with abstractions. Abstract plans preserve the important properties of states and search but omit many details. Simon recognized and described this combinatorial advantage in *Sciences of the Artificial* (Simon, 1969, 1996). He explained how abstractions enable exponential savings by creating big steps or "planning islands" towards abstract goals. Abstract goals describe simplified future states with the expectation that the details of achieving them will be worked out as needed. Subsequent hierarchical planning systems reasoned with abstractions,[25] although their *abstractions were still designed by researchers rather than discovered by the AI systems.*

In classical AI, an agent's goals are represented as desired states. In hierarchical planning systems, goals are organized in hierarchies where the more abstract goals are higher in an abstraction hierarchy. There are different approaches to executing and reasoning about goals and plans. In some approaches, a planning process is initiated by putting a goal on a queue of a planning program.

---

[24] Early AI research includes Newell, Shaw, and Simon's early RAND report on a general problem solver (GPS). This report introduced many concepts of symbolic AI (Newell, et al., 1959). A decade later, Fikes and Nilsson reframed symbolic problem solving in terms of theorem proving in predicate calculus (Fikes et al., 1971).

[25] Building on Richard Fikes' and Nils Nilsson's STRIPS system, Earl Sacerdoti's ABSTRIPS system demonstrated planning islands in hierarchical planning. It augmented predicate calculus representations with annotations for abstraction levels. Mark Stefik represented hierarchical planning in an object-based framework and demonstrated the exponential reduction of the problem space via constraint-based model reinforcement (Stefik, 1981a). For an overview and history of some early problem-solving systems, see AI textbooks such as (Russell and Norvig, 2021).

Other approaches reason over a plan with abstract steps, actions, and goals and decide which steps to refine. This decision process – with meta-goals, meta-actions, and meta-states – can be organized in a metacognition framework for *planning about planning* or *reasoning about reasoning*.[26]

Reinforcement learning (RL) guides agents in choosing actions that lead to desirable goal states.[27] As Richard Sutton and Andrew Barto put it in their introduction to *Reinforcement Learning*,

> "When an infant plays, waves its arms, or looks about, it has no explicit teacher, but it does have a direct sensorimotor connection to its environment. Exercising this connection produces a wealth of information about cause and effect, about the consequences of actions, and about what to do in order to achieve goals. Throughout our lives, such interactions are undoubtedly a major source of knowledge about our environment and ourselves. Whether we are learning to drive a car or to hold a conversation, we are acutely aware of how our environment responds to what we do, and we seek to influence what happens through our behavior. Learning from interaction is a foundational idea underlying nearly all theories of learning and intelligence." (Sutton and Barto, 2018)

Using goals, actions, and states is increasingly important when plans involve many steps and have long time horizons. AI researchers have proposed modifications to RL to enable it to reason hierarchically (HRL). A much studied approach in HRL uses an option model, which specifies initial states, terminal states, and other parameters. In this approach, an agent can plan with options rather than with primitive actions – essentially taking great leaps in a plan. In another variation of RL, Thomas Dietterich proposed an approach where people provide subgoals for hierarchical policies (Dietterich, 2000). In 2003 Andrew Barto and Sridhar Mahadevan wrote an overview of the state of the art of HRL (Barto and Mahadevan, 2003). More recently, Yannis Flet-Berliac surveyed HRL approaches (Flet-Berliac, 2019). Like the early hierarchical planning systems, these HRL approaches do not *discover* abstractions.

### 4.2.2 Research about Human Multi-step Actions and Abstraction Discovery

Parents know that infants gradually begin to act with purpose and increasing skill. A key part of acting with purpose is the ability to perform a *multiple-step sequence* of actions in similar situations. Infants spend time exploring and learning what they can do. Grouping together a sequence of actions for reuse is called *action chunking* and is part of *abstraction discovery*.

Newborn humans exhibit innate multi-step sequences in their feeding reflexes. For example, the *rooting reflex* happens when a corner of an infant's mouth is touched or stroked. The infant turns its head, opens its mouth, and moves (or "roots") in the direction of stroking to start feeding. Precocial offspring of many grazing species carry out multi-step sequences of actions such as walking to avoid predators.

---

[26] At about the same time, Robert Wilensky and Mark Stefik developed AI systems for meta-planning, enabling an AI to represent and reason about plans and its planning (Wilensky, 1981; Stefik, 1981b) Wilensky developed meta-planning in the context of an AI system that reasoned about explicit representations of plans as they are used in stories and narrative, as well as reasoning about the planning process. Stefik focused on meta-planning for representing and switching between deductive and heuristic processes for solving under-constrained subproblems in experiment planning in molecular genetics.

[27] Sutton and Barto's book (Sutton and Barto, 1992, 2018) is the classic presentation of RL. RL is introduced in AI textbooks such as (Russel and Norvig, 2021). Describing it in detail is beyond the scope of this paper. A brief comparison with planning may be helpful. In classic AI planning research, a planner has a model of the environment including a set of possible actions. It employs a search method to find a good or optimal sequence of an actions (the "plan") for an agent to reach a desired goal from a given initial state. To start from a different initial state, replanning is needed. Reinforcement learning starts without a model of the environment. Given a set of possible actions, it interacts with the environment to learn a model (the "policy") that describes optimal next steps for reaching a goal. RL generally requires much more computation than planning *because it develops a policy that works for all states*. In short, planning uses a given model to reach a goal from a given state, whereas RL learns a general policy to reach a goal from any state.

Consider the task of building a tower of toy blocks. Newborn infants cannot build block towers in part because they cannot lift or hold blocks. At 18 months they make towers with two or three blocks. By two years they make towers with several blocks. Placing one block on top of another sets the stage for placing another block. In adults, the learning of multi-step sequences is ubiquitous.

Wendy Wood and Dennis Rünger reviewed computational and neurobiological research[28] on the learning and triggering of multi-step sequences especially in the context of habitual behaviors (Wood and Rünger, 2016). The term "habit" refers to the behaviors that are taken when particular situations arise. The situation is called a trigger. When a trigger appears, a predetermined sequence of actions immediately follows. In behavioral studies, a grouping of memory items is called a "chunk" and creating a set of actions is called "action chunking."

| Trigger | Action |
|---|---|
| A person arrives at the bathroom sink after breakfast. | He gets out the dental floss, toothpaste, and toothbrush. He flosses. Then he puts toothpaste on the toothbrush and brushes his teeth. |
| A pitcher notices a baseball flying at high speed towards his head. | He moves his head out of the way and raises his gloved hand to catch the ball. |
| A driver settles in the driver's seat of his car. | He glances around the seat, checks the mirrors, makes sure the shift is in "park," steps on the brake, and pushes the start button. |
| A woodsman hears rustling and a growl from behind a nearby bush. | He checks his surrounding and looks for the animal. He backs away slowly. |

*Figure 10. Examples of triggers and multi-step responses in human habits.*

B. J. Fogg, who teaches people how to create good habits and defeat bad ones, teaches that the most common human triggers for habits are location, time, emotional state, other people, or an immediately preceding action (Fogg, 2019). Figure 10 gives examples of triggers and multi-step responses for human habits.

In his book *Thinking: Fast and Slow*, Daniel Kahneman explains the fitness advantage of multi-step action sequences (Kahneman, 2011).

> "By shaving a few hundredths of a second from the time needed to detect a predator, this circuit improves the animal's odds of living long enough to reproduce. … Of course, we and our animal cousins are [also] quickly alerted to signs of opportunities to mate or to feed …" (Kahneman, 2011)

Ann Graybiel studied how structures in the basal ganglia chunk learning action sequences by recoding information from the neocortex, which is the locus of higher-level thinking and problem solving (Graybiel, 1998). In other words, *the basal ganglia recode highly processed information from other parts of the brain to create efficient multi-step action sequences* associated with a trigger. As Graybiel found,

---

[28] Wood and Rünger's review summarizes psychological theories starting with behaviorist theories of Thorndike (Thorndike, 1898) through more modern cognitive theories. For example, Tolman's early work *Cognitive Maps in Rats and Men* found that rat and human brains employ a map-like structure in associating situations with multi-step action responses (Tolman, 1898).

> "… this recoding within the striatum can chunk the representations of motor and cognitive action sequences so that they can be implemented as performance units. This … provides a mechanism for the acquisition and the expression of action repertoires ..." (Graybiel, 1998)

Mechanisms for caching action sequences predate the human species. In humans the learning process operates below conscious awareness. Compared to the rapid stimulus-response performance enabled by the recoding, the *learning process for chunking is slow*. The mechanisms need to decide which multi-step sequences should become habits and how general the sequences and triggers should be.

Further research by Smith and Gabriel describes signaling between the neocortex and ganglia in the learning phase (Smith and Gabriel, 2013). The coordinated signaling suggests that the discovery of abstractions involves higher functions of cognition and mechanisms for organizing memory.

Behavioral and brain activity studies using fMRI data have shown that people plan navigation tasks hierarchically (e.g., Balaguer, 2016). To date, however, the data from fMRI and other scanning techniques support only a high level understanding. Computational models and experiments are now beginning to provide a means for deeper understanding.

*4.2.3 Computational Models of Abstraction Discovery*

Computational models of chunking are a staple of computational psychology and AI. For example, John Laird, Paul Rosenbloom, and Allen Newell created a computational model of learning and action chunking in their SOAR system (Laird, et al., 1984). The model determines a trigger and the actions in the sequence. To represent and generalize the range of situations and actions beyond a specific example, the model abstracts the trigger conditions and the action responses. Research continues for how this works in human cognition and how it should be done in open world and incremental learning approaches by AIs.

Adults think routinely in terms of hierarchical abstractions. Asked how to travel from (say) a town in northern California to a hotel in Manhattan, an adult may suggest using ground transportation to reach San Francisco airport (SFO), then taking a flight to one of the airports near New York City, and then taking ground transportation to the hotel. That description divides the travel into three abstract steps and defers detailed planning of actions until later. People create plans for many kinds of problems. They plan quickly in big steps, confident that the in-between steps can be determined as needed.

Without prior knowledge a person could not immediately and skillfully divide a problem into useful subproblems (e.g., the legs of the trip to New York). People find and employ useful abstractions. Babies spend weeks playing with blocks and incrementally learning to build towers. As adults travel, they become familiar with modes of travel and develop mental maps of the areas where they have traveled. Hierarchical planning involves abstract states (e.g., locations on maps), abstract actions (e.g., "travel" and "stack block"), and abstract goals (e.g., "being at a hotel" or "having a tower of blocks").

Hierarchical planning per se has long been part of AI. But how does hierarchical planning relate to action chunking and abstraction discovery? How does an infant or AI create effective goals before it knows very much about its world?

In their paper *How to Grow a Mind: Statistics, Structure, and Abstraction*, Joshua Tenenbaum and his colleagues take a structural approach to modeling how humans discover useful abstractions. The authors draw on famous scientific examples. For example, Linnaeus identified a hierarchical organization for biological species, and Mendeleev identified the periodic structure of the chemical elements as represented in a table. In cognitive experiments with subjects, this research measures the fit of the data to candidate structures for representing the abstractions. In the Tenenbaum and Kemp papers, the best fit of behavioral data to structures is determined using search and optimizing algorithms.

Tenenbaum and his colleagues employed a generative model capable of comparing and optimizing across structural variations beyond the hierarchies typically associated with abstraction in problem solving. It selects structures to organize the data into a containing form. The space of possible forms is described using graphs and grammars. The structures include partitions, chains, ordered lists, rings, limited hierarchies, grids, and cylinders. In an earlier paper, Charles Kemp et al. described the first challenge of the problem as determining the best form for organizing the data and representing its relations (Kemp et al., 2008).

> "The higher-level problem is to *discover the form* of the underlying structure. The entities may be organized into a tree, a ring, a dimensional order, a set of clusters, or some other kind of configuration, and a learner must infer which of these forms is best. Given a commitment to one of these structural forms, the lower-level problem is to identify the instance of this form that best explains the available data [italics added]." (Kemp, et al., 2008).

Focusing on hierarchical structures, Alec Solway and his colleagues asked how people search for useful hierarchical abstractions (Solway et al., 2014). As they put it,

> "Not all hierarchies are created equal. The wrong hierarchical representation can actually undermine adaptive behavior. … It matters very much which specific set of hierarchical representation an agent carries. … While hierarchy can facilitate learning, it also introduces a new learning problem, the problem of *discovering* beneficial rather than disruptive subtask representations." (Solway et al., 2014)

Solway et al. developed a model of the costs of planning with hierarchies. Like Tenenbaum and his colleagues, they framed the problem in terms of Bayesian model selection. They conducted a series of behavioral experiments where subjects were faced with a set of mostly navigation tasks[29] in randomized order. The findings included that people quickly found optimal solutions and discovered task hierarchies spontaneously. How do people do this?

In the study, patterns of brain activity correlated with human behavioral patterns. The levels of brain activity scaled with the costs of hierarchical planning. The study found that the subjects first learned a simple topological and metric model of the world. Then they created a plan structure similar to the structure of the world model and organized their goals and behavior according to that structure. Restated, people planning navigation routes followed external spatial patterns found in the environment to shape their internal abstractions for actions.[30]

How can abstract goals and hierarchical abstractions be discovered for problems *other than spatial navigation problems*? Building on and extending the Solway et al. research, Momchil Tomov and colleagues created and experimented with a computational Bayesian process model for discovering abstract actions and states (Tomov et al., 2020). Their model makes detailed predictions for how the discovery of hierarchical abstractions depends on the topological structure of the search space, the distribution of rewards, and the distribution of tasks in the environment.

---

[29] One non-navigational tasks was solving a Tower of Hanoi puzzle. Although this is not a navigational task, it involves planning a sequence of actions.

[30] Simon's "parable of the ant" explored the relation between the constraints of an environment and the path of an agent. He described an ant walking along a complex path on a beach (Simon, 1996). He suggested that the complexity of the ant's path did not necessarily imply that the ant had a complex reasoning system. The beach sand had a complex contour, shaped by the actions of beach walkers and the wind. He concluded that the ant employed a simple low energy algorithm and that its path resulted from its interaction with the complex contour of the beach surface. Although this example shows the structure of environment shaping the path of the ant, it is *not an example of ants learning* from the structure of the environment. For example, the ant would not "plan" different future paths based on its traversal of the beach path.

Their computational model employs an *online planner* and an *off-line hierarchy discovery component* that incrementally builds a cluster-based representation of task performance in the environment. The planner uses a hierarchical best-first search, which operates with increasing efficiency as the hierarchical model of the environment is built incrementally. As they describe it,

> "The key idea is that the environment is assumed to have a *hidden hierarchical structure* that is not directly observable, which in turn constrains the observations the agent can experience. The agent can then infer this hidden hierarchical structure based on its observations and use it to plan efficiently. Assuming that some hierarchies are a priori more likely than others, this corresponds to a generative model for environments with hierarchical structure, which the agent can invert to uncover the underlying hierarchy based on its experiences in the environment." [emphasis added] (Tomov et. al., 2020)

The online planner requires relatively limited computational resources to solve multiple problems. The hierarchy discovery process learns portions of the hierarchical abstraction that are then used by the planner. The abstraction discovery component builds abstractions from *clusters of action steps taken by the planner*.[31] It favors small clusters and a small number of clusters, dense connectivity within clusters, and sparse connectivity between clusters (not counting the cluster-to-cluster link).

Tomov and his colleagues were interested in phenomena reported by previous researchers in their human behavior studies including detection of bottleneck transitions corresponding to states in different clusters, purposeful hierarchical planning corresponding to the hierarchical model, preference for hierarchies with the fewest clusters, and others. Their simulation experiments showed that their system reproduced human phenomena that were previously reported.

The researchers were also interested in novel phenomena that had not been tested previously with humans. The predicted phenomena involved task and reward distributions. The predictions were that people would cluster together adjacent states that regularly co-occur in a task and that people would cluster adjacent states having similar rewards. These predictions were validated in eight behavioral Amazon Mechanical Turk experiments. They reported how incremental learning of abstractions incrementally improved planner performance across trials.

> "Whether planning something simple like cooking dinner or something complex like a trip abroad, we usually begin with a rough mental sketch of the goals we want to achieve ("go to Spain, then go back home"). This sketch is then progressively refined into a detailed sequence of subgoals ("book flight," "pack luggage"), sub-subgoals, and so on, down to the actual sequence of bodily movements that is much more complicated than the original plan. Efficient planning therefore requires knowledge of the abstract high-level concepts … In this study, we show that *humans spontaneously form such high-level concepts* in a way that allows them to plan efficiently given the tasks, rewards, and structure of their environment." (Tomov et al., 2020)

This computational model of abstraction discovery works for all kinds of problem solving, not just navigation problems. It does not require an agent to select a specified target structure in the environment or to pick from a set of possible target structures. Rather, the structure in the environment or task is assumed to be hidden, discoverable in patterns of agent behavior in solving problems.

---

[31] See (Tomov et al., 2020) for a discussion of the hierarchy discovery and planning algorithms and the design of the behavioral experiments that were used to test the predictions of the model. Their paper includes an extensive comparison of the approach to previous work on model-based and model-free RL, various approaches to hierarchical reinforcement learning (HRL), and option discovery methods.

This computational model has a *slow discovery process* that incrementally improves the performance of a *fast planning process*. Its two-component cognitive structure is consistent with Graybiel's neuroscience experiments involving the basal ganglia where activity patterns in the striatum are modified gradually during the course of learning and slow learning supports later fast planning.

The research on abstraction discovery illustrates the workings and benefits of the multidisciplinary research cycle. Neuroscience research provided a distinct perspective on the largely overlooked problem of abstraction discovery. Cognitive psychology experiments illuminated the cognitive patterns of people using abstraction discovery. AI researchers built a detailed computational model that led to testable functional insights about how learned internal representations can gradually improve a planner's decision making. The net result has been a deeper understanding of how incremental abstraction discovery works, recognition of the slow and fast cognitive processes involved, and a bio-inspired computational model unlike those that were discussed when hierarchical planning became an early AI research topic.

Tomov et al.'s computational model for discovering useful abstractions is powerful, appears to predict human performance data, and is compatible with a developmental approach. Next research steps need to experiment with incorporating the method generally as part of an incremental learning algorithm.

In summary, intelligent agents use multi-step action sequences in the world. As a newborn infant matures, its learning progresses from learning the basics about its body, senses, motor control, and body to learning multi-step actions, abstract sensory percepts, and abstract goals.

- *Abstraction discovery hypothesis.* Abstraction discovery learns useful abstract states, abstract percepts, and abstract actions (chunking of sequences of actions). It has methods for translating these to lower level states, percepts, and actions. It predicts outcomes of abstract actions on abstract states. The abstraction learning system is slow and separate from the motor execution system. Training can involve observing examples generated by motor babbling. The learning process is unconscious, incorporates high-level multi-dimensional data, does not do motor control, and learns abstraction actions and states. It informs the action selection and motor execution systems with learned abstractions that map to possible action sequences.

A recurring theme is that new systems do not replace old systems, but rather, are added to them. Newer systems involving the neocortex and other brain systems learn over time from the high-dimensional contextual conditions that arise in ongoing experience. The neocortex learns but does not directly exercise motor control. Instead, it sends information to parts of the brain that evolved earlier.

### *4.3 Bootstrapping Curiosity and Intrinsic Motivation*

Goals, plans, and abstractions are the cognitive work horses of action planning. However, many important cognitive activities *do not have precise goals*. Such open-ended activities are associated with curiosity, exploration, scientific discovery, creativity, and invention.

How do exploratory activities work? How do the exploratory activities done by children compare to those done by adults? How should these activities be incorporated in the bootstrapping of AIs? What are the neurological and behavioral findings about them? What is the state of the art and what are the remaining challenges for creating computational models of such activities?

We explore these questions about curiosity starting with a concrete example of the toy baby bottle in Figure 11.

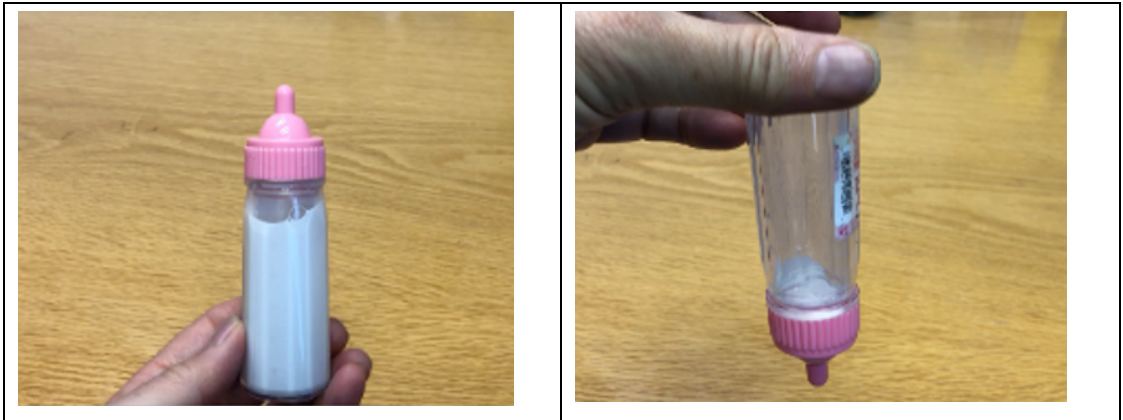

*Figure 11. "Mystery" milk bottle toy. "Where does the milk go?"*

When the toy bottle is upright, it appears to be full of white "milk." When it is inverted, the milk flows down toward the pink nipple end. However, unlike what happens with a normal baby bottle, no milk comes out of the toy bottle.

Infants and some adults see this toy but do not notice anything unusual. Other people realize suddenly and without effort that something about the bottle is anomalous. They have a nearly instantaneous VOE (violation of expectations) experience and may say "wow!" They may think about it and investigate how the toy works. They may notice that the enclosed volume of the main part of the bottle is several times larger than the enclosed volume of the pink nipple. How could the "milk" fit in the small nipple end? They may eventually experience an "Aha!" reaction, a sudden insight when they understand how the toy bottle works.

How does the toy bottle work? Like other mystery devices, it has a hidden mechanism. Inside the bottle there is a sealed transparent cylinder. When the bottle is upright, only a tiny volume of "milk" is needed to occupy the space between an invisible inner cylinder and the outside shell of the bottle.[32] When the small amount of milk flows downwards in the inverted bottle, it fits easily inside the nipple.

---

[32] Readers puzzled about how the boy baby bottle works are invited to consult an explanation at https://babiesplannet.com/how-do-toy-baby-bottles-work-the-truth-explained/ A similar hidden mechanism is the basis of the *magic water fountain*. For a video of a magic water fountain, see https://www.youtube.com/watch?v=a_or5f59A04

For another example, consider the following scenario.

> **Building a first block tower**
>
> Macer just had his second birthday. His parents gave him a set of toy wooden blocks. They put Macer on the floor with his blocks.
>
> Macer's vision and motor control have improved over the past months. Now he can push the blocks around and crawl after them. He picks up a block, turns it over with his hands, and studies it. When Macer lets go of the block, it falls to the rug and comes to rest on one of its faces.
>
> Macer picks up the block again and drops it above another block. It hits the resting block, teeters, and falls off. Macer stares at the two blocks. He picks the first block up again and places it slowly on the resting block. The block starts to tip but Macer pushes it towards the center of the resting block. It stabilizes and does not fall off. Macer stares intently at the two blocks. Then he crawls forward on the rug, accidentally bumping them. The blocks topple and the upper block falls off.
>
> A few days later Macer builds another small tower with two blocks. He concentrates and successfully places a third block on top of the other two.

Why does Macer build the block tower? An unsatisfying and overly simple teleological account might say that Macer has a goal to build block towers. How could a child acquire such a goal? An intrinsic motivation theory would favor a different kind of account for what happens when a child builds its first "tower" with two or three blocks. When the first block tower does *not* fall down, a child might think, "Wow! That's interesting! Usually when I pile things up, they come part and fall down." Children may investigate variations of block placement to see what happens under different conditions. As they explore the world, they learn how block stacking and other things work.

*4.3.1 Research about Human Curiosity and Intrinsic Motivation*

In *The Scientist in the Crib,* Alison Gopnik and her colleagues found that children test hypotheses and explore in ways akin to the scientific method (Gopnik et al., 1999). Among other examples, they call attention to a period in child development known to parents as the "terrible twos." Parents famously dread this period because two year olds can seem to be particularly perverse. Gopnik and her colleagues interpret an example of toddler behavior in this period as follows:

> "While one-year-olds seem irresistibly seduced by the charms of forbidden objects (the lamp cord) … the two-year-olds are deliberately perverse. … A two-year-old doesn't even look at the lamp cord. … he looks, steadily … at you. … The terrible twos seem to involve a systematic … experimental research program …  you and your reaction … are the really interesting thing. If the child is a budding psychologist, we parents are the laboratory rats." (Gopnik et al., 1999)

In this interpretation of the terrible twos, the toddler's behavior is like that of scientists doing experiments to learn how the world works. How do children, adult scientists, and inventors create or discover goals to invent and understand new things? In their book *Breakthrough*! Mark and Barbara Stefik explored these questions in the context of scientists and inventors in Silicon Valley. They interviewed repeat inventors and scientists across Silicon Valley and elsewhere to gain insights about effective practices for invention and discovery (Stefik and Stefik, 2004).

> "When people hear about a breakthrough, they are surprised and often ask 'How did they think of that?' A breakthrough can be surprising either because it works in a surprising new way or because it satisfies a surprising new need.
>
> In either case, the inventor saw something that others missed. … This has to do with how they notice and understand rather than what they are presented with. Sometimes inventors find curiosities in ordinary things. These curiosities become the seeds of their inventions." (Stefik and Stefik, 2004)

Following a suggestion from Joshua Lederberg,[33] the authors identified four strategies of behavior by repeat inventors and discoverers.

> "In the *theory-driven* approach, a mental model or theory provides a way of thinking that leads to insight or invention. In the *data-driven* approach, an anomaly in data reveals a surprising possibility. In the *method-driven* approach, instrumentation enables previously unknown observations, discoveries, and invention. The *need-driven* approach identifies problems and seeks solutions. ... A tag line for … [the theory driven] approach might be 'According to my theory. … The tagline for [the data driven] approach might be 'Now that's strange!" … The tag line for [the method driven] approach might be 'Now I can see it!' … The tag line for the [need-driven] approach might be 'Necessity is the mother of invention.'" (Stefik and Stefik, 2004)

The four approaches are often combined. For example, a scientist observing a novel phenomenon with a new instrument (method-driven approach) may notice that a phenomenon is novel (data-driven approach) and offer a new model to account for it (theory-driven approach).

The approaches emphasize different steps in learning about the world. They increase the probability of discovering or inventing something novel and useful. The data-driven approach takes enough observations of the environment to differentiate what is typical from what is unusual. The method-driven approach exploits new kinds of observations (including observing the effects of actions) to observe what happens in novel conditions. The theory-driven approach creates predictive models for how the world works. It systematically and statistically relates actions taken to their effects in the world. The need-driven approach prioritizes taking actions that might contribute to solving a problem. In the following, these four "approaches" correspond to steps in models of curiosity and intrinsic motivation.

Gopnik and her colleagues offer a caution about comparing the activities of babies with mature scientists and inventors.

> "Babies and young children think, observe, and reason. They consider evidence, draw conclusions, do experiments, solve problems, and search for the truth. Of course, they don't do this in the self-conscious way that scientists do. And the problems they try to solve are everyday problems about what people and objects and words are like, rather than arcane problems about stars and atoms. But even the youngest babies know a great deal about the world and actively work to find out more." (Gopnik et al., 1999)

In short, curiosity drives people, animals, or AIs to learn about the world.

In 1994, George Loewenstein compared several formal psychological frameworks for studying curiosity (Loewenstein, 1994). He noticed that the multiple frameworks asked different questions. How can a rational choice framework explain curiosity? Why does the amount of effort to satisfy curiosity so often exceed the value of the information? How should a reliable and valid curiosity scale be defined? Each psychological framework that he studied worked well for some examples of curiosity but raised issues for others. Ultimately, Loewenstein found the extant psychological frameworks disjoint and incomplete for accounting for the wide variety of behaviors associated with curiosity.

---

[33] The four approaches were suggested to the authors by Joshua Lederberg in a personal communication on October 12, 2002. He also carried out an analysis of his own research and that of close colleagues. See Chapter 3 of (Stefik and Stefik, 2004) for particulars of the inventions and scientific discoveries and stories in the words of the principals.

He suggested that:
> "Psychologists representing diverse intellectual perspectives speculated about the cause of curiosity and invariably concluded that curiosity could be explained in terms of their own preexisting theoretical frameworks. ... [More broadly,] this article views curiosity … as a cognitively induced deprivation that results from the perception of a gap in one's knowledge." (Loewenstein, 1994).

Skipping forward to 2015, Celeste Kidd and Benjamin Hayden followed up on Lowenstein's conclusions in their paper *The Psychology and Neuroscience of Curiosity*. They agreed that earlier research on curiosity did not explain its biological function and neural mechanisms. They also agreed that investigations into curiosity were hindered by the lack of a widely agreed upon definition of what is and is not curiosity. They advocated a simple characterization of curiosity.

> "[W]e provide a selective overview of its current state [of research on curiosity] and describe tasks that are used to study curiosity and information-seeking. We propose that, rather than worry about defining [narrow categories of] curiosity, it is more helpful to consider the *motivations for information-seeking behavior* and to study it in its ethological[34] context. … we favor the rough-and-ready formulation of curiosity as a drive state for information. Decision-makers can be thought of as wanting information for several overlapping reasons just as they want food, water, and other basic goods. This drive may be internal or external, conscious or unconscious, slowly evolved, or some mixture of the above [italics added]." (Kidd & Hayden, 2015)

Kidd and Hayden drew on a methodological approach proposed by the Dutch biologist Nikolaas Tinbergen. Tinbergen proposed four subtopics: function, evolution, mechanism, and development (Tinbergen, 1963). This framing of *curiosity as information-seeking* aligns well with White's motivational drive, Gopnik's observations of children, and the informal studies of invention and scientific practice. It unifies the multiple psychological distinctions under one umbrella. Curiosity motivates the acquisition of knowledge. The acquisition of knowledge benefits organisms by improving their performance and fitness. This framing opens curiosity research to experimentation and modeling.

Findings about curiosity have emerged from a combination of neuroscience investigations, behavioral investigations, and computational modeling. We begin with neuroscience findings about insight and the "Aha!" reaction. Often reported by scientist and inventors when a solution to a problem occurs to them suddenly, the "Aha!" reaction has attracted the study of psychologists for many years. Reflecting on his own "Aha!" moments, Jeff Hawkins observed:

> "Aha moments occur when a new idea satisfies multiple constraints. The longer you have worked on a problem—and, consequently, the more constraints the solution resolves—the bigger the aha feeling and the more confident you are in trusting the answer." (Hawkins, 2021)

Advances in instrumentation for brain scanning led John Kounios and Mark Beeman to carry out controlled experiments.[35] The available temporal resolution of EEG and spatial resolution of fMRI enabled Kounios' and Beeman's to monitor brain activity around "Aha!" moments. The data revealed properties of cognitive processes that are *below conscious deliberation*, providing insights bearing on AI bootstrapping that were not available earlier. By way of contrast, the verbal protocol analyses of human cognitive activities by Karl Ericson and Herbert Simon and others studied only *conscious processes* of deliberative problem solving (e.g., Ericson and Simon, 1993).

---

[34] Ethology is the study of human and animal behavior from a biological perspective.

[35] Kounios and Beeman's experiments followed a method-driven approach – taking advantage of data or "seeing differently." Their experiments used neuroimaging and electrophysiological instrumentation and techniques to study brain activities during "Aha" moments.

> "A series of studies have used electroencephalography (EEG) and functional magnetic resonance imaging (fMRI) to study the neural correlates of the "Aha! moment" and its antecedents. Although the experience of insight is sudden and can seem disconnected from the immediately preceding thought, these studies show that insight is the culmination of a series of brain states and processes operating at different time scales. … insights occur when a solution is computed unconsciously and later emerges into awareness suddenly." (Kounios and Beeman, 2009)

To summarize, Kounios and Beeman's study concludes:

- Insight is sudden, but it is preceded by substantial unconscious processing.
- Some critical components of insight are preferentially associated with the right cerebral hemisphere. Insight culminates with a sharp increase in neural activity in the right anterior temporal lobe at the moment of insight.
- Insights are immediately preceded by a transient reduction of visual inputs that apparently reduce distractions and boost the signal-to-noise ratio of the solution.

A later review article (Kounios and Beeman, 2014) reported that insight can suddenly intrude on a person's awareness when the person is not focusing on the problem. It can also intrude when a person has not yet reached an impasse. These findings bring to mind the famous quote by Louie Pasteur that "In the fields of observation, chance favors the prepared mind." Preparation precedes insight. Preparation can include deliberative problem solving and it often involves unconscious processes. A moment of insight is dramatic, but progress in intrinsically motivated activities is not just about a moment of insight.

While manually designed sequences of increasingly complex experiences and reward functions are useful for guiding the learning of AIs on one or two tasks, manual approaches to specify the experiences and rewards do not scale for AIs that need to perform thousands or more kinds of tasks in open-ended and evolving situations. How do people choose what to do when they are exploring the unknown? How do they choose good challenges and good places to focus their attention?

In his writings about creativity and peak performance, Mihalyi Csikszentmihalyi described an insight into what creative people experience when they are operating "in the flow" in bursts of optimal performance (Csikszentmihalyi, 1991). They experience the greatest internal rewards for mastering challenges when the time and effort for each challenge is intermediate – neither too small nor too large.

*4.3.2 AI Systems with Curiosity and Intrinsic Motivation*

In the following, we draw on reviews of the state of the art by Pierre-Yves Oudeyer, Frédéric Kaplan, and Verena Hafner (Oudeyer et al., 2007a; Oudeyer et al., 2005) and sketch their Adaptive Curiosity (IAC) mechanism.

We begin with a parallel "Act-Predict-Evaluate" computational pattern in Figure 12 as discussed in the side bar. This pattern underlies computational models of decision making and elaborates the high-level Observe-Act cycle in Figure 9.

Oudeyer et al. offered a model that learns to predict the outcomes of a robot's actions and that actively chooses actions related to the novelty of the prediction based on an internal metric.

Restated, early curiosity systems sought out novel situations and learned from them. Using the terminology of Oudeyer et al., early learning systems included (1) a model **M** that learns to predict the sensorimotor consequences when a given action is executed in a given context, and (2) a meta-learning model **metaM** that learns to predict the errors that **M** makes in its predictions.[36]

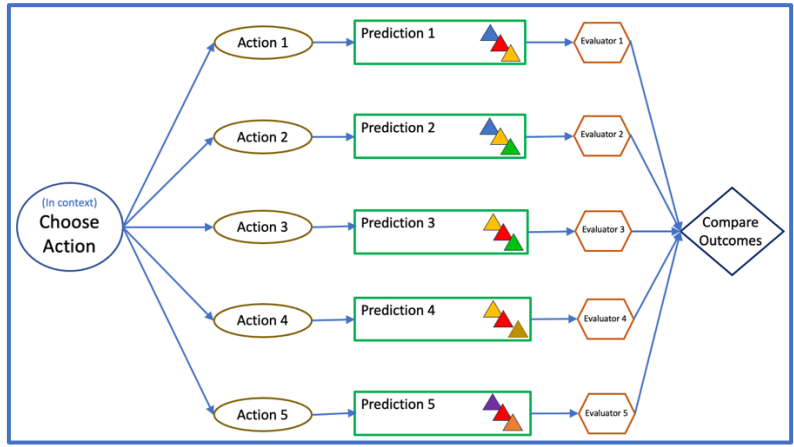

*Figure 12. Pattern including **observations, actions, predictions**, and **evaluations**. Computational versions of this pattern have been used to model choice making in different kinds of activities including diagnosis, design, optimal experiment design, and curiosity.*

Oudeyer and his colleagues divide earlier computational models of intrinsic motivation into three classes.

> (1) The *error maximization* approach chooses the action at each step where **MetaM** predicts the largest error in the prediction of **M**. In short, it focuses on examples where its ongoing predictions fail the most.

**A Parallel Act-Predict-Evaluate Pattern**

This side bar describes a basic computational pattern for choosing an action. Variations of the pattern are used in computational models for choosing actions in different kinds of activities.

In the basic version, an agent in a situation uses a Predictor to predict what would happen for each of its action choices. For each choice, an Evaluator computes a value for the predicted outcome. The agent chooses an action that has the greatest utility.

In variations of the basic pattern, a situation may be only partially observable. The situation may have probabilistic distributions of features. The model may have a single "Predictor" model applicable for all its actions or have a different specialized Predictor for each choice. The evaluation criteria may combine competing considerations such as costs, benefits, and risks. The outcome may be represented as a probabilistic distribution. Models may include assumptions about noise.

Variations of this framework have been applied to a wide range of decision problems. In a design problem, the agent makes choices to achieve the greatest value that meets design criteria. In planning problems, the agent chooses multiple coordinated actions that satisfy plan constraints and maximize the plan's utility. In diagnostic problems, the agent chooses an action (such as the location to place a probe) with the goal of maximizing information gain and narrowing the search space of diagnostic hypotheses. In a repair problem the agent considers both the cost of diagnostic probes and the cost of repairs. In reinforcement learning (RL) formulations, an agent maximizes a cumulative reward across multiple decisions, allocating credit across intermediate decisions, balancing between the exploration of new cases and rewards based on previous knowledge (exploitation).

Early computational models of curious agents included provisions to consider both immediate utility and cumulative learning.

Although the error maximization approach is effective in situations where **M** can learn a mapping in a deterministic environment (even with Gaussian noise), it fails when the environment is too complex for its learning program. For example, the robot can get stuck when it runs into a wall and then bounces off in unpredictable ways. Failing to predict its bounces accurately, the robot keeps focusing on them. This behavior leads to getting stuck, but not mastery. The robot keeps trying to predict the bounces. A second approach avoids this problem.

---

[36] This computational pattern is consistent with the theme in Bruce Buchanan's AAAI presidential address that characterized creativity as search at a meta-level (Buchanan, 2000).

(2) The *progress maximization* approach tries to avoid getting stuck in noise. It adds a knowledge gain assessor (**KGA**) that predicts the mean error rate of **M** in the sensorimotor contexts in the near future. It evaluates action choices using the difference between the expected mean error rate of the predictions of **M** and the mean error rate in its near past. It chooses the action that will lead to the greatest decrease of the mean error rate. Rephrased, it avoids repeating actions when they do not lead to improved predictions.

Although this second method avoids getting stuck, it analyzes the agent's behavior assuming that there is only one kind of activity. If the robot's choices include different kinds of activities such as visual tracking and grasping objects, the apparent rates of recent change computed by the **KGA** are neither easy to understand nor meaningful. This observation leads to a third approach.

(3) The *similarity-based progress maximization* approach organizes its memory of situations into groups of similar situations and monitors the evolution of error rates separately for each group. Rather than lumping together all actions in its close past, it takes into account different kinds of actions for different functions. However, a challenge is that this approach needs a computationally effective way to develop a *measure of similarity* in order to organize situations into appropriate categories.

Addressing these limitations of earlier approaches, Oudeyer and his colleagues proposed an *intelligent adaptive curiosity* (IAC) approach. Robotic IAC systems move towards cognitive situations where they can learn things. IAC is adaptive because the criteria for which situations are attractive varies over time. IAC moves away from situations that are too predictable and ones that are too unpredictable.

The IAC formulation is akin to a reinforcement learning formulation with a mechanism providing a positive or negative reward each time an action is performed. In contrast with the typical actor-critic formulation of an RL formulation, the IAC critic is internal rather than external.[37]

Parallel to Csikszentmihalyi's observations, IAC chooses challenges that increase in complexity *at the right pace*. This amounts to an explanation for the efficacy of a "Goldilocks Principle" where a medium level of effort offers reasonable progress without stalling.

> "… infants' activities always have a complexity which is well-fitted to their current capabilities. Children undergo a developmental sequence during which each new skill is acquired only when associated cognitive and morphological structures are ready. … Eventually, they [must] decide by themselves what they do, what they are interested in, and what their learning situations are. … Thus, they construct by themselves their developmental sequence." (Oudeyer et al., 2007a)

As research interest has grown in intrinsic motivation and exploratory behavior without goals, Andrew Barto observed that psychology and neuroscience see reward signals as just one part of motivation.

> "… Psychologists distinguish between extrinsically motivated behavior … and intrinsically motivated behavior ... Is an analogous distinction meaningful for machine learning systems? Can we say of a machine learning system that it is motivated to learn, and if so, is it possible to provide it with an analog of intrinsic motivation? … the answer to both questions is

---

[37] See (Oudeyer, 2007a) for detailed descriptions of the IAC sensors and processing and mathematical characterizations for assessing predictions of IAC's models, its action selections, and its learning progress. The paper (Oudeyer, 2005) describes the experimental setting for their "playground experiment."

assuredly 'yes,' and that the machine learning framework of reinforcement learning is particularly appropriate ... [for formulating this combination] … contemporary psychology and neuroscience indicate that the nature of reward signals is only one component of the complex processes involved in motivation. … Because RL addresses how predictive values can be learned and used to direct behavior, RL is naturally relevant to the study of motivation." (Barto, 2012)

Other examples of using RL to model intrinsic motivation include combining estimation and control in RL without reward engineering (Singh et al., 2019), integrating temporal abstractions with RL for efficient deep RL in complex environments (Kulkarni et al., 2016), and building models for exploring their space given various physical constraints (Frank et al., 2016).

In a review editorial[38] about the field of extrinsic motivation, Vieri Santucci, Pierre-Yves Oudeyer, Andrew Barto, and Gianluca Baldassare suggested that the field of intrinsic motivation still has puzzles and challenges for building computational models and experiments. They characterized intrinsic motivation in part in terms of the discovery of new goals. ("I didn't know we could do that!") They summarized the state of the AI research as follows:

> "Notwithstanding the important advances in Artificial Intelligence (AI) and robotics, artificial agents still lack the necessary autonomy and versatility to properly interact with realistic environments. This requires agents to face situations that are unknown at design time, to autonomously *discover multiple goals/tasks* and to be endowed with learning processes able to solve multiple tasks *incrementally and online.*" [italics added] (Santucci et al., 2020)

Annie Xie, Frederick Ebert, Sergei Levine, and Chelsea Finn reported on an AI system whose exploration and model building led it to use a "method-driven" strategy. The AI combined its models about simple physics and its models for action to choose to use some objects in its world to move other objects (Xie et al., 2019). It used objects in its environment as *tools* to enhance its actions.

The behavior of intrinsic motivation and curiosity can be understood as *goal discovery*. The behavior for goal discovery complements the behavior of *abstraction discovery* in the previous section. Abstraction discovery is about creating abstraction actions and goals that generalize actions and goals previously taken by an agent. Restated, it is about noticing interesting abstractions. Tomov et al.'s computational model describes that process.

Goal discovery is about analyzing the current situation for interesting anomalies and opportunities that were not previously encountered. Restated, it is about noticing interesting states as goals. Oudeyer et al.'s analysis describes computational models for goal discovery. Their analysis framework is compatible with a developmental learning-based approach.

Next research would experiment with incorporating the approach as part of a *perceive-act cycle* for an AI, or as more elaborately described, an *observe-predict-evaluate-act-learn* cycle. This is a larger behavioral cycle than that for abstraction discovery.

In summary, the behavior required for effective learning in the environment combines features of prediction, internal rewards for learning, chunking, and hierarchical planning.

> *Goal discovery hypothesis.* Intrinsic motivation – as a driver of curiosity and scientific discovery – searches for and selects new abstract goals based on experiments and interesting anomalies. To explore possible goals, it maps higher level goals to possible lower level goals, makes predictions, takes actions, and learns. Training involves observing examples generated by lean exploring.

---

[38] Oudeyer and colleagues reviewed theories of intrinsic motivation, curiosity, and learning (Oudeyer, et al., 2016). Their review compared the merits and challenges of the different theories from an engineering and computational perspective for building robots. The paper also reflects on the application of these insights in the creation of educational technology for humans.

With intrinsic motivation, AIs could potentially bring curiosity and creativity to their activities. Like creative humans, they could experiment to improve its models of the world, inspired by anomalous events.

The repeat inventors and researchers described earlier have deliberate models for creating favorable conditions for learning and getting novel scientific results. The four strategies from the *Breakthrough!* book are examples of such deliberate strategies. As in the cautions by Gopnik et al., infants do not explore the world "in the self-conscious way that scientists do." Nonetheless, every scientist and inventor starts out as a child and some of those children acquire these strategies either as self-developed competences or as socially developed competences. A vision for future AIs has them ultimately acquiring these strategies too.

### *4.4 Bootstrapping Imitation Learning*

Imitation learning is learning how to do something (e.g., a task) by observing another person do it. Imitation learning is sometimes called observational learning.

Imitation learning is an important way to acquire socially developed competences. Phrased in terms of agents, a learning agent learns from a demonstrating agent without needing to discover the relevant goals and multi-step action sequences on its own. Afterwards, the learning agent can demonstrate the skill.

The sidebar describes the related engineering discipline of "Learning from Demonstration" (LfD). LfD simplifies the programming of robots. Overviews of LfD by Harish Ravichandar et al. and Brenna Argall et al. are available (Ravichandar, 2020; Argall et al., 2008).

> **Engineering Practices for**
> ***Learning from Demonstration* (LfD)**
>
> *Learning from Demonstration* (LfD) refers to engineering practices used for producing practical (and specific) robot applications with less programming than detailed manual specification. AI research on imitation learning seeks general methods for autonomous learning of competencies by observation.
>
> Ravichandar et al. review the art and the limitations of different demonstration approaches including kinesthetic teaching, teleoperation, and passive operation (their term for imitation learning), and approaches for correcting robot behavior. They describe engineering choices for controlling the representations of what is learned (e.g., low-level control, long horizon planning, and multi-step sequences of actions) and the compatibility of the representations with control programs.
>
> As Ravichandar et al. report in their overview:
>
>> " learning from demonstration is the paradigm in which robots acquire new skills by learning to imitate an expert. The choice of LfD over other robot learning methods is compelling when ideal behavior can be neither easily scripted (as is done in traditional robot programming) nor easily defined as an optimization problem [for reinforcement learning] but can be demonstrated." (Ravichandar et al., 2020)
>
> Traditional robot programming requires programming expertise to specify the actions that a robot takes to perform a task. LfD practices are simpler than manual robot behavior design. They employ machine learning to learn context and constraints from structured demonstrations.
>
> The best engineering practices of LfD advance as technology advances become available from AI research in computer vision, modeling, and imitation learning. For example, Ajay Mandlekar and his colleagues combine human-in-the-loop teleoperation and motion planning techniques to train robots more quickly on "long horizon" tasks (Mandlekar et al., 2023). In their approach, the motion planner hands off to a human parts of tasks that it cannot do automatically. This enables the human to spend less time on demonstrating tasks. This approach decomposes a task into planning-based and learning-based segments. As they describe it,
>
>> "By soliciting human demonstrations only when needed and allowing a human to participate in multiple parallel sessions, our system greatly increases the throughput of data collection while lowering the effort needed to collect large datasets on long-horizon, contact-rich tasks." (Mandlekar, 2023)
>
> In summary, LfD is an applied engineering practice that employs techniques from research on imitation learning. The engineering and application effort for preparing a robot for a particular task is shared between (1) high-level experts that set up an application framework for the robot and the range of tasks, and (2) application technicians that teach the robot and may edit representations of what the robot has learned to adjust its behavior manually.

*4.4.1 The Correspondence Problem*

In *Nine Billion Correspondence Problems* (Nehaniv, 2007), Chrystopher Nehaniv defines imitation learning as when a learning agent observes and then produces a behavior that is similar to the behavior of a demonstrating agent.

Nehaniv's broad definition of the correspondence problem raises questions. What is matched? Which aspects of the imitated behavior need to be similar, and which can be different? For example, suppose that the learning agent has limbs with different dimensions than the demonstrating agent – so the motion trajectories of its limbs are different. The "nine billion" term in the title is a nod to the many ways that the learning agent could need to make multiple adjustments in what it pays attention to and what it does.

In natural settings, imitation learning often operates very quickly. The importance and high human performance of imitation learning raises intriguing questions:

(1) How can the correspondence problem be solved generally for AIs?
(2) How does imitation learning – including solving the correspondence problem – work so fast?
(3) How does it work with just a few observations, that is, with so little data?

The following fictional "Traveler and Bakers" scenario[39] illustrates five competence sets required for imitation learning.

**The Traveler and the Bakers**

A traveler from a distant land came to an Italian mountain village on its annual baking day in late December. This was the traditional day for men of the village to gather at the community oven. They make the hard bread that their families will eat with hearty soup over the next few months. The traveler came from a town that mostly ate rice and not much wheat. He was unfamiliar with the hard dark bread of alpine tradition and with how it was made.

The traveler watched the bakers working. One baker was at a trough. A table in the kitchen had a dusting of a brown powder. Occasionally, a baker would sprinkle some water on the dough that he was kneading, roll it in the powder on the table, and then shape the loaf.

The bakers put each loaf on a long wooden paddle and pushed it into the hot oven. What were the bakers doing?

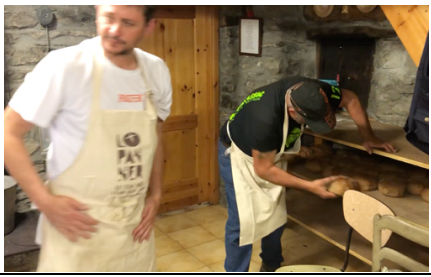
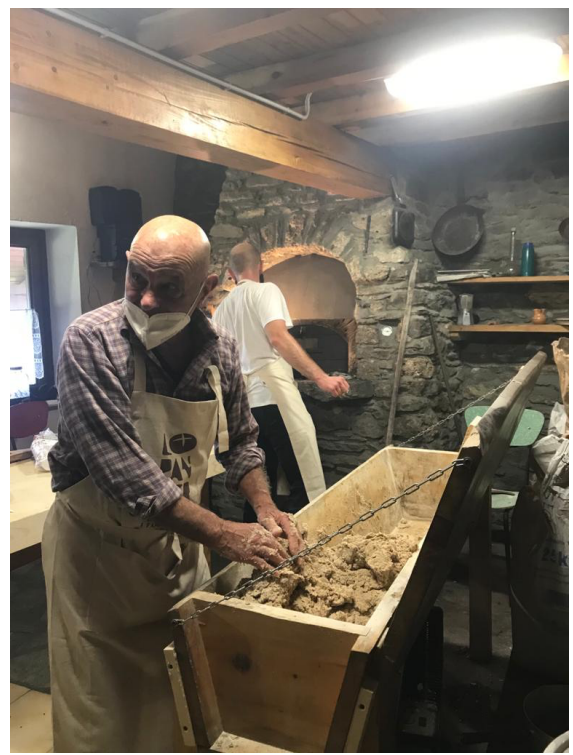

---

[39] The photos used to illustrate the bakery in the fictional "Traveler and Bakers" scenario are from the mountain village bakery in Vetan, Italy. They were taken by Barbara and Mark Stefik in 2021.

The first two competence sets have been discussed already. They are prerequisites for the next three competence sets which address the correspondence problem.

A first competence set[40] required for imitation learning is for the learning agent to (1) acquire models of the world including its own body and how it senses, moves, and acts in the world. As described previously for people, it is where people learn how to see, crawl, walk and do things in the world.

A second competence set[41] required for imitation learning is for the learning agent to (2) develop models of its own abstract goals, states, and actions. For example, by forming abstractions people generalize and adapt more quickly to other situations.

A third competence set extends perception and modeling (3) *to understand* a demonstrating agent's physical actions and the setting. In the photos of the scenario, the bakers are human, their arms and hands are visible, and the table, dough trough, and oven are in view. If the bakers had been wearing clothing that concealed their movements, understanding their physical actions would be more difficult.

A fourth competence set (4) *ascribes goals and purpose* to the actions of a demonstrating agent. People already model how their actions achieve their own goals. This competence set is about understanding a demonstrator's goals. In the scenario, the traveler does not initially understand the purpose of the bakers' actions. Only when the loaves are taken from the oven does the traveler realize that the purpose of the actions was to produce loaves of bread. Observing the overall process, the learner understands the purposes of each of the steps of manipulating flour, water, dough, oven paddles, and so on.

A fifth competence set for the learning agent is (5) *to translate* what it sees to what it can do. Suppose that a baker is very strong, and that the traveler is not. When the observer's body or the available materials differ from those of the demonstrator, the observer may need to modify the steps and use tools to accomplish the same purpose. For example, the traveler in the scenario may later need to substitute a different kind of flour or to use an electric oven rather than a woodburning one. Suppose that the observer has different limbs than the observer. In such cases, the learning agent needs to find different ways to achieve analogous results. The mappings and possible substitutions can be guided by reasoning from abstract operations and goals rather than from the specific actions of the demonstrator. For example, a learner might reason from an abstract step like "put a loaf in the oven," rather than from the individual movements of the demonstrator's appendages.

Chrystopher Nehaniv groups the overall challenge of imitation learning -- mapping the learning agent's perceptions of the demonstrating agent to actions of the learning agent – as the *correspondence problem* (Nehaniv, 2007).

The correspondence problem as defined by Nehaniv includes challenges that are not typically part of LfD technology as discussed in the sidebar. For example, when people train a robot through teleoperation, they operate the robot's own limbs. No adjustments are needed to account for differences in limbs or other effectors. An LfD system does not typically use sensors to observe a human. The LfD system does not typically discover new goals or abstractions. The goals for the LfD system are entered by a human rather than inferred by a learning system. As in the sidebar, LfD systems can be built with a combination of human-in-the loop interfaces and AI technology. They may require a human operator or trainer to set parameters on behavior or change internal representations of plans and goals.

---

[40] Section 4.1 describes how an agent can learn models of its world, including its own body. Section 4.2 describes agents acquiring goals, multi-step action sequences, and abstraction discovery.

[41] Section 4.3 describes how agents can discover new goals (intrinsic motivation).

Developmental AIs will be expected to operate more "in the wild" than today's robotic systems which move in highly confined and constrained settings and do largely repetitive tasks. In developmental AI, imitation learning builds on supporting competences such as object recognition. Imitation learning will be triggered by inherent drives such as curiosity when an AI observes a person doing something new – and then creates a new goal to learn how to do it. Imitation learning supports other competences such as coordination and teaming, where new members of a team learn how to substitute for others. Aspirationally, developmental AIs will combine multiple competences as they coordinate skillfully with people, learn from them and with them, and enhance the quality of teamwork.

Imitation learning predates the human species and occurs in non-human species. A famous and subsequently controversial[42] story of imitation learning by animals concerns the spread of potato washing skills by free ranging macaque monkeys on islands off Japan. Researchers introduced sweet potatoes to Koshima Island by putting them on a beach where the monkeys could find them. A young monkey named "Imo" was observed putting potatoes in the water and washing off the mud and sand before eating them. Over the next few years, researchers observed other monkeys following that practice and teaching it to their offspring.

Babies expand and accelerate their learning about the world by learning from other people. Learning by imitation as well as humans do it, however, is beyond the state of the art of current AI. How does it work? Do newborn babies learn by imitation? Does the ability to learn by imitation improve with maturity and experience? How could an AI bootstrap its ability to learn by imitation to the level of a skilled human adult?

*4.4.2 Research about Human Imitation Learning*

Researchers in several fields study imitation and social learning (Nehaniv and Dautenhahn, 2007). The research is multidisciplinary including neuroscience (e.g., Kang et al., 2021; Meltzoff et al., 2018; Saby et al., 2013; Rizzolatti et al., 2004, 1996); developmental and cognitive psychology (e.g., Williams, et al., 2022; Ramsey et al., 2021; Jones, 2009; Oudeyer et al., 2007b; Meltzoff et al., 1997), and animal studies (e.g., Gariépy, 2014; Galef et al., 2005; Tomasello et al., 2003; Galef, 1988). Ahmed Hussein and his colleagues reviewed computational methods for imitation learning (Hussein, et al., 2017).

In broad strokes, brain research focuses on mechanisms and activity of brain areas, cognitive research studies cognition and development, and AI research creates detailed multi-level models of imitation learning.

Addressing the question of whether imitation is innate in humans, Andrew Meltzoff and M. Keith Moore studied very young infants imitating facial expressions (Meltzoff et al., 1977). Gopnik and her colleagues gave a personal and high-level description of this research:

> "Twenty years ago … [Andy Meltzoff made] a startling discovery. One-month-old babies imitate facial expressions. If you stick your tongue out a baby, the baby will stick his tongue out at you; open your mouth, and the baby will open hers ... to demonstrate that this ability was really innate, he had to show that newborn babies could imitate. So, he set up a lab next to the labor room in the local hospital. … In order to imitate, newborn babies must

---

[42] More systematic methodologies for understanding imitation learning in animals have come into practice since the sweet potato report (Imanishi, 1957) first appeared and became a textbook example. Subsequent researchers have raised definitional distinctions, competing hypotheses, and issues with the evidence including a lower than expected rate of cultural transmission on the island. Andrew Whiten gives an overview of this line of research and the more complete models and nuanced methodologies for studying skill acquisition and transmission in animal populations (Whiten, 2000). The subsequent neural and cognitive research explores differences in the learning abilities of various species.

somehow understand the similarity between that internal feeling [of kinesthesia] and the external face that they see." (Gopnik, Meltzoff, and Kuhl, 1999)

Meltzoff and Moore's study concluded:

"Infants between 12 and 21 days of age can imitate both facial and manual gestures: this behavior cannot be explained in terms of either conditioning or innate releasing mechanisms. Such imitation implies that human neonates can equate their own unseen behaviors with gestures as they see others perform." (Meltzoff et al., 1977)

This research established that some imitation learning in humans is innate. How does the brain do it? To understand this question, neuroscientists set up experiments that scan brain activities in situations where a human (or animal) watches and imitates the behavior of another person or animal. The research focused on the neural activity of motor neurons, that is, the neurons that control an agent's own actions. Giacomo Rizzolatti and colleagues reported findings from their experiments that scanned the brains of macaque monkeys that were watching other monkeys perform activities.

"In area F5 of the monkey premotor cortex there are neurons that discharge both when the monkey performs an action and *when he observes a similar action made by another monkey* or by the experimenter. … We posit … that this motor representation is at the basis of the understanding of motor events [by the monkey and others]." (Rizzolatti et al., 1996)

The brain areas of both monkeys and people include maps of body parts such as a hand or a foot. As described earlier, somatotopy is the point-for-point correspondence of an area of the body to a specific point on the central nervous system. These findings by Rizzolatti and colleagues showed that mirror neurons in the monkey's brain that receive sensory signals from a body part are also activated *when the observing monkey sees a demonstrating monkey using its corresponding body part* in an activity.

Subsequent research about imitation learning mostly used non-invasive stimulation and scanning technologies (fMRI, TMS, EEG, MEG). Depending on the particular technique, activity is measured over a volume of the brain containing potentially millions of connections rather than measuring the activity of a single nerve cell. For this reason, the findings of the later research on imitation learning refer to mirror mechanisms, mirror phenomena, mirror systems, and mirror neuron circuits rather than to individual mirror neurons.

In 2004 Giovanni Buccino and his colleagues carried out an fMRI study of brain mechanisms involved in imitation learning by musically naïve participants playing guitar chords after observing a guitarist play (Buccino et al., 2004). They concluded:

"The results showed that the mirror neuron system is indeed at the core of imitation-based learning. They also indicate that during imitation learning, *the activity of the mirror neuron circuit is under control of the middle frontal lobe* … and anterior mesial cortical areas. … In conclusion, the results … identified the core circuit involved in vision-to-action translation. They also showed that other areas … play an important role in learning …" (Buccino et al., 2004)

Stepping back, questions arise about the nature and speed of the cognitive processes underlying imitation learning. Recall the five required competences for imitation learning:

1. *Acquire a model of the world* including its own body and how it senses, moves, and acts in the world.
2. *Develop the model of its own abstract goals, states, and actions* to facilitate adapting activities for different settings.
3. *Perceive and model the demonstrating agent's actions* and the setting of the activity.
4. *Ascribe goals and purpose* to the actions of the demonstrating agent.
5. *Translate* what it sees to what it can do.

Is the execution of the last three competences fast? As described in the following, this question has been investigated experimentally by multiple researchers and the answer is "yes." Although relatively rare cases of imitation learning may involve extended learning or reasoning (i.e., slow thinking), research finds that the usual performance of humans and animals in the wild that imitation learning is fast and is enabled by prior learning.

In 2005, Marcel Brass and Cecilia Heyes reviewed competing theories about brain mechanisms and the correspondence problem (Brass and Heyes, 2005). As they summarized,

> "*Specialist theories* suggest that this correspondence problem has a unique solution; that there are functional and neurological mechanisms dedicated to controlling imitation. *Generalist theories* propose that the problem is solved by general mechanisms of associative learning and action control. *Recent research in cognitive neuroscience, stimulated by the discovery of mirror neurons, supports generalist solutions.* … Generalist theories … imply that mirror neurons – and other neural systems … are active during both action observation and action execution … mirror neurons acquire their [learned] properties in the course of ontogeny as a side-effect of the operation of general associative learning and action control processes." [italics added] (Brass et al., 2005)

Supporting this position, Maddalena Fabbri-Destro and Giacomo Rizzolatti summarized findings by researchers about mirror mechanisms in human and animal studies.

> "The functions mediated by the mirror mechanism vary according to its location in the brain networks. Among the networks endowed with a mirror mechanism (mirror systems), the most studied is the one formed by the inferior parietal lobule and the ventral premotor cortex. The network transforms sensory representation of observed or heard motor acts into motor representations of the same acts. Its function is to give an immediate, *not cognitively mediated* [italics added], understanding of the observed motor behavior." (Fabbri-Destro and Rizzolatti, 2008).

Similarly, in a later paper, Giacomo Rizzolatti and Leonardo Fogassi described an expanding set of specialized functions for mirror systems discovered in further investigations about the rapid solution of the correspondence problem.

> "… 'canonical neurons' [are]… responsive to the presentation of three-dimensional objects. … 'mirror neurons' [are]…responsive to the observation of motor acts performed by others … the main property of canonical neurons is to match the shape and size of the observed object with a specific type of prehension, whereas the main property of mirror neurons is that of matching observation of hand and mouth motor acts with the execution of the same or similar motor acts. This matching mechanism enables the observing individual to achieve an automatic understanding – i.e., an understanding *without inferential processing of other' goal-directed motor acts*." (Rizzolatti and Fogassi, 2014)

Along similar lines, Dmitrii Todorov et al. modeled the interactions of the basil ganglia and the cerebellum in motor learning (Todorov, et al., 2019). Their studies supported two conclusions: *The basil ganglia is active and important in executing the motor control of imitation; and other parts of the brain are involved in learning how to do it.* These cortical neuron systems adjust the motor controls to handle variations in context. This division of labor between the basil ganglia and the cortex and other parts of the brain is analogous to the patterns described earlier for abstraction discovery and the processes underlying intrinsic motivation.

*4.4.3 AI Systems that Learn by Imitation*

Figure 13 shows a notional framework for comparing AI approaches to imitation learning and for relating them to findings from the neurosciences and cognitive psychology. In this framework the learning system for imitation learning advises the sensing and control system and is separate from it.

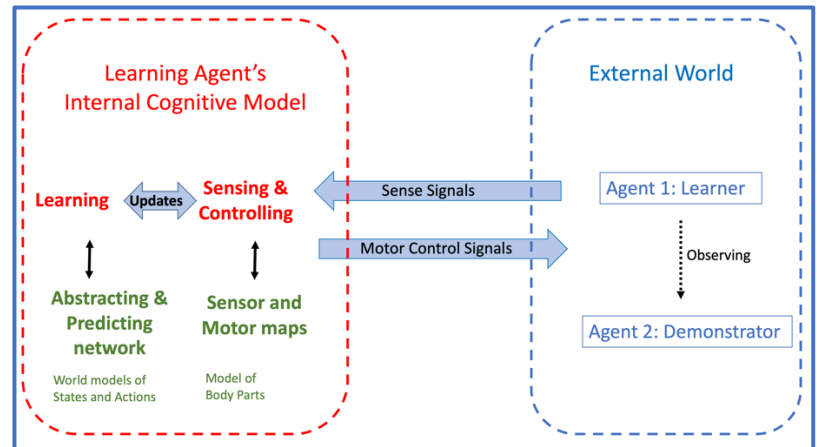

*Figure 13. Notional agent framework for imitation learning.*

Inspired by the neuroscience research, Stefan Schaal asked, "Is imitation learning the route to humanoid robots?" (Schaal, 1999). His paper reviewed the state of the art for humanoid robots, emphasized the need for a learning approach to govern behavior, estimated the scale of the state-action space and the combinatorial implications for a reinforcement learning approach, reviewed findings from cognitive science on imitation learning, and reviewed the nature of the perception-to-actions problem from a computer science perspective.

> "… research into imitation [learning] needs to include a theory of motor learning, of compact state-action representations or movement primitives, and of the interaction of perception and action. … these three components cannot be studied in isolation: perceptual representations serve motor representations, motor representations facilitate perception, and learning provides the mutual constraints between them." (Schaal, 1999)

Schaal concluded:

> "None of the approaches in this paper could provide satisfying answers to questions of appropriate perceptual representations for imitation, motor representations, and the learning of those representations." (Schaal, 1999)

More than two decades later, both of Schaal's conclusions still ring true. (1) Research has continued in neuroscience, psychology, AI, and robotics. (2) AIs have not yet achieved adult-level general competence via imitation or other kinds of learning. Imitation learning research projects in AI starting with activities similar to those of young children have not yet reached levels comparable to those of older students and adults.

AI research has led to more advanced concepts, embodied systems, and modeling for imitation learning (e.g., Haddadin, 2023; Arulkumaran et al., 2021; Torabi et al., 2019; Todorov et al., 2019;

Hussein et al., 2017; Cangelosi & Schlesinger, 2015; Ugur, et al., 2015; Law et al., 2014; Asada et al., 2009, Demiris & Dearden, 2005; Demiris & Johnson, 2003; Schaal, 1999; Brooks, 1991). In the following we sketch findings from this research and remaining challenges.

In 2005 Yiannis Demiris and Anthony Dearden proposed an architecture and experiments to combine other forms of learning with learning by imitation (Demiris et al., 2005). The agent used motor babbling to produce activity samples for training a Bayesian belief network to learn motor actions. They combined the babbling with examples from human trainers to drive imitation learning.

In 2009 Minoru Asada and colleagues surveyed research on brain development and cognitive development and proposed creating robots that would develop higher cognitive functions starting with sensorimotor development through social learning (Asada et al., 2009). They predicted that manual design of competences would usually fail because it would be based on shallow understandings of the requirements. Their paper cautiously framed an embodied approach for acquiring all of the competences at once rather than by developing modules for each of the competences separately. Their work included research with a child-sized robot.

Law et al. proposed a longitudinal and staged approach for developing competences. They reported findings from experiments using an iCub humanoid robot for competences akin to those developed in the first few months by human children (Law et al., 2014).

Extending this line of thinking, Emre Ugur and colleagues proposed a developmental framework where advanced stages of cognitive development for a real (embodied) robot would start with infant sensorimotor development (Ugur et al., 2015). Among other things this research demonstrated examples where the AI categorized groups of actions according to features that it could see demonstrated by others. It then advanced from simple to more complex sequences of actions.

In 2017 Ahmed Hussein and colleagues reviewed the cumulative state of the art of imitation learning and explored ways of combining the approaches that researchers have explored separately (Hussein et al., 2017). Their findings emphasize the importance of creating abstract models:

> "[Imitation learning approaches open] the door for many potential AI applications. … However, specialized algorithms are needed to effectively and robustly learn models as learning by imitation poses its own of challenges. … A key challenge in imitation is generalization to unseen scenarios." (Hussein et al., 2017)

In 2021 Kai Arulkumaran and Dan Lillrank investigated other learning techniques (not necessarily bio-inspired) to imitation learning, including adversarial and deep techniques, reinforcement learning techniques, and inverse reinforcement learning (Arulkumaran, 2021).

In the normal course of competence development, a human observing agent acquires competences for modeling its own body, senses and controls in early childhood. It acquires competences for its own perception and actions and for its own *abstract* actions, states, and goals. When the observing agent observes a demonstrator agent in action, it receives signals associating corresponding body parts and perhaps probable goals of the demonstrating agent. In broad strokes, imitation learning composes models from earlier competences to provide a rapid solution to the correspondence problem.

Table 2 summarizes relationships for the five steps for the correspondence problem to other competences.

| Correspondence Problem Step | Relation to Other Competences in Trajectory |
|---|---|
| 1. Acquire model of the world. | Perceiving, Understanding, and Manipulating Objects (4.1) Requires identifying those objects in the world that have goals and purposes of their own. |
| 2. Develop a model of its abstract goals and states. | Goals, Multi-Step Actions, Abstraction Discovery (4.2) <br> Curiosity and Intrinsic Motivation (4.3) |
| 3. Perceive and model the demonstrating agent's actions. | Requires **correspondence mapping** of the actions of the demonstrating agent's action to its own possible actions. |
| 4. Ascribe goals and purposes to the demonstrating agent. | When observed actions are like its own previous actions, a **correspondence mapping** to them (4.2) is required. <br> When observed actions are different from its previous actions, **correspondence mapping** to predictions by intrinsic curiosity (4.3) is required. |
| 5. Translate what it sees to what it can do. | (The overall correspondence problem.) |

*Table 2. Relating the Correspondence Problem to competences.*

In summary, imitation learning builds on previous competences and requires additional abilities. A required new competence is that the learning/observing agents need to notice and identify those objects in the world that have goals and purposes of their own. This identification builds on the curiosity and intrinsic motivation competence where observing agents notice agents that have unusual behaviors. For example, the anomalous agents move of their own volition rather than just by the laws of simple physics; they may make noises or talk; and they may appear to coordinate with other such agents. Learning agents also need to develop a theory of mind (teleological model) of the demonstrating agents. They need to make correspondence mappings between the demonstrating agent's actions and goals and their own possible actions and goals.

Drawing on neuroscience research on imitation learning for humans and animals, the sensorimotor systems in the brain provide hints or biases for solving the correspondence problem. Imitation learning is a special case of curiosity.

- *Imitation learning hypothesis.* Imitation learning acquires new competencies by observing the activities of other agents. The imitation learning system is slow and separate from the goal execution system and the motor execution system. Training involves observing task activity by demonstrator agents. Fast imitation learning requires rapidly finding solutions to the correspondence problem. A supporting and slower learning process is unconscious, incorporates high-level multi-dimensional data, does not directly set goals and does not do motor control. It informs the goal execution system about learned possible goals and mapping to concrete goals and actions.

## *4.5 Bootstrapping Imagination, Coordination, and Play*

When people coordinate with others, they model other people's goals and actions. They plan their own actions accordingly. They imagine what other people will do if they take one particular action versus another. This teleological modeling required for coordination is similar to the teleological modeling required for imitation learning.

Consider the following driving scenario for a human automobile driver (Bob). Although the scenario is written for a human driver, a similar scenario could apply for an AI agent of a self-driving car.

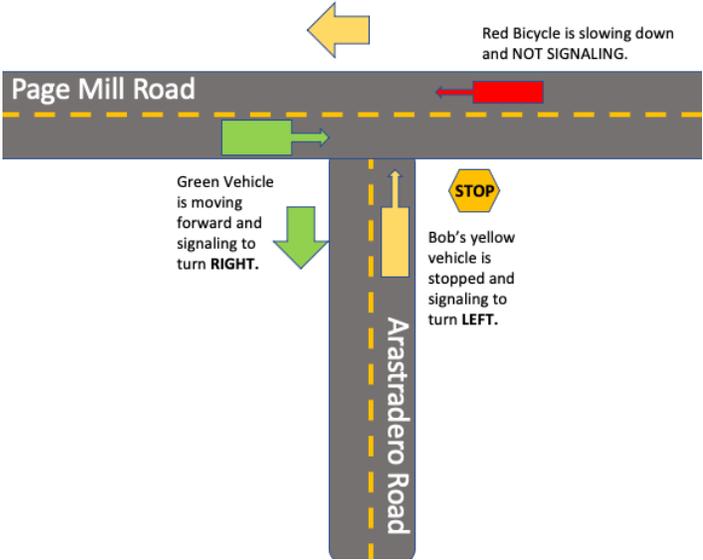

**Modeling the Goals of Other Drivers**

Bob was driving a yellow convertible on Arastradero Road approaching Page Mill Road. He depressed the brake and stopped as he approached a stop sign at the intersection.

A green vehicle on Page Mill Road approached from the left. It was slowing down and signaling that it would turn right on Arastradero Road.

Ahead and to the right of Bob, a red bicycle was slowing down. The bicycle rider was not signaling.

Bob had his foot on the brake pedal. The green vehicle on his left began to turn right onto Arastradero Road.

Bob considered what to do next. He could proceed ahead and turn left or he could continue to wait and see what the bicyclist does. If the bicyclist rode forward, Bob could safely turn left, keep left of the bicyclist, and proceed on Page Mill Road. If the bicyclist turned left onto Arastradero Road, however, Bob should not turn left. That action would be unsafe because Bob's car would cross the bicyclist's path. Bob was not yet sure what the bicyclist was going to do. Picking the safer option, Bob waited for the bicyclist to complete its turn. While Bob waited, and without looking at Bob or signaling, the bicyclist abruptly turned left and followed the green vehicle on Arastradero Road.

Drivers know that they need to plan for the possible actions of other drivers, bicyclists, and pedestrians. In this scenario, the driver of the green vehicle communicated with Bob by using a turn signal. To signal a turn, a bicyclist needs to lift a hand off the handlebars. Bicyclists sometimes give only a brief hand signal and then put both hands back on the handlebars to maintain their balance. The bicyclist might have signaled quickly when Bob was not looking.

In this example, the intentions of the bicyclist are ambiguous for Bob. The bicyclist could either (1) slow down but continue riding forward, or (2) turn left and pass in front of Bob's vehicle. To avoid a potential collision, Bob stops and watches the bicyclist.

A theory of mind hypothesis is that people imagine themselves in the situations of other agents and simulate the choices that other agents could make. Making predictions about others requires cognitive processes similar to those that people use in making choices for their own actions. They imagine what they would do if they had the same goals and the same options as another person.

In summary, human children gain early experience interacting with others in their second and third year. In play and other life situations they increase their competences for understanding the goals of others and predicting what they will do.

*4.5.1 Research about Stages of Children's Play*

Participating in play activities with more than one person requires modeling the goals of others and modeling the collective goals and activities of a group. Play is a setting for practicing coordination in sporting events and other activities through the school years and beyond. When toddlers play, they acquire self-developed and socially developed competences for coordination.

Mildred Parten systematically studied social participation within groups of children in the Nursery School of the Institute of Child Welfare at the University of Minnesota (Parten, 1933). She organized systematic observational studies in six stages of play as shown in Figure 14.[43]

| Birth to 2 years | **Unoccupied Play.** Children are relatively still, observing materials around them without much organization. |
|---|---|
| 2 years | **Solitary Play**. Children busy themselves manipulating materials, observing how the world works, and mastering control of their movements. They do not seem to notice or acknowledge other children. |
| 2 ½ to 3 ½ years | **Onlooker Play**. Children watch other children play but do not join them. They may engage in some communication with other children. |
| 2+ years | **Parallel Play.** Children play separately from others. They are close to them and begin to mimic their actions. |
| 3-4 years | **Associative Play**. They practice what they are seeing and may coordinate their activities. |
| 4+ years | **Cooperative Play**. Children coordinate their activities and take on roles in joint tasks with shared goals. |

*Figure 14. Parten's Stages of Play.*

Parten studied how children advance during their first four years. Observers found only the youngest children in the unoccupied stage. As children mature, they take more interest in other children, spend increasing time in solitary play, then onlooker play, parallel play, and so on.

Researchers often reference Parten's stages (e.g., Drew, 2021). For example, during unoccupied play they begin to move their bodies with apparent purpose. During onlooker play children mimic the behavior of other children. During associative play and cooperative play, they develop communication skills and engage more with other children.

---

[43] Claire Hughes and Judy Dunn caution that Parten's research is sometimes interpreted as meaning that three year olds are solitary and do not interact with other children (Hughes & Dunn, 2007). Subsequent research interprets Parten's stages as referring to the form of interactions among children. The stage descriptions do not account for children's interactions through gestures and other nonverbal communications.

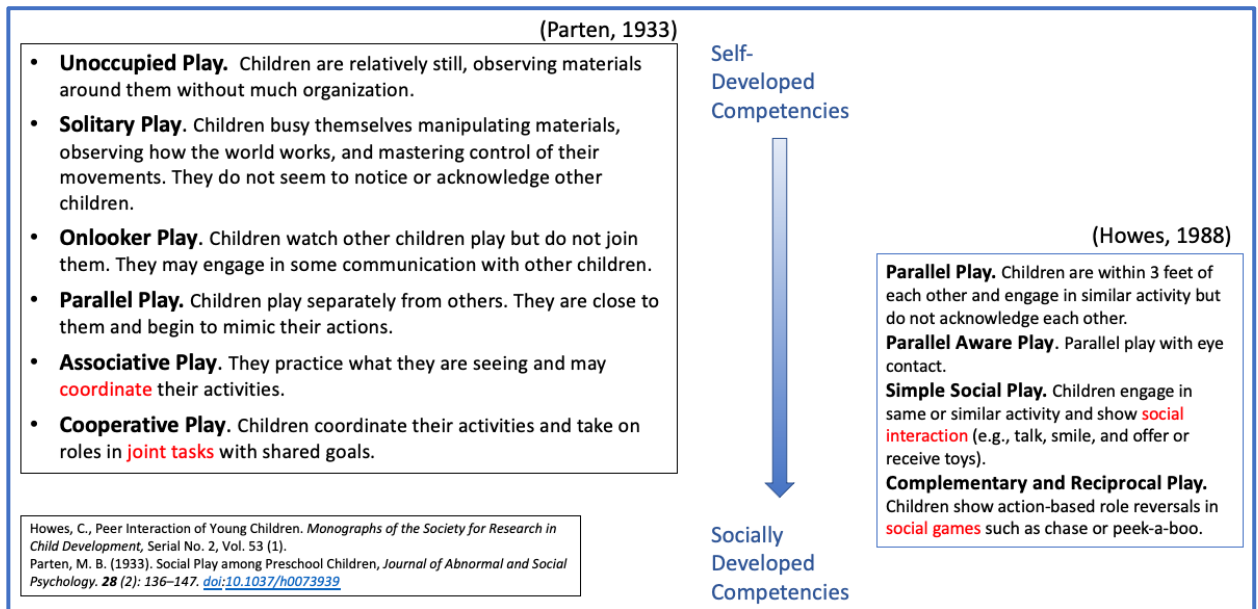

*Figure 15. Across multiple studies, later stages of play increasingly support socially developed competences.*

Figure 15 shows a slightly revised description of stages from more recent studies (Howes, 1988). The shading of the downwards arrow suggests how the later play stages increasingly involve social interactions. Many competences involved in play activities enabling social behavior and social learning develop rapidly during the toddler period and early childhood (e.g., Hay et al., 2007).

*4.5.2 Research about Pretend Play, Imagination, and Coordination*

Angeline Lillard describes *pretend play* as a kind of play that transforms the here and now.

> "… a child might pretend that a stick is a horse, and gallop it around, or that he himself is a king, or that a group of friends are various family members engaged in a pretend fair. Pretending can involve the substitution of one object for another, the attribution of pretend properties, and even the conjuring up of imaginary objects. … It always involves the intentional projection of a mental representation on to some reality. … Pretend play has been found in every culture in which it has been sought … and it emerges on a similar timetable. … Pretending also is entwined with early language. In sum, excepting for autistic children, pretend play is a universal of early childhood." (Lillard, 2007)

Consider the following scenario involving a five-year-old and his grandfather.[44]

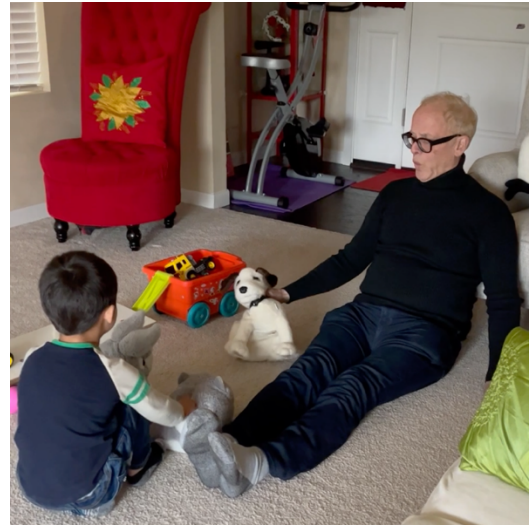

**Dogs Can't Fly**

Macer and his grandfather are at Macer's home. Macer has a toy stuffed hedgehog named Fluffy and a toy stuffed rabbit named Bunny. A toy stuffed penguin named Penguin is on a chair behind Macer's grandfather. He holds a black and white toy stuffed dog named Dog. He has just sat down.

**Fluffy** (*voiced by Macer*)**:** "Let's race again."

**Dog** (*voiced by Macer's grandfather*)**:** "I'm resting. Besides, you fly. You always win."

**Fluffy**: "That's because you cheat. You don't touch all the [designated] places.

**Dog**: "I don't have wings. You and Bunny can fly, which is strange. Only Penguin has wings, and even he doesn't fly much."

**Macer**: "Penguin is dead. He was bad."

**Dog**: "Grandpa told Penguin to behave. He is just resting. I don't have wings."

**Macer** (sounding exasperated): "C'mon!"

Macer jumps up with Fluffy and Bunny. His toy stuffed animals start to race around the room.

Grandpa gets up. Animating Dog to run on the floor, he takes big steps chasing Macer. Dog pulls ahead of Bunny and Fluffy.

Macer giggles and picks up his pace. He starts to run and flies Fluffy and Bunny. On the next lap Fluffy and Bunny bounce off a toy box, the fireplace, and then bounce off Grandma in a chair.

**Dog**: (growls)

Dog runs faster.

Macer throws Fluffy and Bunny at the finish chair.

**Fluffy and Bunny**: "We win!"

In the *Dogs Can't Fly* scenario, Macer and his grandfather give voice to stuffed animals. They pretend that the toys are alive, have goals, and have a conversation about racing around the room. Participating in the pretend play, Macer speaks for two of the stuffed animals. He imagines what each animal wants and what it will say in the racing game using his game rules. The conversation moves back and forth between referencing the real world and staying within an imagined reality. Dog argues that he has no wings and cannot fly. (Grandpa may not want to up from the floor). Fluffy, perhaps reflecting Macer's enjoyment of the game, wants to race again.

Each person creates and maintains mental models of himself, the other person, and imaginary mental states of the toy stuffed animals. Such play exercises similar teleological modeling to what is needed in real world situations of human coordination.

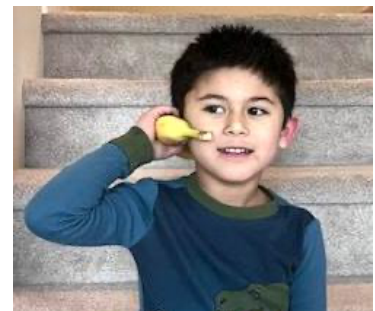

*Figure 16. Child pretending a banana is a phone.*

In pretend play, object substitution is where people pretend that an available object in the real world is something else. Substituting a banana for a phone, the child in Figure 16 pretends to have a conversation with a person who is not present or does not exist. In pretend play, there is an agreement among parties to act as if the substituted object is a different real object, and to act accordingly. This agreement enables interaction in a rich pretend world. From a perspective of cognitive modeling, the popular use of object substitution suggests that young children can easily carry out flexible symbol-like reasoning about object identity.

---

[44] The *Dogs Can't Fly* scenario is based on multiple real life episodes as shown in the photo.

*4.5.3 AI Systems for Imagination, Coordination, and Play*

Coordination has been studied in several fields and contexts. A few examples provide an orientation. Thomas Malone and his colleagues wrote about theories of coordination in the contexts of organizational management structures (e.g., hierarchies versus markets) (Olson, et al., 2012; Malone et al.,1993; Malone et al., 1987a) and intelligent information sharing (Malone et al., 1987b; Malone et al., 1986). Reid Smith and colleagues wrote about coordination in multi-agent systems as distributed problem solving (e.g., Smith, 1980). RoboCup soccer is an AI context and contest for research on team coordination challenges in games with robot players. A specialized variation of team coordination in command and control of swarms of autonomous systems (e.g., Chung et al., 2018). This early AI research on AI coordination does not discuss the *acquisition of competences* for coordination.

In support of their experiments, researchers in developmental robotics have created child-sized humanoid robots such as the iCub (Metta et al., 2008) and the $CB_2$ (Minato et al., 2007). Because the research involved only an individual robot playing, it modeled curiosity, but not coordination. For example, in 2005, Pierre-Yves Oudeyer and his colleagues conducted experiments with an iCub *playing alone with blocks* in a setting with toys (Oudeyer et al., 2005). as platforms for exploring experiments in robotic cognition, perception, and language use (e.g., Cangelosi et al., 2015; Asada et al., 2009).

Further research would explore the synergy and co-development of related competencies and play patterns when AIs engage in play with other robots or humans. Next research steps need to experiment with incorporating the method generally as part of an *observe-predict-evaluate-act-learn* cycle for AIs in play groups.

In summary, human children gain early experience interacting with others in their second and third year. Play experiences provide a rich setting for deepening teleological and social models.

- *Learning for Imagination, Coordination, and Play hypotheses*. The learning system involved in play and coordination is slow and separate from the goal execution system and the motor execution system. Training involves observing and learning in interactive activity with other agents. The learning process is unconscious, incorporates high-level multi-dimensional data, does not directly set goals and does not do motor control. It modifies its model of the other agent learning to predict what the other agent will do. If the actions by the other agent are novel or unexpected, and effect the agent's plan, it informs the execution system with updates for acting and predicting outcomes of action.

*4.6 Bootstrapping Communication and Language*

Communicating in natural language is the main way that people share information and coordinate their joint activities.

To interact effectively with people about communicating and collaborating, future AI systems should be fluent in natural language. Embodied AIs (robots) will be expected to communicate not only in speech, but also with gestures and by reading and writing. They will need to use natural language to interact with and reliably learn from digital and other media.

As discussed in a companion paper (Stefik, 2023a), communication and collaboration require many of the same competences. Communication is a prerequisite for several of the ten challenges of human-machine teaming (Klein et al., 2004) including maintaining common ground, collaborative planning, and negotiating goals.

*4.6.1 Three Generations of NLP Technology*

> **Three Generations of NLP Technology**
>
> - The **first generation** of NLP technology is *symbolic* NLP. It uses finite state and other grammar representations and parsers (e.g., Joshi, 1991; Jones, 2001). The content resources for this approach include grammars, parsers, and organized networks of words and concepts (e.g., WordNet). The content resources are created manually. Applications of symbolic NLP include information retrieval, text summarization, spelling correction, parsing, parts of speech tagging, entity extraction, question answering, and sentiment analysis. In interactive applications, the systems are programmed to respond narrowly to queries, but do not otherwise acquire rich world models. The technology is used for tools for linguists, chatbots, translation assistance, and other areas. Research included structured communication models for discourse and narrative (e.g., Grosz, 2018).
>
> - The **second generation** of NLP technology uses *statistical* approaches (e.g., Manning and Schütze, 1999). Applications included information and document retrieval, text analytics, topic analysis, and machine translation. Content resources for this approach included corpora with thousands of documents that enabled automatic development of baseline probabilities for word frequencies and test sets for comparing the performance of algorithms (e.g., for document retrieval and topic analysis).
>
>   - For several decades, systems for classifying, retrieving and extracting have been developed for information analysts. They automated repetitive text processing, freeing up time for analysts and others to focus their attention and effort where an understanding of the meaning of documents is required. These applications typically combine first and second-generation NLP technology.
>
> - The **third generation** of NLP technology uses *deep learning* approaches including ones with transformers and transfer learning (see Otter et al., 2021). The main content resources produced by this research are pre-trained large language models (LLMs) such as BERT, GPT-4, and MT-NLG (Devlin, et al., 2019; Bonmasani et al., 2021). This generation has had notable successes including performance improvements in natural language translation (Lecun et al., 2015) and text generation. It surpasses the performance of earlier approaches for benchmark linguistic tasks including paraphrasing, part-of-speech tagging, named entity recognition, and question answering. The third generation is evolving rapidly. Some versions of it are discussed in recent textbooks on natural language processing (e.g., Azunre, 2021).

The sidebar describes three generations of computer technology for natural language processing (NLP) that have been developed and applied over several decades. Figure 17 summarizes examples of their linguistic services, commercial applications, and associated linguistic resources.

Despite their applications and many successes, the three generations of NLP technology have not enabled AIs to converse fluently with people for coordination and other social interactions. Although AI and linguistics researchers have long recognized *language acquisition* as a research challenge (e.g., Chomsky, 2023; Bisk et al., 2020; Tellex et al., 2020; Tamari et al., 2020; Marcus &

| NLP Models | Example Service | Commercial Applications | Required Content Resources |
|---|---|---|---|
| Symbolic NLP | Given a sequence of words, is it a valid sentence? | • Spelling & grammar correction<br>• Spam filtering<br>• User interfaces for retrieving structured Information from databases | • Professional linguist-generated grammars<br>• Ontologies with thousands of concepts |
| Statistical NLP | Which articles in a corpus are similar to a given article? | • Text-based Search<br>• Topic identification | Corpus repositories with thousands of documents<br>Parallel corpuses with parallel sentences in multiple languages |
| Generative Large Language Models | Given a word (or sentence), what is a likely next word (or sentence)? | • Machine translation<br>• Emerging chatbot applications<br>• Essay and code generation<br>• Question answering | Corpus repositories with hundreds of billions of words<br>Context tuning models and "prompt generators" for every application. |

*Figure 17. Three generations of natural language processing (NLP) technology.*

Davis, 2019; Maurtua et al., 2017; Collobert et al., 2011; Cangelosi et al., 2015; Bloom, 2000; Harnad, 1999), it has not been a focus of the mainstream of this research.

### 4.6.2 Research on Communication and Language

Several animal species are known to communicate with language (e.g., Bakker, 2022). Using audio recorders and machine learning technology, researchers over the last few decades have shown that other animal species communicate to coordinate their joint activities and to teach their young. What a species communicates about depends on its communication needs for its joint activities. Animal species employ gestures and use language to warn of dangers and to coordinate hunting and foraging. As discussed in later sections, babies use gestures and nonverbal communications before they learn to speak.

For some species (e.g., whales and bees), researchers have created technology that enables them to understand some of what animals are saying and to communicate with them. Non-human communication sometimes uses audio frequencies outside the range of human hearing.

Research on communication and language in different species brings a wider lens to understanding the utility and development of commuication and language. Evolution favors the development of communication competences when they improve fitness. Communication enables members of a species to share information and to coordinate their activities.

Animal groups expand and change their languages to meet changing needs. Over time, separated groups develop dialects. Dialects help members of a group to recognize when an individual is a member of their group. New members joining a group (or being born to it) must learn the group dialect and keep learning as it changes. The changing nature of dialects demonstrates that language requires learning.[45]

---

[45] Researchers have long debated the extent to which language competences are innate. There is now little doubt that language competences are the result of innate, self-developed, and socially developed acquisition. Characterizing the contributions of each kind of acquisition remains a subject of ongoing research.

Based on physical evidence from the fossil record and genetic evidence, human language appeared about 50,000 years ago. This is much later than when early humans evolved from their early hominid ancestors approximately 200,000-300,000 years ago.[46]

The development and use of human language correlates with changes in human behavior reflecting advances in human abilities to coordinate and collaborate (e.g., Tomasello, 2019). As Eric Horvitz reflected,

> "… the first recorded tool making by an ancestor of homo sapiens – home habilis – [lived] in the Olduvai gorge in Africa … Homo habilis was two million years ago … [Later,] humans [lived] in the caves of Spain and France, … around 400,000 years ago to about 30,000 years ago. And the most amazing thing … is a finding that the architecture of homo sapiens – our mind and body plan -- have not changed in any significant way for 400,000 years. But something very mysterious happened around 40,000 to 45,000 years ago that changed everything about who we are and led to civilization. It's called … the symbolic explosion. … leading to homo sapiens that are so different – buildings, structures, civilization as we know it, technology, deep understandings … What was the mystery? … We are different, and we continue to be different … in that we have evolved ourselves through the tools of our imagination … starting with language. Language is a human invention. … language led to collaboration, specialization, and an accumulative society that could build on itself over time." (Landay, J., et al., 2023)

Researchers hypothesize that human natural language provided fitness advantages for coordinating hunting, food gathering, agriculture, and other team and community functions (e.g., Pinker and Bloom, 1990). Competences for gestures and primitive sounds preceded structured speech and the development of complex language.

Facility in language also enables people to draw on oral and written knowledge that is socially developed and accumulated over multiple lifetimes. Today, people uniquely share knowledge geographically in public and online libraries and information resources.[47]

Language competence, knowledge of the world, and skill at social interactions all expand enormously during childhood. In their second year, most human children can ask for help and learn about the world. They learn that people are sometimes helpful, sometimes don't quite understand, and sometimes can be joking or teasing.

As Celia Brownell and Claire Kopp put it,

> "… it is important to keep in mind that however much the toddler has raced forward in development, much lies ahead. True, she has put infancy behind her with first words, first steps, first use of cultural tools like spoons and crayons, first evidence of self-recognition, and first instances of pretense and peer play. But she remains immature along many dimensions that characterize the more advanced preschool repertoire, including autobiographical memory, planful problem solving, flexible use of emotional and behavioral regulatory strategies, conversational fluency and narrative, and more abstract perspective on self and other. It is unlikely that a toddler could converse with the sophistication, yet charming naiveté of a 4-year-old who asked her mother, "Do you think tooth fairies are

---

[46] A comprehensive discussion of the origins and development of human language is beyond the scope of this paper. Interested readers can find perspectives of researchers from several fields in (Christensen et al., 2003).

[47] In addition to evolutionary perspectives, a social principle in democratic societies is that sharing knowledge in public libraries and in online resources has social value. For example, it has long been argued that access to reliable information helps people to keep informed, to avoid misinformation, and to discover opportunities (e.g., Wikipedia, 2023b).

smart? If I put this pebble under my pillow, will the tooth fairy leave money?" (Brownell & Kopp, 2007a)

How do children come to understand the production and meaning of natural language so well? How can they learn so much with just a few observations, that is, with so little data? How do they learn it so quickly? Joshua Tenenbaum and his colleagues frame these questions as follows:

> "Consider the situation of a child learning the meanings of words. Any parent knows, and scientists have confirmed … that typical 2-year-olds can learn how to use a new word such as "horse" or "hairbrush" from seeing just a few examples. We know they grasp the meaning, not just the sound, because they generalize: They use the word appropriately (if not always perfectly) in new situations. Viewed as a computation on sensory input data, this is a remarkable feat." (Tenenbaum et al., 2011)

How do children do this?

Researchers distinguish two broad areas of language acquisition: learning to communicate and learning language. Erica Hoff characterizes these capabilities as follows:

> "*Linguistic competence* is the ability to produce and understand well-formed meaningful sentences. *Communicative competence* is the ability to use those sentences appropriately in communicative interaction." [emphasis added] (Hoff, 2014)

### 4.6.2.1 Human Acquisition of *Communication Competences*

*Communication competences* include using gestures and language to communicate. Using verbal and nonverbal means, adults communicate with purpose – such as to make a request, to promise something, or to agree or disagree with others.

Figure 18 milestones in children's communication development, starting with nonverbal communications.

Before babies can talk, they communicate with gestures including facial expressions, gaze, and pointing.

Figure 2's scenario *Opening a Door for Dad*

| Age | Communication Competence Developments |
|---|---|
| 0-10 months | Social interactions involving smiling, gesturing, and making sounds. Observing people saying things and responding to what is said in social situations. |
| 10-12 months | Children have communicative goals but do not use adult like language forms. Coordinated pointing to objects and babbling. |
| 12+ months | Children have goals and communicate them using adult like language forms. |
| 12 – 24 months | Speech acts have just 1 word. Example functions:<br>• greet,<br>• call for attention,<br>• request to continue an activity,<br>• propose ending an activity, … |
| 2 - 3 years | • Children take turns speaking.<br>• They may initiate topics.<br>• 23% of conversations include narrative. The average conversation has 1.7 clauses.<br>• A narrative may be a few unrelated ideas with no coherent links.<br>• Narratives tend to be a basic description with no causal or time links. |
| 3 – 4 years | • Stories contain a central character, topic, or setting.<br>• Stories may have structure – initial event, actions, consequences.<br>• Joining words may connect story elements. |
| 4.5 – 5 years | • Stories have a logical sequence of events.<br>• Stories show some cause and effect or temporal ordering. |
| 5 years | • 35% of conversations include narratives.<br>• The average conversation has 2.8 clauses. |

*Figure 18. Milestones in communication development over the first five years. This table is adopted from information in (Stanford, 2023; Spelke, 2022; Hoff, 2013; Brownell & Kopp, 2007; and Tomasello, 2001).*

gives examples of nonverbal communication between a toddler, Macer, and his father, Nick. Macer had been watching his father struggle to hold a stack of packages as he tried to open a closet door.

Nick made eye contact with Macer. Macer understood that his dad wanted help. He got up and opened the closet door. Such situations are familiar to parents and caregivers of toddlers.

For another example of early nonverbal communication, a child may point to an object and look to a parent. The "looking" is the child checking for evidence that the parent saw the pointing gesture and acknowledges the communications act. Children may make a fuss when they try to get something that is out of reach. In this way, children use pointing as a deliberate communication act.

Sotaro Kita's book *Pointing: Where Language, Culture, and Cognition Meet* reviews research on communicating by pointing a finger (Kita, 2003).[48]

> "Pointing has captured the interest of scholars from different disciplines who study communication: linguists, semioticians, psychologists, anthropologists, and primatologists. … The prototypical pointing gesture is a communicative body movement that projects a vector from a body part. This vector indicates a certain diretion, location, or object. … It is a foundational building block of human communication." (Kito, 2003).

Nameera Akkhtar and Carmen Martínez-Sussman summarized research findings about gazing and pointing by infants:

> "Perhaps the most studied gesture is the point – the extension of the index finger to an object or event – that emerges at the end of the first year. Early studies (e.g., Bates et al., 1975) described two distinct functions of 1-year-old infants' points: imperative (instrumetal pointing to request the object pointed at) and declarative (pointing to share interest in the object or event pointed to) … gaze may often be used [as a communication signal] without pointing … By 12 months babies can reliably follow an experimenter's gaze to an object and will spend more time inspecting that object than if the experimenter had turned toward it with her eyes closed." (Akhtar & Martínez-Sussman, 2007).

As described by Sotaro Kita,

> "Pointing is one of the first versatile communication devices than an infant acquires. Pointing emerges out of its antecedent behaviors, such as an undirected extension of the index finger, several weeks before the first spoken word." (Kita, 2003)

George Butterworth summarized research on how pointing serves infant communication:

> "… at 10 months babies sometimes point at an object, then turn to the mother as if to check with her, whereupon they point at the mother. …Checking is strong evidence both for communicative intent and for the deictic nature of the gesture because the audience is being 'interrogated' for comprehension of the referent. Thus, it is possible that components of pointing which are particularly closely related to syllabic vocalization, can be observed early

---

[48] A thorough review of research on nonverbal communication and the creation of theories of mind by children is beyond the scope of this paper. A few examples may be helpful for interested readers. Jean Piaget did extensive early observational research on infant cognitive development in several books including *The Construction of Reality in the Child* (Piaget – translated to English by Margaret Cook, 1954). Elizabeth Bates, Laura Benigni, Inge Bretherton, Luigi Camaioni, and Virginia Volterra's book *The Emergence of Symbols: Cognition and Communication in Infancy* reports on research from several perspectives about cognition, communication, and language (Bates et al., 1979). Douglas Frye and Chris Moore's book *Children's Theories of Mind: Mental States and Social Understanding* reviews research on the acquisition of theories of mind starting with mental models of objects that move with volition and leading to theories about how children build mental models of others (Frye et al., 1991). In his paper "Theories of Mind in Infancy," Chris Moore reflects on different kinds of evidence for theories of mind (Moore, 1996). In this position paper on bootstrapping developmental AIs, our purpose is more about using such theories to inspire analogous bootstrapping trajectories for AIs. An experimental framework would compare an AI's micro-abilities in experiments to children's micro-abilities. Restated, our AI interest is more about finding ways to test whether the AIs are advancing and acquiring prerequisite sub-competences rather arguing the merits of competing models of neural computation or human cognition.

in development. … Pointing occurred only under conditions where a partner was available for communication." (Butterworth, 2007)

In summary, cognitive development and model building precede and accompany language development. Psychologists have long realized that children's language learning takes place in a context of extensive individual and social learning (e.g., Lindsay & Norman, 1977). Michael Tomasello characterizes the preparatory foundations required for human language acquisition as follows:

> "… to acquire linguistic conventions in the situations in which they encounter them, young children must have … powerful skills of social cognition. … the key social cognitive skill is children's ability to perceive the intentions of the adult as she acts and speaks to the child. … the understanding that other persons have intentions towards my intention states – is the very foundation on which language acquisition is built." (Tomasello, 2001).

Reflecting on cognitive theories that children have "folk theories of mind,"[49] Andrew Meltzoff suggested:

> "Failure to attribute mental states to people confronts one with a bewildering series of movements, a jumble of behavior that is difficult to predict and even harder to explain. … Normal children give elborate verbal descriptions of the unobservable psychological sates of people, indicating that they relate observable actions to underlying mental states. … One has to understand that there is not a single 'theory of mind,' but rather a succession of different theories, in particlar, an early mentalistic one." (Meltzoff, 1995)

Meltzoff designed experiments to test whether eighteen-month-old children understand the intentions of others (Meltzoff, 1995).[50] In one experiment children were asked to re-enact what an adult tried to do but failed to complete successfully. He found that children could understand an adult's intended acts by watching his or her failed attempts. Restated, they demonstrated their understanding of the adult's goals even though the adult failed to complete a task.

In the context of a competence trajectory, the *discovery of communication* involves both abstraction discovery and goal discovery as described in previous sections. As toddlers learn that others have goals, they learn that they can get them to do things if they say the right words. They build up experience about what people do, what they want to do, and where they might need help. Before ten months, children's vocal behavior has effects, but it is not done with clear intent to communicate. In later months, children recognize that their sound making has effects, but they have not yet learned proper linguistic forms. Still later, they learn to express themselves using linguistic forms that adults can interpret.

Claire Kopp summarizes developments in perception, pointing, babbling, and social interactions over the first few months.

---

[49] A detailed discussion of competing interpretations of experimental observations about infant communication is beyond the scope of this paper. Elizabeth Bates, Luigia Camaioni, and Virginia Volterra discuss the onset of *communication before speech* begins (Bates, Camaioni & Volterra, 1975). They review research on the cognitive and social developments that prepare children for *discovering communication*.

[50] As described in Section 4.4 on imitation learning by mature humans, ascribing goals and purpose to the actions of a demonstrator is one of the required competences for imitation learning and a step of process of solving the *correspondence problem*. Meltzoff's research as described briefly in this section is about a child's early ability to ascribe mental states to another person who has demonstrated an activity or even failed part way in demonstrating an activity. This research is consistent with the theme that prerequisite elements of adult-level competences are developed in parallel by very young children.

> "Toward the end of the first year, social acts often involve a coordinated sequence of behaviors: a baby turns to his father, pulls on his arm, babbles, points to a toy, and then looks back at his father. This group of behaviors … [is] often labeled joint attention… the baby invites an adult to share his interests and his world! … The exchanges can take the form of eye contact, vocal communications, or emotional mimicry (smiling to one another). … Being able to initiate social interactions also delights them, and they try to keep these exchanges going." (Kopp, 2013)

The study of communication acts including speech analyzes the competences that are required to create and receive them. Erika Hoff offers an example of a speech act where a fictional character, Dennis the Menace, goes to a neighbor's door and says, "My mother wants to borrow a cup of ice cream."

By way of illustration, Hoff analyzes this fictional communications act as follows:

> "The intended function or illocutionary force of Dennis' speech act is to request, while the form, or location is a declarative sentence. … Dennis's communicative intention was not just to get ice cream, … Rather, the intention was … to create a belief in the listener's mind …" (Hoff, 2014)

The teleological model that the Dennis the Menace character has of Mrs. Wilson is inadequate to fool her. The example shows that understanding speech acts requires having nuanced mental models of others.

What matters in communication when people teach each other? How can communication skills improve a learner's efficiency in learning competences?

Paul Shafto, Noah Goodman, and Michael Frank investigated this question (Shafto, et al., 2012). Their perspective drew on concepts from Herbert Paul Grice, who described maxims for effective communication (Grice, 1975).

Here are examples of Grice's maxims:

- Make your contribution as informative as is required for the purposes of the exchange.
- Do not make you contribution more informative than is required.
- Do not say what you believe is false.
- Avoid obscurity of expression.
  - … from (Grice, 1975)

A learner learns more effectively by taking into account the goals of a teacher. The learner recognizes that the teacher's goal is to provide information efficiently, adhering to Grice's maxims. The learner then takes into account that the information conveyed by the teacher is exactly the information that is required specifically for the purpose. This focuses attention and can disambiguate among the possible meanings of what a teacher says.

Generally, a teacher should include all important information that is not common sense. If the instruction is intended to prepare the student for multiple situations, a teacher should explain the variation in the situations and proper actions. If it turns out later that a teacher leaves out important matters or conveys incorrect information, then the student may decide that what the teacher says needs to be tested further and verified.

The authors illustrate these points with scenarios and Bayesian analyses. To provide a baseline of communications and learning, Shafto et al. compare learning efficiency where (1) an agent learns by itself experimentally using physical evidence, (2) where it learns by imitation learning from goal-directed demonstrations, and (3) where it learns from communicated actions. For example, a learner proceeding entirely by experimentation would need to include extra steps in a multi-step activity or leave out necessary conditional tests that are required in variations of the situation. Their research shows that under Gricean-like assumptions, an agent can correctly learn and generalize cause-and-effect examples with fewer communications and examples.

As their conclusions summarize:

> "Our approach suggests that intuitive psychological reasoning may provide a strong lever by which learners can capitalize on other's knowledge to learn about the world. … If other people are viewed to act in random or even malicious ways, then learning on the basis of their actions will likely be very difficult. However, if people are viewed as approximately rational, goal-directed agents or as knowledgeable and helpful teachers, then the problem of learning becomes much more tractable. These assumptions make it possible for learners to learn [efficiently] what others already know, rather than rediscovering all knowledge form the ground up." (Shafto et al., 2012)

In summary, communication with gestures precedes speech. Competences acquired in early childhood, such as competences for perception and motor control and competences for multi-step actions, goals, and abstraction discovery, are the basis of a child's model of its own body and mind. As they advance in their early years, they notice that some "objects" in their environment do not simply move following like solid and rigid objects. For example, some objects have separately moving parts like arms and legs. Some objects are apparently "other" people and their behavior can be understood as having goals, senses, and motor control just like the observing child.

Children notice that their parents look at things before they reach for them and that they make notices and use gestures to direct attention. With these observations children build simple models of what other people do and begin to communicate with gestures before they learn to communicate with speech. In short, at the time when they learn to communicate with speech like other people, they already have things that they want to communicate about and have observed how other people communicate.

For the most part, the study of communication including nonverbal communication has been a topic in psychology and developmental psychology, but not a mainstream topic in AI or computational linguistics. The next section considers competences in the trajectory where children acquire linguistic competences.

### 4.6.2.2 Human Acquisition of *Linguistic Competences*

We begin with human linguistic competence development for hearing, sound production, and speech.[51]

| Age | Speech and Language Developments | Notes |
|---|---|---|
| 0-3 months | **Phonology, Prosody, and Intonation** | Infants perceive some phonological contrasts, "bat" versus "pat"; or "bat" versus "bad". They become sensitive to language prosody and temporal variations in intonation that reveal boundaries of phrases and emotional emphasis. |
| 3-6 months | Name recognition. Marginal babbling. | Infants attend to repetitions of their own name. Vocal play with sounds. E.g., cooing, squeals. |
| 6-9 months | Canonical babbling and Words | Language-like syllabic sounds such as "dada" or "bababa". Infants recognize words learned in isolation. They begin to connect words to their referents. Infants learn the abstract ordering of function and content words. They learn that unction words like "the" and you" form heads of phrases. They learn that the complements of "the" are nouns and the complements of "you" are verbs. |
| 10-12 months | **Grammar and Referents**. Participating in babble-gesture combinations. | Observing people saying things and responding to what is said in social situations. Infants identify with confidence the sounds of their language, the grammatical class of each word, the role each word plays in a sentence, and connections between words and their referents. Infants expect that speech will be efficient and relevant to the situation that the speaker and child communicate about. |
| 12-18 months | 50+ word vocabulary | Using gestures to answer questions and to ask for help. Children ask about the names of things – "What's that?" They help others in simple ways. They see others using language to get help. |
| 18-24 months | **Telegraphic speech.** Kids use two-word combinations like "more milk" or "mommy's hat", "Car go," "Daddy work." This is called telegraphic speech. | Typically, the words are in correct grammatical order but connecting words are missing. Kids might use multiple 2-word sentences in a short narrative. |
| 24 months | **Longer sentences**. Producing two or three-word sentences such as "Me tired" and "Toy go there." | Producing successive sentences to communicate about events or causes. |
| 36 months | 1000+ word vocabulary. Using more adultlike grammar. Understanding and participating in speech acts. | Turn taking in games and in speech. When a parent says "You left toys all over the floor" the child may start to pick them up and put them away. By 36 months, children begin to bargain in earnest with their parents and others. In this period, speech engagement with other children occurs increasingly in play and preschool settings. |
| 48 months – 60 months | More consistently handling special cases for regular and irregular verbs, tenses, and plurals. 10,000+ word vocabulary. | Being more mindful of the feelings of family members and friends. Having better understanding of time. |

*Figure 19. Timeline of typical milestones of children's speech and language development over the first five years. This table is adopted from information in (Stanford, 2023; Spelke, 2022; Hoff, 2013; Brownell & Kopp, 2007; and Tomasello, 2001).*

*Speech* and *word usage* are actions that parents easily observe and where research has historically focused. Figure 19 presents familiar *speech milestones* for children during their first five years.[52] This table is about vocabulary size and observable changes in the intelligibility of speech. Although these developments are easily observable, focusing on speech and vocabulary leads to an exaggerated "big bang" impression of language acquisition.

Children's hearing is functional before birth. Research shows that babies in the womb give different responses to different musical sounds.[53] Newborns also differentiate human voice sounds from other sounds. Soon they begin to see and hear their caregivers clearly. By the first month, a baby's hearing responds to pitch – more to high-pitched sounds like a mother's voice than to lower-pitched sounds.

---

[51] As with the study of neural processing for other competences, neuroscience research on language processing and language acquisition augments behavioral studies. As anticipated by Michael Arbib and Giacomo Rizzolatti (Arbib & Rizzolatti, 1996; Rizzolatti & Arbib, 1998), the areas in the sensorimotor brain maps that connect to body parts (e.g., to hands, feet, and the mouth) for sensing, motor control, and imitation learning are also involved in language and speech. For example, action verbs in language involving (say) the hands activate regions or adjacent regions to the sensorimotor maps for the corresponding body parts (e.g., Hauk et al, 2004; Glenberg et al., 2002). As such, meaning connects to action. Neuroscience research in this area has mostly been limited to the study of single words and isolated sentences. Broader neuroscience investigations of human natural language processing in speech, dialog, and reading are challenging to carry out due to the large number of experimental variables to be controlled.

[52] The timing of speech milestones varies for different children. Studies report variations of 1-2 months in the centerline for phenomena, although there is much agreement on the order that they appear.

[53] Spelke reviews research about what infants learn in the womb about the properties of their local languages (Spelke, 2022).

Months before parents observe early infant speech, babies watch other people communicate and start to learn features and elements of their language. Neuroscience research shows that they start with classifying the sounds in their local language.[54] In the first three months, they begin to recognize different sounds, rhythms of speech, and intonation. They begin to recognize words spoken in isolation.

By ten months children understand what some of the words refer to. They watch people, point to objects, notice what is happening and what people are talking about, and are taught by others. They distinguish between content words (like nouns and verbs) and function words (like prepositions).

By twelve months children can identify with confidence the typical sounds of the local language. At eighteen months, they start making their own sounds approximating speech.

Mothers and other caregivers first speak to babies in *motherese*.[55] Motherese is an exaggerated form of speech used to help babies learn language. It involves speaking slowly and in a higher than normal pitch, talking about things they can see together, repeating phrases, and exaggerating intonation and pronunciation. This practice directs an infant's attention to salient elements of speech. Motherese is spoken around the world by parents speaking different languages.[56]

In the first weeks, babies' larynxes become more flexible enabling them to make sounds approaching human speech. They learn to make speech-like sounds as they engage in social exchanges – cooing and smiling. By six weeks some of a baby's vocal sounds are recognizable – like "ah," "eh," and "uh." Production of these sounds is called *marginal babbling*. These sounds are at first accidentally emitted.

Speech babbling follows the same development pattern as "motor babbling," where children observe their own unintended actions and then learn from them. They learn to recognize different sounds, how to produce them, and what their effects are. Over the next three months marginal babbling changes to canonical babbling, where babies make repeated syllabic sounds like "dada" and "bababa."

Linguistic competences include syntax, morphology (the forms of words), regular and irregular verb forms, overregularization and overgeneralization, categories in grammars, parts of speech, popular language versus formal language, the innateness of grammar, similarities and differences across human languages. Research overviews of linguistic competences are available in (Spelke, 2022) and (Hoff, 2014). Figure 20 lists typical milestones of linguistic competence development for human children during this period.[57]

---

[54] See (Spelke, 2022) for an overview of early infant classification of sounds and speech.

[55] Examples of motherese and the responses of babies are available on the web (e.g., UW I-Labs, 2020)

[56] Forms of motherese are also used by mothers in non-human species, such as bats (Bakker, 2022).

[57] Elizabeth Spelke gives a detailed summary of the perceptual and cognitive challenges of mastering word recognition and meanings (Spelke, 2022)

| Age | Linguistic Competence Developments |
|---|---|
| Prenatal | *Phonology*. Infants perceive some phonological contrasts, "bat" versus "pat"; or "bat" versus "bad". |
| 0-3 months | *Prosody and Intonation*. Infants become sensitive to language prosody and temporal variations in intonation that reveal boundaries of phrases and emotional emphasis. |
| 3-6 months | Infants attend to repetitions of their own name. |
| 6-9 months | *Words*. Infants recognize words learned in isolation. They begin to connect words to their referents. Infants learn the abstract ordering of function and content words. Infants learn that function words like "the" and "you" form heads of phrases. They learn that the complements of "the" are nouns and the complements of "you" are verbs. |
| 10-12 months | *Grammar and Referents*. Infants identify with confidence the sounds of their language, the. grammatical class of each word, the role each word plays in a sentence, and connections between words and their referents. Infants expect that speech will be efficient and relevant to the situation that the speaker and child are communicate about. |
| 18-24 months | *Telegraphic Speech*. Infants use 2-word sentences and drop function words (telegraphic speech). They likely have mapped familiar content words like "cup" to representations from the three core knowledge systems (Object, Form, and Agent). They understand speakers use such words in speaking about objects with particular forms and functions. |
| 24-30 months | **Longer sentences**. Infants use 3-word and longer sentences, with words in correct grammatical order. |

*Figure 20. Milestones of early linguistic competence developments in toddlers (Spelke, 2022) and (Hoff, 2014).*

Peter Eimas and colleagues showed that infants as young as one month are responsive to speech sounds and able to make fine distinctions and to categorize sounds (Eimas, et al., 1971). At around 18 months, children begin to combine words. As they interact and try to communicate with others and are schooled in grammar, their ability to speak correctly improves. At 18 to 24 months children use 2-word expressions such as "give milk" or "dirty shoe." This is called telegraphic speech. In telegraphic speech, the words of a sentence tend to be in the correct order, suggesting that children are beginning to recognize some elements of grammar. The sentences include main meaning words and leave out connecting words. Sometimes the toddlers group telegraphic sentences together as a short narrative.

Below are some of Kopp's examples of a toddler's sentences during an afternoon at a day-care center.

"One child's vocabulary … included sentences that reveal several kinds of knowledge:
- 'Finished juice already' (knowledge of the amount of time expected to complete a task)
- 'Oh, oh, dropped cracker' (past tense; knowledge that food is not to be dropped)
- 'I can't turn around' (knowledge of own limitations)
- 'Billy took play dough from Jill' (awareness of others' rights of possession).'"

(Kopp, 2013)

### 4.6.2.3 Human Acquisition of Reading and Writing

Beyond speech, modern human natural language communication includes reading and writing including iconic symbols. Today, the literacy of "reading and writing" extends to interacting with digital media.

Human competences for reading and writing develop later than those for understanding and generating speech. Children generally encounter reading when their caregivers read them stories. They point to favorite pictures. They learn about turning pages and that reading in English proceeds from left to right, and from top to bottom on a page. For digital media they learn about pointing, selecting, zooming, and scrolling.

The descriptions of reading milestones in Figure 21 were created for parents and teachers. Compared with milestones for learning to speak, the timing and depth for learning to read and write is more variable.

Children learn reading differently, but by age 4 most children have learned the alphabet and can sing the "ABC's." They learn the sounds associated with letters. They recognize a few sight words. They learn to sound out words, learning to recognize words from their favorite stories and common words in their environment such as words on containers. By age 5 they learn about plurals, some forms of verbs, and irregular words. They connect the spoken words that they learned earlier and their meanings to written words. By ages 6 and 7, they can read simple sentences on their own and re-read sentences that don't make sense to them. They can spell many words. By age 8 they can read some materials that they want to know more about. They begin to learn how to find information in digital media.

At about the same time as they learn to read, children begin learning to write. By age 4 they learn to draw lines and shapes and can manage a pencil and crayons. By age 5 many children have experience with keyboards and texting. They recognize letters on the keys and can type out words. By age 8 they write longer sentences and can correct their own writing.

| Age | Reading Milestones | Writing Milestones |
|---|---|---|
| 2 – 3 years | *Alphabet and phonics.* English-speaking parents read to toddlers. Kids begin to learn the alphabet and most of the sounds associated with individual letters. Kids can point to familiar objects in pictures. | |
| 4 years | *Letters*. Many children know the alphabet. They know that words are made of letters and that reading goes left to right. They recognize a few "sight words" such as their names or words on containers without sounding them out. | *PreWriting*. Children begin using pencils and crayons. They begin interacting with digital media. |
| 5 years | *Words*. Children identify the beginning, middle, and ending sounds of words. They say new words changing the first letter such as *cat* to *rat* or *hat*. They can answer and ask questions about stories like *who*, *what*, *where*, *when*, and *why*. They can match words that they hear to words on a page. They learn that some letters have different sounds, and some are silent. They learn regular and irregular spelling rules and how patterns of letters make more advanced sounds including long and short vowel sounds and multiple consonant sounds (e.g., "th" and "ph"). | **Scribbling and Early Typing**. Children can write some numbers and letters. They recognize many letters on keyboards and can type words. They correct their typing mistakes. |
| 6-7 years | *Spelling and Independent reading*. Children know many high frequency words like *a*, *and*, and *the*. They get familiar with punctuation and capitalization. They re-read sentences that don't make sense. They use context to sound out and understand unfamiliar words. They can spell many words. | *Early Writing*. Children write simple notes and messages. Their handwriting is readable by others. |
| 8 years | *Reading to learn*. Children read books and digital sources to learn about things that interest them. They become familiar with prefixes, suffixes, and root words – like those in helpful, helpless, and unhelpful. They summarize stories. They can contrast information from different texts. They understand similes and. metaphors. | *Writing to communicate*. Children correct their own writing. They learn about spelling irregularities in written language. |
| Middle & High School | *Different Genres*. Analyze themes, characters, and symbolism and how they develop in the text. Understand satire, sarcasm, irony, and understatement. They explore and understand different genres such as biographies, fiction, literature, news, digital media, and topical magazines. | *Different Genres*. Children explore writing in different genres such as short stories, reports, and reviews. |

*Figure 21. Milestones for reading and writing. This table is adopted from educational resources for parents and teachers including (Morin, 2023; Zettler-Greeley, 2022; WriteReader, 2019).*

By middle school and beyond, reading and writing enable children to get information and to pursue topics that interest them. As they experience different genres, such as fiction, humor, opinion, rhetoric, advertisements, and so on – they can compare what they read to what they have experienced. They build on their capacity for critical thinking, recognizing cases where what is written differs from common sense or their own experiences.

*4.6.3 AI Systems that Acquire Communication and Language Competences*

In the beginning of their book *The Scientist in the Crib*, Alison Gopnik and her colleagues call out three difficult and much-debated philosophical and scientific problems relating to knowledge, communication, and language:

> "… the problem of knowledge—is one of the oldest and most profound problems of philosophy. … Three versions of the problem are especially important and puzzling … We'll call them the *Other Minds problem*, the *External World problem*, and the *Language problem*." [emphasis added] (Gopnik et al., 1999)

These three problems are about how it is possible (1) to recognize and understand that other beings have minds and to communicate with them, (2) to create predictive models of the external world by perceiving it and interacting with it, and (3) to acquire and use language for communication.[58]

Mainstream AI has largely been blind to these scientific and philosophical problems.[59] Nonverbal communication and language acquisition are also not main topics in computational linguistics. In contrast, language acquisition has long been studied in psychology (e.g., Lindsay and Norman, 1977). Yonatan Bisk and his colleagues (among others) argue that language acquisition will be required for future AI systems:

> "Language understanding research [in AI] is held back by a failure to relate language to the physical world it describes and to the social interactions it facilitates. … successful linguistic *communication* relies on a shared experience of the world. It is this shared experience that makes utterances meaningful." (Bisk et al., 2020)

The following section reviews selected advances in the state of the art of AI and other fields in language use and language acquisition over the last few decades. The arc of research in AI projects shows an evolution in the field where the working definitions of language acquisition changed over time to become more comprehensive and realistic. Meanwhile, advances across the virtuous research cycle of Figure 1 have created an opportunity for fresh multidisciplinary approaches to the three problems and for bootstrapping developmental AIs in particular.

---

[58] Section 4.1 describes approaches for the *External World problem* with findings from the neurosciences and developmental psychology about multimodal perception, object recognition, and model building by infants together with AI models of the competences. Section 4.2 describes findings from the sciences bearing on the *Other Minds problem* together with AI models for goals, multi-step actions, and abstraction discovery. Sections 4.3 through 4.5 describe findings bearing on curiosity, imitation learning, imagination, and play that bear on human competences that are co-developed with language competences.

[59] For example, in their discussions of natural language processing, introductory AI textbooks describe the model of language as "a probability distribution describing the likelihood of any string." They do not cover how a computer could *acquire* a natural language or use language for communication. Following approaches from logic formalism, they relegate the "meanings" of words to "semantic rules," rather than exploring analyses such as those by Herbert Clark, Susan Brennan, and others about how people *create common ground about the shared meanings* of terms and their communications (Clark & Brennan, 1991). For a related perspective on AI's lack of attention to the three philosophical and scientific problems, coverage in mainstream AI textbooks about (say) models of computer vision is typically not connected to models of language. See the sidebar on common grounding and perceptual grounding of meaning.

### 4.6.3.1 SHRDLU (1971)

In 1971, Terry Winograd's doctoral dissertation was a landmark research project showing how a computer could understand natural language (Winograd, 1971).[60] Winograd's SHRDLU system took typed commands and answered questions. Among other things, it disambiguated words and sentences using context and semantic information. The core of Winograd's dissertation was about the use of procedural knowledge representations for natural language understanding. The project did not address language acquisition.

SHRDLU's world was contained completely in computer memory. SHRDLU represented a symbolic virtual Blocks World of scenes with 3D objects. A grammar linked to procedures defined how a subset of English was processed. Statements in a grammar described objects, relations, and actions in the world. A mapping process associated the grammatical expressions about its world to logical expressions evaluated on data in its database. The assumptions and design choices of SHRDLU shaped the approaches of much later research.[61]

Like other AI projects at this time, SHRDLU used symbolic processing to model cognition. In the following we describe SHRDLU's approach as a background for the language acquisition challenges tackled by later AI systems. Recalling the three hard problems for communication by intelligent agents, SHRDLU sidestepped all of them. Since Blocks World had no agents, SHRDLU did not address the *Other Minds problem* or create teleological models of others. Its world of objects and actions was defined and closed. Objects were 3D figures. They had properties and relations to each other. For example, a red block had different properties than a green one. SHRDLU did not learn about new kinds of objects. In Blocks world, no surprising wind-up toy car could appear or race around autonomously.

SHRDLU also sidestepped the perception challenge of the *External World problem* since it had no external world. It maintained only an "inner world" that was a memory model to represent the inventory, properties, and positions of all of the objects in a scene. To answer a question about an object or relation, SHRDLU consulted its memory. It's actions were MOVETO, GRASP, and UNGRASP.[62] It did not employ a realistic physics model for objects. An object was supported by one other object, or it was on the ground. Other than SHRDLU's virtual hand, there were no bodies in its world.

SHRDLU also sidestepped the *language acquisition* aspect of the language problem. Its language model was pre-programmed. SHRDLU did not learn syntax, new words or their meanings. It used a manually built language model with procedural semantics.

Many AI research projects over the next decades made similar simplifying assumptions. For example, many projects used an internal model of a Blocks World or had a small set of known kinds of objects. Many following AI systems had no perception system and did not need to recognize objects from perceptual signals. Creating AI systems that acquire language like children requires going beyond SHRDLU's simplifying assumptions.

---

[60] Winograd's dissertation was titled *Procedures as a Representation for Data in a Computer Program for Understanding Natural Language*. This dissertation influenced subsequent AI research in knowledge representation and many other areas and including framing tests relevant to tests for commonsense reasoning (Levesque et al., 2012). In later research, Winograd deepened his investigations into fundamental questions about cognition and communication.

[61] Winograd's dissertation influenced subsequent AI research in knowledge representation and many other areas. In later research not reviewed here, Winograd's research and writings returned to fundamental questions about cognition and communication and moved well beyond the assumptions of his dissertation research. His work inspired later tests for common sense reasoning in open worlds (Levesque et al., 2012).

[62] This description of SHRDLU actions leaves out some details. For example, if a user issued an impossible command, an underlying planner in SHRDLU would try to carry it out, fail in execution, and then undo all the attempted actions in the command.

#### 4.6.3.2 $L_0$ (1991-1996)

Two decades after SHRDLU, Jerome Feldman and his colleagues proposed a language acquisition challenge problem in their report, *Miniature Language Acquisition* (Feldman et al., 1990). The report was motivated by their concern that the field of cognitive science was fragmenting. The authors noted that although cognitive science included the topics of visual perception, language, inference, and learning, individual papers tended to be about one of the topics without reference to the others. The authors argued that this siloed practice impoverished the study of language acquisition as follows:

> "… language acquisition is defined in a very limited way. … 'language acquisition' usually means just syntax acquisition and research has not attempted to characterize how people learn to describe what they see." (Feldman et al., 1990)

Their report proposed a challenge problem for AIs to learn a subset of natural language from picture-sentence pairs in children's books. To demonstrate and test language understanding, the system would be given a new scene and asked to answer questions about it. Unlike SHRDLU, a solution to their challenge problem would employ perception by using a camera aimed at a page to see a drawing of the scene. In the challenge problem the computer would need to discover the grammar, the objects, and the relations.

The authors proposed simplified test conditions for the knowledge acquisition challenge including (1) a grammar for a simple language ($L_0$), (2) a few scenes in a 2-D variant of Blocks World, and (3) a sample set of propositional spatial relations for geometrical objects.

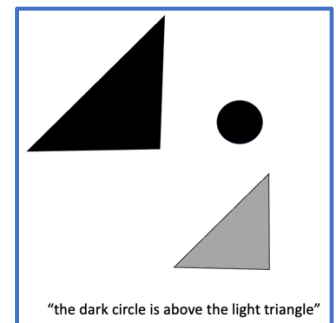

*Figure 22. Example pairing of a 2-D Blocks World picture with a sentence. Adapted from (Feldman et al., 1996).*

In 1996, Feldman and his colleagues reported on the results of five years of research on the $L_0$ project (Feldman et al., 1996). Recognizing the challenges of developing an integrated system for the challenge of answering questions based on perceived pictures from children's books, they later simplified this problem in several ways and organized their research presentation into multiple projects that addressed different subproblems as follows:

> "The system is given examples of pictures paired with true statements about those pictures in an arbitrary natural language. (See Figure 22.)
>
> The system is to learn the relevant portion of the language well enough so that given a novel sentence of that language, it can determine whether or not the sentence is true of the accompanying picture." [Figure numbering adjusted.] (Feldman et al., 1996)

The first $L_0$ project was a manually programmed system to demonstrate language acquisition in the $L_0$ framework. It relied on researchers programming knowledge about language from an informant. This project was discontinued after providing proof-of-concept demonstrations for multiple languages.

The next two $L_0$ projects involved language acquisition, based on Stephen Omohundro's incremental learning approach of *best-first model merging* (BFMM) (Omohundro, 1992). See the sidebar for a summary of BFMM.

The second $L_0$ project, the doctoral thesis of Andreas Stolcke, focused on *learning syntax* without representing the meanings of the words.

Using BFMM, it successively improved a stochastic context-free grammar model from a set of grammatical sample sentences (Stolcke et al., 1994). The sample sentences are assigned initial prior probabilities. The model-merging step generalizes grammar instances by rewriting them with non-terminal (abstract) terms. Goodness criteria assign a higher posterior probability to grammars having fewer rules and fewer abstract terms. The goodness criteria also prefer short productions and replace repeated sequences of terms by an abstract term (chunking). A working grammar is produced by repeatedly performing merging and chunking operations in a best-first manner until no more changes are required.

> **Best-First Model Merging (BFMM)**
>
> The design of incremental machine learning approaches for creating collaborative AIs and other open world applications is an open research problem. This sidebar describes the **Best-First Model Merging** technique (Omohundro, 1991) that was used in the $L_0$ project. **BFMM** predates mainstream deep learning approaches and the generative AI approaches of large multi-modal models. This technique does not necessarily require the vast computational resources typically needed to train large unified models. Its concepts are bio-inspired and include ideas that could be adapted for self-developed learning.
>
> BFMM dynamically and incrementally develops the structure of a neural architecture. Its learned models can be used both for recognition and for prediction. Early in its learning process, BFMM memorizes the details of its individual experiences. As it accumulates more examples, it gets enough data to generate and validate more abstract (and complex) component models. BFMM can generalize its piecewise model by replacing two of its component models with a single merged model. The merged model depends on the combined data from the two previous models and can come from a larger parameterized class. A merged model is never more complex than can be justified by the data. The technique merges component models to create more complex abstract models according to the following criteria:
> - *Continuity prior*: Prefer continuous models over discontinuous ones.
> - *Sparseness prior*: Prefer decompositions where the models directly affect each other in a sparse manner. (Sparsely-interacting components)
> - *Locality prior*: Prefer models in which the data decomposes into components which are directly affected by a small number of model components.
>
> In summary, BFMM generates a piecewise component model rather than a single parameterized model of a domain. Omohundro's paper illustrates the technique with three applications. He cites mathematical analyses of biological stimulus/response functions by Roger Shepard (Shepard, 1987) and statistical analyses that estimate dependencies from data by Vladimir Vapnik (Vapnik, 1982) as justifications for the technique.

In addition to having a model of syntax, this project needed a way to populate its limited world model. $L_0$'s world model was represented in a database. For example, processing the example sentence "The dark circle is above the light triangle" in Figure 22 would result in database entries for feature-based semantics similar to the following:[63]

- There exists an object Triangle #1.
- Triangle #1 is dark.
- There exists an object Triangle #2.
- Triangle #2 is light.
- There exists an object Circle #1.
- Circle #1 is light.
- There exists a relation Above #1.
- The TR (trajectory) of Above #1 is Triangle #1.
- The LM (landmark) of Above #1 is Circle #3.

The authors summarized the achievements of their system's syntax learning as follows:

> "We have no illusions that our method can, by itself, derive arbitrarily sophisticated grammars for the more complex kinds of linguistic phenomena. Such a task would require considerable work both on the representations involved and the necessary learning algorithms." (Feldman, et al., 1996).

The third $L_0$ project, the doctoral thesis of Terry Regier, focused on learning single word concepts for spatial events and spatial relations given a syntax for the language.[64] It demonstrated learning single word semantics based on visual grounding and used the Omohundro technique for creating generalizations. Details of Rieger's approach on the $L_0$ project for learning *perceptually-grounded semantics* are described in (Rieger, 1991). An ongoing constraint and theme in the project was learning in the absence of explicit negative evidence.

Figure 23 gives an example of a scene template to guide learning from scene instances objects in spatial relations. The template is intended to guide the system in recognizing novel scenes not previously encountered. The TR and LM symbols in the pattern are interpreted as referring to arbitrary shapes. The label LM (for landmark) in the template indicates the object that is the reference object. In a novel scene, the LM object may have a different size, shape, or position. The label TR (for trajector) indicates the object that is related to the reference object. The label "Above" is the name of the relation demonstrated in the scene. The template could be a static image or a movie that demonstrates an action. In this example, the template demonstrates spatial relation where the trajector TR is above the reference object LM.

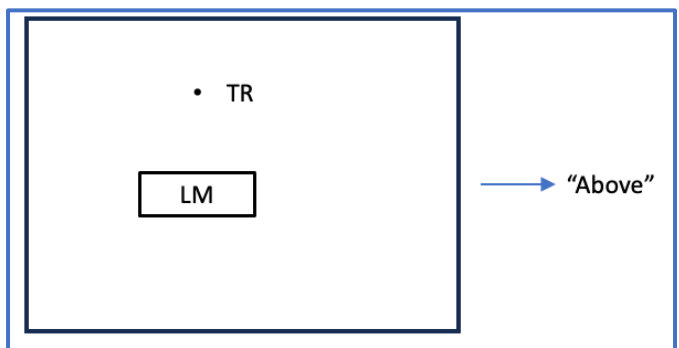

*Figure 23. Example input for learning the meaning of a term for a spatial relation. Adapted from (Rieger, 1991).*

---

[63] This project included additional built-in rules including "language independent constants" naming the relations (e.g., TOUCH and ABOVE), the relation argument names (e.g., trajectory, landmark, relation name), and the classes of objects (e.g., CIRCLE, TRIANGLE). The perceptual grounding of terms was not part of the first project.

[64] After the $L_0$ project, Rieger continued in related research and published extensively on learning and related linguistic and perceptual problems (e.g., (Rieger, 2003)).

Rieger called out issues with this characterization of the problem statement. The learning process needs to generalize appropriately to scenes that are not part of the training template. The template does not describe necessary constraints about generalizing. For example, an interpretation constraint is that the LM object should not be above the TR object or vertically aligned and horizontally off to a side of it. The template does not address what to do with edge cases of object shape such as a small TR object inside a hole in a large LM object. In hindsight, these representational inadequacies arose from complications from the design objective that learning should takes place without negative evidence. The reader is referred to (Rieger, 1991) for details of the approach. In broad strokes, the approach has assumptions about how parameterized spatial regions are assigned in terms of visual cells (or pixels) in bitmaps. The relations considered in the project include Above, Below, On, Off, Inside, Outside, Left of, and Right of.

$L_0$ advanced the art of language acquisition in AI. Like many exploratory projects, it made simplifying assumptions. It may be instructive to step back and consider some greater challenges that human babies face and master but that are not addressed by the research in $L_0$.

- *Perceiving objects and words.* The objects of $L_0$ have simple shapes and come in only a few types. Words (separated by white space) are trivial to locate in text. In contrast, babies need to separate and recognize the words and objects in messy perceptual signals. For example, objects in the real world can have variable 3D perspectives. They can move, be obscured behind other objects, and can have varying shapes and moving parts.
- *Unsupervised incremental learning.* Learning in $L_0$ depends on labeled examples of words, word categories, objects, etc. to enable supervised learning. In contrast, babies need to figure out parts of speech, word usage, and grammar in context from their sound and visual perceptions of the world, people, and social situations without having labels as guides.
- *Complex worlds and human interaction.*[65] Communicating and collaboration were not explored in the $L_0$ Project. The $L_0$ world is 2-D, has no agents, and has no invisible teleological state. In contrast, babies need models of the real world, models of other agents and models of the internal states of other agents. They learn to communicate and collaborate. They learn words for all of these.

Other $L_0$ projects were planned to investigate models for world worlds, agents, and cognition. At the time of their paper, these proposed projects were at an early conceptual stage and not yet developed.

In summary, the $L_0$ project recognized several scientific challenges to language acquisition and provided a concrete early example of an approach. It advanced the state-of-the art in *linguistic competence*. Although the project developed a limited competence for perceptual grounding, it relied on hand-coded biases and simplified data. The topics of agency, common grounding, and communication competence were deferred to later research. See the sidebar on Common Grounding and Perceptual Grounding.

### 4.6.3.3 CELL (2002)

A few years after $L_0$, Deb Roy and Alex Pentland reported on a computational model ("CELL") for *word acquisition* that learns from multimodal sensory input (Roy and Pentland, 2002). By demonstrating speech and camera-based visual grounding of word meanings, CELL advanced the state of the art of AI language acquisition.

---

[65] These rich competences and their acquisition are described in earlier sections. The findings of neuroscience and developmental psychology required for understanding were mostly out of scope for $L_0$.

The research goal for CELL was not to create an AI able with a rich linguistic understanding, parsing, or segmentation of speech. Rather, it was to create an AI able to *discover a few words* from unsegmented speech input and to learn their association with simultaneous images. They addressed three research questions.[66]

> "First, how do infants discover speech segments which correspond to the words of their language? Second, how do they learn perceptually grounded semantic categories? … [and third,] How do infants learn to associate linguistic units with appropriate semantic categories?" (Roy et al., 2002)

Briefly, the experimental set up included human mothers and their babies in a controlled setting. The mothers talked to their babies and placed different objects on a lighted table. Participants were given sets with seven objects. The parent's speech with the infant was recorded for later offline processing. The experimental situation was not set up as a "teaching" situation where caregivers would coach their children and speak in motherese. Instead, it involved mothers talking normally to kids about the objects that they picked up. The mothers were *not told to teach their infants words*. They were told to take one object at a time, play with it, and then return it to an out-box.

In parallel to the mothers and babies, a camera filmed the objects from deferent visual angles while the mother spoke to the baby. The mother's speech was recorded for later offline processing. The input for language acquisition combined the video and audio feeds.

The visual equipment for capturing images of objects used a color CCD camera on a mechanical arm that could move the camera through a range of vantage points, adjusting the elevation and angle of the camera. The target objects – toy trucks, toy cars, toy dogs, toy horses, keys, balls, and shoes – were placed one at a time on a turntable. A set of 209 images were captured for each of the 42 objects used in the study. A set of 15 images was randomly selected for each object ahead of the acquisition sessions. The data collection involved individual caregivers and their infants in six sessions over two days.

CELL processed each participant's data as a separate experiment. Details of the algorithms, representations of object shape and sound, experimental questions, and findings are reported in (Roy et al., 2002).

CELL's experimental situation was more challenging than the set up for $L_0$. The AI was not given a 2-D image, text input, and labeled training pairs. The mother's speech was fluid and was not scripted. The AI had to process images of real 3D objects and process fluid speech.

Nonetheless, this experimental setting was simpler than what babies experience when they acquire language in their natural home habitats. An infant's natural world does not involve highly controlled experimental situations. For example, in acquiring competences for object and speech perception in a non-experimental setting, infants deal with noisier input, multiple people speaking at once, and various interfering sounds. Although a toy car has complex shape, in the experimental setting it was placed by itself on a well-lit white table and recorded from several perspectives. In contrast, infants are presented with multiple objects at once in changing and complex contexts rather than ideal views of isolated objects.

CELL's language acquisition experiments did not involve understanding phrases or sentences, complex meanings, or relationships. Although the speech input for the experiment was fluent, the only words under study for language acquisition were the nouns that referred to the seven kinds of toy objects in the experiment. CELL had to recognize words in speech, that is, to identify words and word boundaries in the sound stream. Word pronunciations varied.

---

[66] Although these research questions are phrased in terms of infant-driven research and involved mothers and their children, the experiments were about an AI acquiring words and their meanings.

CELL's model computational model was built on existing technology. This included shape models, phoneme collections, and learning algorithms that had been developed earlier and separately for computer vision, machine learning, and speech recognition.

In contrast to CELL, infants learning language follow their own interests as they see and manipulate objects. They develop language competences while they develop competences for object perception, goals, and abstractions. They spend many months on their "full time jobs" developing models about themselves and their world. In contrast, CELL did not have any goals or take any actions in the world. It did learn verbs or learn syntax. It did not create models of people or their goals. CELL did not learn language by interacting with people. It did not speak to people in order to influence them.

In summary, CELL's research went beyond the experiments of $S_0$ in several ways. It went beyond the three generations of NLP technology in its demonstration of acquiring words and their meanings in an actual (albeit simplified) setting. Although the language competences that CELL developed were much less than those acquired by toddlers, the project explored new technical approaches and research questions. It convincingly learned the meanings of nouns for a few toys.

#### 4.6.3.4 Connecting Language and Perception (2008)

In his 2008 paper *Learning to Connect Language and Perception,* Raymond Mooney identified research challenges for language acquisition (Mooney, 2008). He reviewed the language acquisition approaches of several AI systems including ones that he contributed to (e.g., Kate & Mooney, 2007) and systems created by other groups (e.g., Gorniak & Roy, 2005). To handle the complexity of the real world, Mooney argued that NLP should automate the learning process and use machine learning technology to acquire natural language including its semantics.

At that time, other state-of-the-art supervised machine learning approaches labeled the training data by pairing natural language sentences with *formal representations of meaning* (MRs). To reduce ambiguity and the amount of labeling, some systems incorporated parallel visual data. In these approaches, the subsystems needed to segment images, recognize objects, detect events, and provide probabilistic information. To advance the state of the art, Mooney advocated developing an embodied system approach for language acquisition that would learn language like a child and use perceptual signals and their abstractions to ground language.

> "To truly understand language, an intelligent system must be able to connect words, phrases, and sentences to its perception of objects and events in the world. Current natural language processing and computer vision systems make extensive use of machine learning to acquire the probabilistic knowledge needed to comprehend linguistic and visual input. However, to date, there has been relatively little work on learning the relationships between the two modalities. … Ideally, an AI system would be able to learn language like a human child, by being exposed to utterances in a rich perceptual environment. The perceptual context would … ground the system's semantic representations in its perception of the world." (Mooney, 2008)

Mooney did not sidestep practical utility by simplifying the definition of language acquisition. His paper was a call-to-arms for biologically-inspired experiments. He encouraged other researchers to join with him in addressing the hard research challenges.

### 4.6.3.5 ITALK (2014)

In 2014, Frank Broz and his colleagues published a progress report about learning and language acquisition in the multidisciplinary ITALK[67] project (Broz et al., 2014). As they described it,

> "Within the framework of human linguistic and cognitive development, we focus on how three central types of learning interact and co-develop: individual learning about one's own embodiment and the environment, social learning (learning from others), and learning of linguistic capability." (Broz et al., 2014).

The iCub humanoid robot[68] as shown in Figure 24 was used across several related research groups as a vehicle for experiments in AI language acquisition. This ambitious research program involved researchers in the neurosciences, cognitive and developmental psychology, and AI.

The ITALK project was organized around multiple projects using the iCub platform. This led to a demonstration that integrated several separately developed systems. Broz and his colleagues summarized the resulting demonstration as follows:

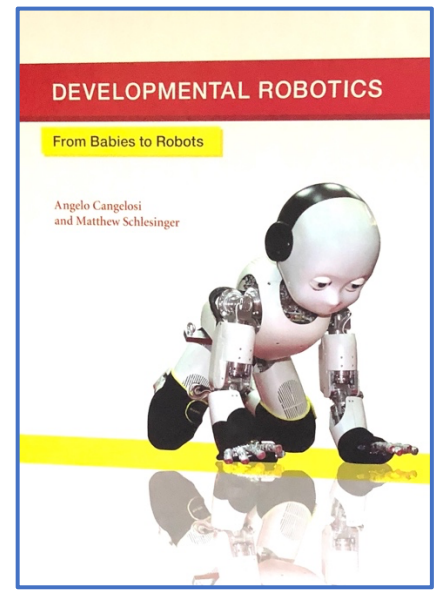

*Figure 24. Photo of the cover of Angelo Cangelosi and Matthew Schlesinger's book "Developmental Robotics" (Cangelosi et al., 2015), showing an image of the iCub humanoid robot.*

> "The ITALK project addresses development up to roughly 2 years in age, when the ability to communicate in short sentences emerges. … At the first stage, the robot performs simple actions on objects and learns word object associations from the human. During the second stage, the robot learns word-action and word-attribute associations from the tutor's verbal descriptions of actions it performs on objects. And during the third stage, the robot demonstrates simple grammar learning based on the speech it has been exposed to and learns to compose complex actions from simple ones. (Broz et al., 2014)

In summary, developmental robotics research on language acquisition stopped at the "Talking Gap."

### 4.6.3.6 Developmental Robotics Projects on Communication (2015-2023)

In their comprehensive book *Developmental Robotics*, Angelo Cangelosi and Matthew Schlesinger reviewed the advances, state of the art, and ongoing challenges for developmental robotics.[69] The authors reflected on the developmental robotics community and the experiments and demonstrations that it had carried out.

---

[67] ITALK is an acronym for Integration and Transfer of Action and Language Knowledge in Robots.

[68] A paper by Nikos Tsagarakis and his colleagues describes the technical specification, purpose, and rationale for the iCub humanoid robot. It was designed as an embodied robotic child with the physical height, mass, and cognitive capabilities of a 2.5 year old human child (Tsagarakis, 2007). Many other robotic children were designed for experiments in developmental robotics at about the same time, including the CB² (Minato, 2008). Cangelosi et al. review human baby and adult-sized robots in (Cangelosi et al., 2015).

[69] *Developmental robotics* is also called *cognitive robotics* and *epigenetic robotics*. In contrast, the similar sounding term "collaborative robotics" is typically used more narrowly to describe industrial robots (also called *cobots*) that are designed for workplace safety more than for research on robot cognition. A focus of collaborative robotics is on force limiting robot arms and other devices that stop moving if they bump into people. Custom commercial cobots and applications for warehouses, construction, manufacturing and other applications are available through several companies (e.g., Brooks, Collaborative Robotics, Epson, Fanuc, HP, Mitsubishi, and Universal Robots). Although collaborative robotics

> "The tendency [in the developmental robotics community] to produce models that focus on isolated language development phenomena is in part due to the early maturity stages of the developmental robotics discipline, and in part due to the inevitable complexity of an embodied, constructivist approach to language and the related technological implications. Given the early stages of developmental robotics research, the studies … mostly specialize on one cognitive faculty. … in most studies, researchers tend to start with the assumption of preacquired capabilities and investigate only one developmental mechanism." (Cangelosi & Schlesinger, 2015 – page 272)

Several papers in this period reviewed technical approaches for discourse and natural language and recommended directions for further research (e.g., Ahn et al., 2022; Dubova, 2022), Cangelosi & Asada, 2022; Marge, et al., 2022; Tamari et al., 2020; Tellex et al., 2020; McClelland et al., 2019).

Marina Dubova reviewed the technical approaches for NLP and their shortcomings. She identified variations on *grounding* as a recurring theme (e.g., Dubova, 2022). See

> **Grounding of Meaning and Cognition**
>
> When they are learning a language, babies see the external world and hear others communicating. They create cognitive models of the world, of other people, and of communication. They construct meanings for words and expressions by their association with objects and actions in the world. They learn language and meanings from the demonstrations and teachings of other people.
>
> Spoken words and sentences are patterns of sound; written words are patterns of markings. Their meanings can be about directly observable things in the world, abstractions, and things that can be inferred but not directly observed. For example, the world "dog" can refer a particular nearby pooch, or it can refer to a kind of animal – generalizing to create a class based on multiple observed individual dogs. Abstractions can refer to cognitive concepts such as goals and dreams. Mapping patterns of expressions to perceptions of the world is called **perceptually grounding**.
>
> Establishing common ground between two parties requires shared perception of a shared world, discussion, and negotiation. When either party does not understand the meaning of a communication, they may point to objects or drawings in the setting, give examples of objects, actions, and events, and discuss the salient features to establish **common ground** (Clark & Brennan, 1991). They notice breakdowns in communication and repair them to maintain their shared understanding (Klein et al., 2004; Stefik, 2023). Crucially, common grounding is not just a mental activity of an individual. It requires social processes to create, validate, and maintain common understanding.
>
> Allen Newell defined a **symbol** as a representation that gives **distal access** to information (Newell, 1990). Newell was referring to **internal symbols** in cognitive processes. **External symbols** – such as those in spoken and written languages – also provide distal access to information via the perceptual associations that agents learn (e.g., Stefik, 1995). See Lawrence Barsalou's review *Grounded Cognition* for a comprehensive discussion extending the concept of grounding of symbols to socially grounded of cognition more generally (Barsalou, 2008).
>
> The three generations of NLP systems described earlier do not use perceptual grounding or common grounding to learn and establish meanings. Current AIs typically lack perception and the foundational competences that humans use to model the world. They lack competences to model each other's mental state, and to communicate and collaborate effectively including establishing, maintaining, and repairing common ground.
>
> Dictionaries *describe* the meanings of words. They depend on their readers already knowing the meanings of the words used in the definitions. To illustrate, suppose that an agent comes upon a dictionary that defines an "alpine pear" as "the sweet fruit of copper clad trees that grow in the clouds above the Swiss Alps." An informed agent with adequate world models would realize that this proposed definition is inconsistent with commonsense models of the real world.

the sidebar on grounding of meaning and cognition. One important kind of grounding for language is *common grounding* or shared understanding. Common grounding is about the condition where two people in conversation talk about something and develop a workable understanding of when they are talking about the same thing and when there is an ambiguity or confusion.

Natural language can refer to anything at all. Here are some examples that illustrate challenges in grounding:

- Language can be about a specific *context*.
- It can be about objects that exist in that context or objects more generally.

---

does not focus as much on the cognitive requirements for collaboration, the engineering developments in this field are preparing the way for human-safe AI collaborators. Collaborative robotic systems are typically programmed for lifting, placement, and tool application in repetitive and narrow applications that do not require general competences or rich world models for task work or teamwork.

- It can be about actions or activities taking place in a context.
- It can be about proposed actions that might take place in a future context.
- It can be about *perceptions* of objects.
- It can be about what someone is thinking.
- It can be about a person's goals and beliefs.
- It can be about characters in fictional stories.
- It can be about what one agent thinks that some other agent is thinking.
- It can be about a particular thing at a particular place at a particular time -- such as the Stefik family dog whose name starts with "S," who barked out a stairway window at our home in California at 2:00PM Pacific Time on the afternoon of April 10, 1998. (Although the preceding sentence is fairly precise, it does not convey the visual appearance of that particular dog or what the dog was barking at.)
- It can be about things imagined that never were and never will be.

Angelo Cangelosi and Minoru Asada wrote about the absence of social grounding processes in typical systems build with using NLP models. As they summarize in *Cognitive Robotics* (Cangelosi & Asada, 2022):

> "NLP-based models, on the other hand, have been widely used to handle dialogue with conversational agents and complex lexicons. However, in these models the robot is not able to autonomously ground the words it uses for sensorimotor knowledge, and it must rely on the hand coding of the word-meaning mappings defined by the system designer." (Cangelosi & Asada, 2022, pp. 407)

More recently, generative approaches based on large language models have become widely available. Although they may appear to use language fluently, they have provoked concerns and debates about whether they know enough (or can know enough) about what they are talking about. Developers outside of the research community also have also avoided the natural language acquisition challenge. They created systems that carry out tasks using off-the-shelf NLP technology. Relying on the first and second generations of NLP technology, these AIs cannot converse broadly or deeply in natural language.

In summary, although the three earlier generations of NLP technology have enabled the creation of useful interactive AI systems, their shortcomings have limited their utility in creating systems that learn what they need to know. These approaches are based on assumptions about AI systems that differ radically from how communicating species operate. The language systems are not integrated with perception. The meaning of their language is not grounded perceptually or cognitively. They do not test or develop common ground with their users. They do not develop models of the world or of people and other agents.

In contrast, language in communicating species is part of a complex of competences for perception, for modeling the world, for modeling other agents, for learning, for coordination, and for communicating with purpose.

- *Learning for Language and Communication hypotheses.* Competences for communication require the same kinds of effective world modeling, teleological modeling, and social modelling as imitation learning, coordination, collaboration, and play. The hypothesis is that the acquisition of communication and language competences by AIs requires the same core competences that are acquired by people. AIs require experiences in a similarly rich world and social experiences to develop them.

Strong capabilities in communicating with language are crucial for the success of future AI systems. In alignment with these requirements and observations, the next section explores how the trajectory of competence acquisition works holistically for people and other communicating species and how it could work for bootstrapping AIs.

# 5.0 What's Next for Developmental AI?

This section proposes a research roadmap that addresses gaps in the current state of the art of developmental AI.

We begin by considering an example of the technology readiness of some current AIs that were created differently and their persistent limitations.

## *5.1 AI Technology Readiness – The Case of Self-Driving Cars*

There is much to applaud about the advances in autonomous vehicles (AVs) since the first DARPA Grand Challenge in 2004. In that competition, none of the prototype vehicles finished the test course (DARPA, 2004). Today, many automobile companies offer vehicles with advanced capabilities for autonomous driving. However, current AVs still make strange, concerning, and sometimes life-threatening mistakes in deciding what to do.

Based on her experience with the National Highway Safety Administration, Mary Cummings reviewed cases where self-driving cars crashed, contributed to other vehicles crashing, blocked city streets, and cases where the AIs failed to deal with uncertainties and unusual conditions in driving situations (Cummings, 2023). For example, in San Francisco, police and fire fighters waved and shouted without effect at a misbehaving AV (Eskenazi, 2023). The emergency workers had to disable it. In another news story, Cruise's driverless taxi service was suspended in October 2023 when a pedestrian was trapped under a driverless taxi and dragged for 20 feet (Lu & Metz, 2023). To continue testing the AVs, the General Motors subsidiary was required to have human safety drivers in the taxis ready to take over in an emergency.

**Sample Expectations for Future AI Applications**

**Self-Driving Cars**
- Recognizes situations where children could suddenly and dangerously chase a ball into the street or lose control of a bicycle.
- Recognizes that a person in a parked of car is distracted and may carelessly open a car door and step into traffic.

**Sous-Chef Robot**
- Helps the chef while she focuses on the preparation of a favorite dish including anticipating the need to slice vegetables, wash dishes, and turn down the burner for pots that are about to boil over.
- Teaches a new robot team member about the chef's preferences.
- Develops a delicious new recipe variation when some ingredients are not available.

**Nanny Robot**
- Tends to the care of young ones (including robot children), guides their development, and keeps them safe.
- Consults with parents and others on how the children are doing.
- Talks to them (sometimes in motherese); sings to them; arranges an appropriate learning environment.

**Research Assistant Robot**
- Prepares for a team discussion by gathering recent reports from others in the field.
- Is skeptical of reported findings from a research team that are inconsistent with a model's predictions.
- Notices that some procedures designed by a researcher will not work for an experimental situation.

**Engineer/Designer Robot**
- Recommends design constraints that improve sustainability and repair.
- Proactively anticipates and suggests work arounds when materials or parts are not available.

**Hardware Store Assistant Robot**
- Advises customers about available replacement parts for repairing equipment they have.
- Carries goods to a customer's vehicle or arranging pickup and delivery services.

These failures reveal shortcomings in two abilities: (1) making appropriate decisions in uncommon situations and (2) understanding people and their goals.

The sidebar gives examples of future AI applications and social expectations for them. Similar to the case of current AVs, future AIs for these applications would need to act appropriately in uncommon situations and to communicate and coordinate with people.

Erik Vinkhuyzen and Melissa Cefkin bring a social science perspective to the study of self-driving cars. They investigated how AVs interact with pedestrians, bicyclists, and other cars. Reporting on the driving decisions of commercial AVs, they said:

> "… driving is not only a technical but also a social skill … [This] is obvious from the moment one progresses from driving in a parking lot to driving on the public roads. Indeed, for anyone to manage their movement through time and space, whether in a car, on a bike, or as a pedestrian, is a social act that involves the interpretation of cultural signs and signals ..." (Vinkhuyzen et al., 2016)

Self-driving cars have rules for common cases, such as for staying in their lanes on roads and highways. AVs need to exercise judgement about when they could temporarily and carefully violate such rules. For example, consider an AV taking action to avoid stopped traffic. The AV could swerve slightly out of its lane when another car blocks the lane to wait for a pedestrian. When people do something unusual like this, they typically may make eye contact with other people and signal their intentions.

In summary, current AVs lack robust abilities to understand people, to coordinate with them, and to communicate and learn from them.

## *5.2 Competence Acquisition Gaps in Developmental AI*

When we compare the abilities of current developmental AI research systems with the abilities of people, three competence acquisition gaps stand out. These are the Nonverbal Communication Gap, the Talking Gap, and the Reading Gap. Most people bridge these gaps while they are children. The competences in the gaps support communicating with others, coordinating activities, and learning from other people.

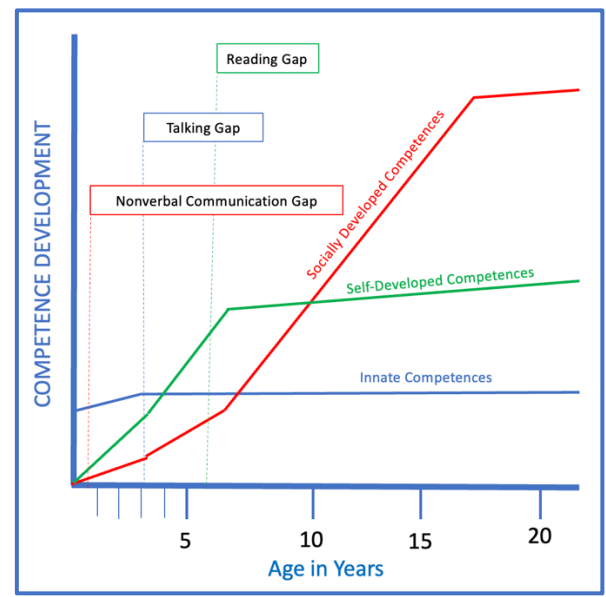

*Figure 25. Notional levels of innate, self-developed, and socially developed competences for people.*

To emphasize the importance of bridging these gaps, **Error! Reference source not found.** depicts notional levels of competences acquired over time for people.

- The blue curve sketches the level acquired via *innate* competences.
- The green curve sketches the level acquired via *self-developed* competences.
- The red curve sketches the level acquired via media.

In short, people learn to do many things during their lives. They are born into the world with innate competences (blue curve). If that was the sum of their learning, humans would be among the lesser animals. As children mature and experience the world, they learn how to do many things that no other species can do (green line). If human learning stopped there, every person would need to figure everything out for themselves. Human life would be much as it was when the human species first appeared. In broad strokes, our civilized lives today are a consequence of our ability to work together and to learn from each other (red line).

The field of developmental AI has not deeply investigated the Nonverbal Communication Gap, where children learn to communicate with pointing, gaze, and other gestures. Children bridge the Talking Gap at about three years of age when the first stages of purposeful speech typically appear.

When people do not learn to read, they are blocked from full participation in many activities in life. Developmental AIs will need to cross this gap in order to expertly access and fuse social developed information from media.

Table 3 lists competence areas on the development trajectory for children. The competences develop in parallel and in small steps.

| Competence Area on Bootstrapping Trajectory | Description | Impact of Competence |
|---|---|---|
| *Section 4.1* Perceiving, Understanding, and Manipulating Objects | Children build mental models of the world and of their bodies. | Perceptual, motor, and cognitive foundations for later competences. |
| *Section 4.2* Discovering Goals, Multi-Step Actions, and Abstractions | Children employ compound and abstract actions and goals and build models of them. | These cognitive actions instantiate and generalize models of the world and support efficient planning about possible actions. |
| *Section 4.3* Curiosity and Intrinsic Motivation | Children use models to make predictions about changes depending on context and their actions. They notice prediction failures and anomalies, update their models, and manage their attention. | Expanding to open-ended activities and discovering new kinds of goals. These are cognitive foundations for exploration, discovery, creativity, and invention. |
| *Section 4.4* Imitation Learning | Children notice that others do interesting things. They create teleological models of the goals and actions of others. They translate the goals and actions of others to their own goals and actions. | Expanding capabilities by learning how others do things, rather than learning everything on their own. |
| *Section 4.5* Imagination, Coordination, and Play | Children notice that others do things together. They learn in play that people can have different roles and goals. People can collaborate or compete. In combination with communication skills, children earn to be skeptical and interpret what others say. | Expanding capabilities and doing things with others. These are cognitive foundations for communication and collaboration They learn that others may not do what they want. |
| *Section 4.6* Language and Communication | Children notice that others point and make sounds to influence others. They discover that they can influence others with nonverbal and later spoken communication. Later they learn to talk and to read and write. | Communication competences enable the acquisition of knowledge that was developed by others. |

*Table 3. Broad competence areas on the trajectory of human competence development.*

*5.2.1 The Nonverbal Communication Gap*

Parents typically pay much attention to their child's speech development. In contrast, they pay less attention to their earlier nonverbal development. Linguistics and AI also pay little attention to nonverbal communication. In contrast, pet owners often judge the intelligence of their animals from their non-verbal communications.

Nonverbal communication precedes verbal communication. Nonverbal competences are integral to three remarkable "discoveries" by children: (1) the discovery of self, (2) the discovery of others, and (3) the discovery of communication.[70]

The *discovery of self* includes sensorimotor competences – perceiving, understanding, and manipulating objects. These competences could be described as a child learning about its body and its place in the world. Self-discovery further includes a child's learning about its mind – learning to have goals, take multi-step actions, and form abstractions. Following our trajectory competence groupings, self-discovery also includes curiosity and intrinsic motivation.

The *discovery of others* involves the sub-competences of imitation learning. These start early when a child observes others doing activities – well before it understands much about the world or has substantial motor control.

Most children cry at birth. Later, they may look at their mothers and smile when they are soothed. Consider a child months later pointing at a banana from a highchair at lunch time. The child then watches to see if their parent notices the pointing. The communication actions continue when a parent observes the direction of the child's pointing, looks at the banana, fetches it, and gives it to the child. This closed loop of two-way communication continues when the child and parent then smile at each other. In this way, learning the idea of communication precedes the learning of language.

*5.2.2 The Talking Gap*

Research on the ITALK project stopped at a development stage analogous to human children at about two years of age (Broz et al., 2014). In comparison to studies in psychology and education research, experiments of robot play only studied robots playing *alone*. It did not create situations where robots could use non-verbal communication such as pointing to direct the attention of other robots or where multiple robots could coordinate their activities with gestures or language.

Experiences and interactions of during the stages of play appear to provide children with powerful training experiences for learning teleological reasoning, communication and coordination. Informally, parents and pet owners refer to such play periods as having the purpose of "socializing" their charges.

Table 4 overlays the progression of communication and language competences with Parten's six stages of children's play. Parten's first stage – unoccupied play – corresponds to when human infants begin to learn how to perceive and manipulate objects. They fuse perception signals from vision, hearing, proprioception, and touch. They build mental models of the objects in the 3D world around them including their bodies. They build naïve physics models and predict what will happen next in the world. They cannot routinely perceive and move objects and are not yet ready for play.

By Parten's second stage – solitary play – children have developed the perceptual skills to recognize objects and 3D motion. They have also mastered early competencies for moving their bodies and manipulating toy objects. They are learning about multi-step actions. As they engage in solitary play, they discover abstractions and refine their mental models of objects and actions.

---

[70] The "discovery" of self could alternatively be characterized as the "development" or "invention" of self. These three "discoveries" are similar to the three difficult philosophical and scientific problems quoted from Alison Gopnik in Section 4.6.4: the *external world problem*, the *other minds* problem, and the *language* problem.

The Talking Gap occurs between Parten's third stage – onlooker play – and the fourth stage – parallel play. At this stage, children develop teleological models for their own goals and develop teleological models for the goals and actions of other people. They notice that people make noises (speak) and influence the behavior of each other. They explore communication with other children.

| Period | Parten's Stage of Play | Corresponding Competence Development |
|---|---|---|
| Birth to 2 years | 1. **Unoccupied Play.** Children are relatively still, observing materials around them without much organization. | **Perceiving, understanding, and manipulating objects.** Building mental models of objects in the world and their bodies. (Section 4.1) |
| 2 years | 2. **Solitary Play.** Children busy themselves manipulating materials, observing how the world works, and mastering control of their movements. They do not seem to notice or acknowledge other children. | **Goals, multi-step actions, abstraction discovery.** (Section 4.2) |
| 2 ½ to 3 ½ years | 3. **Onlooker Play.** Children watch other children play but do not join them. They may engage in some communication with other children.<br>– Talking Gap begins about here. – | **Curiosity** and **intrinsic motivation.** (Section 4.3)<br>Co-development with communication and language development (Section 4.6) |
| 2+ years | 4. **Parallel Play.** Children play separately from others. They are close to them and begin to mimic their actions. | **Imitation learning** (Section 4.4)<br>Co-development with communication and language development. (Section 4.6) |
| 3-4 years | 5. **Associative Play.** They practice what they are seeing and may coordinate their activities. | **Imagination, coordination, and play.** (Section 4.5)<br>Co-development with communication and language development. (Section 4.6) |
| 4+ years | 6. **Cooperative Play.** Children coordinate their activities and take on roles in joint tasks with shared goals. | **Communication and language.** (Section 4.6)<br>Further co-development with earlier competences. |

*Table 4. Relating Parten's Stages of Play to Competence Development*

Below are the time periods during which childhood communication and language developments typically occur:

- 12-18 months – mainly 1 word speech acts. Approximately a 50 word vocabulary.
- 24 months – two or three-word sentences.
- 36 months – 1000+ word vocabulary. More adult-like grammar. Children take turns speaking.
- 48 months – 10,000+ word vocabulary.

Parten's stages involve increasing levels of communication and language competence. Play is a major locus of practice first in nonverbal communication and then in talking.[71]

---

[71] A discussion of research on play deprivation and its effects on early child development is beyond the scope of this paper. Erika Hoff's book *Language Development* provides an introduction to the relevant research (Hoff, 2014). In the context of the COVID pandemic, there have been reports that play deprivation can damage early child development (e.g., Brown, 2018; Sparks, 2022).

In Parten's fourth stage – parallel play – children learn by watching the activities of other children. They mimic them and learn by imitation.

In Parten's fifth stage – associative play – children watch activities where other children do things together. They develop models for coordinating their activities with other children.

By Parten's sixth stage – cooperative play – children's language development is more advanced. They increase the range of possible activities with pretend play and take on more roles. Their expand their teleological models to joint tasks and shared goals.

Communications during play is not just about describing or acting on objects in the world – involving sentences like "This is a red block" or "Put the blue block on the green block" as in AI Blocks World scenarios. As play experiences develop, children's communication supports negotiation about taking turns, the goals and rules of games, and showing someone how to do something. In short, coordinating and collaborating provide motivation for learning to communicate.

Part (a) of Figure 26 shows an experiential station at the Tacoma Children's Museum in Tacoma, Washington. The museum provides a series of connected water-filled trays at waist height for children. Water circulates from the top tray to lower trays with waterfalls, turbulence, eddies, colliding toys, and so on. The objects in the tanks include water toys – little boats, ducks, and other objects that children may have encountered at bath time or in swim classes. This is an active learning environment.

In this museum, kids learn about manipulating objects in water, the physics of objects in water, and also skills for collaboration.

Children at the museum often play in pairs and small groups where they learn from each other. Occasionally, an older child or adult shows a younger child how to do something involving multiple boats and other water toys. The children watch and learn from others in this environment. They observe other people moving their arms and hands as they carry out sequences of actions. Through imitation learning, they pick up new understanding. They teach each other and tease each other. They pick up social skills during their play.

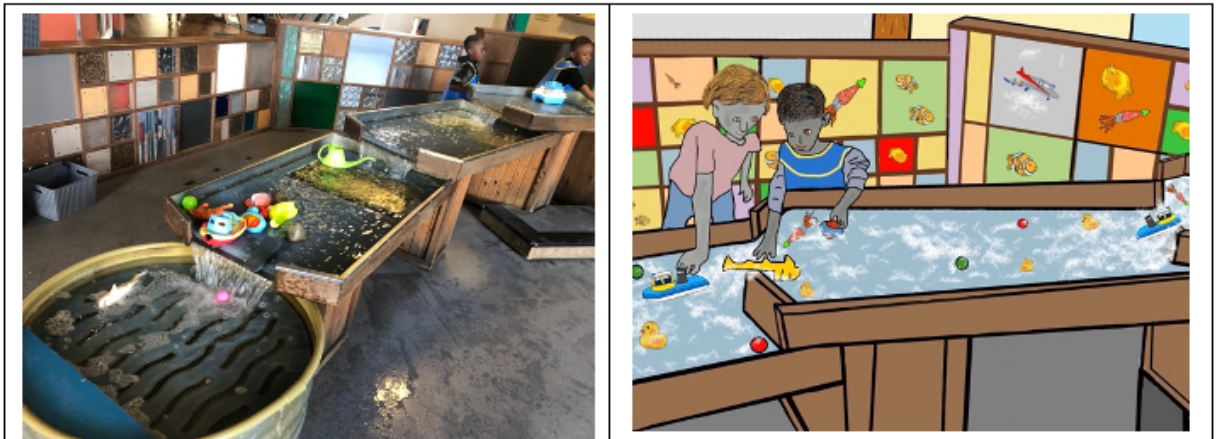

*Figure 26. (a) Water tank room at the Children's Museum in Tacoma, Washington. (b) Drawing of two fictional android "children" playing and manipulating objects in flowing water from (Stefik, 2022b).*

This museum setting is similar in design and intent to many preschool and kindergarten settings. The encouragement of exploration in the Tacoma museum is similar to educational practices for children in Montessori schools and in other experiential education settings.

Part (b) of Figure 26 is a drawing of imagined "robot children" in a similar setting where they would play and learn with human adults, other robot children, and perhaps human children.[72] Such a setting could provide an experimental environment for training "child robots" to engage in free play, to learn by imitation, and to be taught socially developed competences. Lacking such settings, developmental AI has not created an opportunity to study the role of play in creating much value for communication and coordination. Similarly, although large language models and multimodal models are richer settings for learning than vision alone, they miss an opportunity to bootstrap robots in learning to communicate.

### 5.2.3 The Reading and Writing Gaps

Children's first experiences with books and reading are usually when parents and other caregivers read to them. Their first books are mostly pictures. Physical books are 3D objects with pages. As in Figure 27, reading can be a relaxing and comfortable experience. Favorite books are read over and over again. As language skills develop, children become engaged in simple stories. The stories are typically about simple experiences and social situations. They can include characters in fantasy worlds with animals or vehicles that talk.

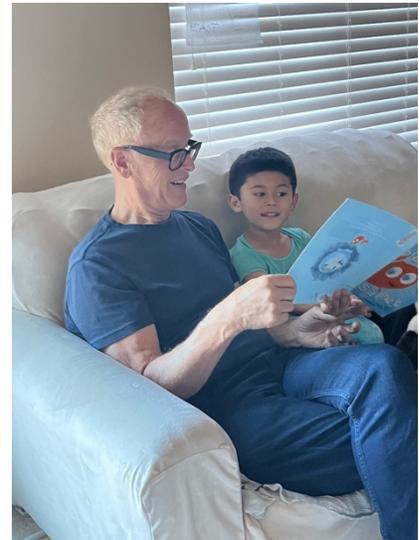

Children advance in the variety of books that they read or have read to them. They become experienced with different genres of books and stories. What happens in fantasy books is different from what happens in real life. Children recognize that animals and vehicles do not talk in real life. For example, real bears don't have cottages, talk, or wear clothing. The little engine does not really talk about trying hard ("I think I can!"). Children develop skills in knowing that stories and their authors say things for their own purposes. They need to determine how to understand the information that that they read.

*Figure 27. Macer and his grandfather reading a book.*

When reading to children, parents may move a finger from word to word as they read. Such pointing draws attention to printed words and structure that a child may not notice when it first sees a page of a book. Readers repeat the same words aloud every time a book is read. Children learn that the visual printed words on the pages correspond to the words that they hear and have the same meanings. In languages like English, the words are arranged in lines and are read left-to-right on a page with lines ordered from top-to-bottom. Children learn favorite words on a page and recognize them when they appear in other books and places.

By about 4 years of age children typically recognize letters and know the alphabet. By 5 they know beginning and ending sounds of words and can sound words out phonetically from their spellings. By age 6 or so, they begin reading independently.

As children read and discuss what they have read, they learn to be selective about how they fuse information from different sources. Whether they read for fun or play video games and watch movies, children continue to take in information from many different sources. As they advance in school, they may discuss the news of the day from different parts of the world. With these different experiences at home, with friends, and at school they learn to pay attention to the sources of information.

---

[72] The drawing is from (Stefik, 2022b).

Their early interaction with others on the playground and elsewhere provides children with experiences that serve them later for assessing the suitability and trustworthiness of socially developed information. In a school context, young students write essays on different topics and rely on different reference sources. They receive feedback about the information sources that they use and may get lessons about critical thinking skills.

Figure 28 adds education research to the virtuous research cycle of Figure 1. Prototype projects taking a developmental robotics approach have generally not continued to a point where the AIs have learning experiences corresponding to those discussed here.[73] In order to address later periods of learning, education research is an important addition to the virtuous research cycle for developmental AI.

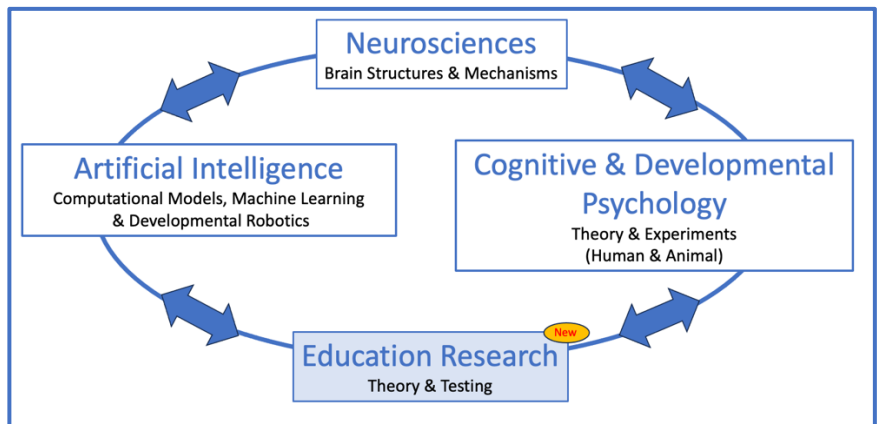

*Figure 28. Education research can bring knowledge and testing experience for competences on the trajectory of school that are acquired by age children.*

### 5.3 Towards a Research Roadmap

We recommend three interleaved research threads as organizational elements in a research roadmap:

- *Incremental and continuous machine learning for embodied developmental AIs.*
    - The training challenges for developmental AIs differ from those of deep learning and internet AIs. As with deep learning, substantial data is needed for learning. However, developmental AIs are created by embodied systems that learn by experience – interacting with and experimenting with the world. This includes interacting with people. They would develop competences in parallel in small steps along a competence trajectory. Embodied AIs would perceive and process *streams* of multi-modal information. They would build models of the world around them. They would learn continuously and incrementally.

        Although neuroscience and AI researchers have theories and concepts about the nature and operation of human learning "machinery" (e.g., (Zador et al., 2020); (Hawkins, 2021); (Hinton et al., 2021; 2018)), there are many open questions. Developing general learning systems for embodied systems is a substantial challenge for progress in developmental AI.

- *Multi-disciplinary testing and evaluation to assess what a developmental AI has learned.*
    - How can we determine what an AI has learned when it interacts with the world (or with a person)? What does it get right or wrong? What has it learned or mislearned from its interactions with the world or with people?

---

[73] Pushing the leading edges of research on developmental robotics and education, Pierre-Yves Oudeyer and his research collaborators have explored the use of AI models to inform the design of education technologies (Oudeyer et al., 2016b, 2017b), the modeling of curiosity and intrinsic motivation (Oudeyer et al., 2016a), and scaling of learning theories (Oudeyer et al., 2017a).

Conventional approaches for testing system behavior were not designed for systems that learn. They assume that systems do not change after they have been tested and that the possible conditions leading to failure are known ahead of time. In contrast, the behavior of learning systems depends on things learned early, things learned later, and on surprises in new situations. Competent learning systems need to bring multiple sources of information to bear in deciding what to do in novel situations and how to generalize from their experiences. Potentially they can choose new experiences as experiments in order to understand how best to generalize.

The proposed research would experiment with computational techniques and combine testing and evaluation practices drawing on multi-disciplinary collaboration of neurosciences, cognitive psychology, and education. For example, neurosciences produce "brain maps" of the functions and activities of a brain and developmental psychologists produce focused experiments to repeat particular behaviors. One idea is to use maps of AI representations analogous to brain maps from neuroscience and violation of expectation experiments from psychology. We anticipate various approaches to building computational models to help understand what is learned. Restated, we suggest using psychologically-motivated scenarios with behavioral and computational "probes" to test, map, and analyze behavior changes and competence boundaries in the neural networks and other parts of embodied AIs. It could include visual analytics to produce maps of neural networks interactive data models to explore and make sense of experimental findings. This research would extend the methods and goals of DARPA's earlier explainable AI (XAI) program.

- *Interactive learning environments for embodied AIs.*
    o What learning environments would be effective for developmental AIs to learn the skills that human children learn? What settings can be effective for enabling the physical experimentation and social interactions like those that children have? How can the interactions with humans be safe, effective for AI learning, and teach appropriate human values? How can the substantial amount of required human interaction for "parenting" and "teaching" of AIs be arranged?

    Setting up learning environments where AIs can safely learn and explore the world is expensive, analogous to the costs of parents raising children. Providing multiple robots for experiments in real world settings would be a significant expense.

Developmental AI aims to advance the multi-disciplinary methodologies for bootstrapping and potentially offering a new kind of foundation model – *experiential foundation models.* These models would be created as developmental AIs interact with the world. Adapting them later for particular uses would be akin to training people to work in new areas.

The aspirational goal is to create AIs that could be able to operate in more situations. In the context of robots, developmental AI would potentially open up many currently unavailable applications in businesses and homes. Since training robots with the gap competences would be similar to training for people, developmental AIs would be practiced in interacting with people in task-oriented situations.

The bootstrapping approach for developmental AI is aimed at creating the kinds of robotic AI partners that have long populated our human imaginations and stories. We hypothesize that it may be a more direct and robust research route to human-compatible AIs than mainstream approaches.

We offer our gratitude to our colleagues for decades of research in the virtuous research cycles. We hope that the development of such AIs will be of widespread benefit.

*Acknowledgments*. We thank Robert Hoffman and William Clancey for their comments and suggestions on early drafts of this position paper and an earlier companion paper. Thank you to colleagues who engaged with us and provided perspectives and comments including Steve Cousins, Peter Denning, Edward Feigenbaum, Mike Frank, Celeste Kidd, Ben Kuipers, Ray Levitt, Charles Ortiz, Shantanu Rane, and Ben Shneiderman.

Thank you to our colleagues at SRI International including PARC. Thank you to Kai Goebel, director of the Intelligent Systems Laboratory, for support as we developed earlier versions of these ideas over several months.

58. Cangelosi, A., Asada, M. (eds.) (2022) *Cognitive Robotics*. The MIT Press. Cambridge, MA. https://direct.mit.edu/books/oa-edited-volume/5331/Cognitive-Robotics
59. Carey, S. (2004) Bootstrapping and the Origin of Concepts. *Daedalus* 133 (1), pp. 59-68. https://dash.harvard.edu/bitstream/handle/1/5109360/Carey_Bootstrapping.pdf?sequence=2&isAllowed=y
60. Carey, S. (2009) *The Origin of Concepts*. Oxford University Press. New York, New York.
61. Carlson, A., Betteridge, J., Kisiel, B., Settles, B., Hruschka Jr., E.R., Mitchell, T.M. (2010) Toward an Architecture for Never-Ending Language Learning. *Proceedings of the Conference on Artificial Intelligence (AAAI)*.
62. Champion, A. (2022) The Amazon Effect and Its Impact on E-commerce. *Flowspace Blog*. https://flow.space/blog/amazon-effect/
63. Chen, B., Zhu, C., Agrawal, P., Zhang. K., Gupta, A. (2023) Self-Supervised Reinforcement Learning that Transfers using Random Features. *arXiv* https://arxiv.org/pdf/2305.17250.pdf
64. Chomsky, N., Roberts, I., Watumull, J. (2023) The False Promise of ChatGPT. *New York Times* (Guest Essay), March 8, 2023. https://www.nytimes.com/2023/03/08/opinion/noam-chomsky-chatgpt-ai.html
65. Christiansen, M.H., Kirby S. (2003) *Language Evolution*. Oxford University Press. Available online at https://academic.oup.com/book/12818/chapter/163042125
66. Chung, S-J., Paranjape, A., Dames, P., Shen, S., Kumar, V. (2018) A Survey on Aerial Swarm Robotics. *IEEE Transactions on Robotics* 34( 4) pp 837-855.
67. Clark, H., Brennan, S. E. (1991) *Grounding in Communication*. In Resnick, L.B., Levine, J.M., Teasley, SS. D. (eds.) Perspectives on Socially Shared Cognition, American Psychological Association. pp. 127-149. http://www.psychology.sunysb.edu/sbrennan-/papers/old_clarkbrennan.pdf
68. Collobert, R., Weston, J., Botton, L., Karlen, M., Kavukcuoglu, K., Kuksa, P. (2011) Natural Language Processing (Almost) from Scratch. *Journal of Machine Learning Research* 12 pp. 2493-2537.
69. Csikszentmihalyi, M. (1996) Creativity – and the Psychology of Discovery and Invention. New York Harper Perennial.
70. Csikszentmihalyi, M. (2008, 1991) *Flow: The Psychology of Optimal Experience*. Harper Perennial, New York.
71. Cummings, M. L., (2023) What Self-Driving Cars Tell U About AI Risks. *IEEE Spectrum* 60(10) pp. 30-36. https://www.londonreconnections.com/2023/what-self-driving-cars-tell-us-about-ai-risks-ieeespectrum/
72. DARPA (2004) Grand Challenge 2004 https://www.esd.whs.mil/Portals/54/Documents/FOID/Reading%20Room/DARPA/15-F-0059_GC_2004_FINAL_RPT_7-30-2004.pdf
73. Defense Advanced Research Projects Agency (2018). *Broad Agency Announcement: Machine Common Sense*. HR001119S0005. https://research-authority.tau.ac.il/sites/resauth.tau.ac.il/files/DARPA_Machine_Common_Sense_MCS-HR001119S0005.pdf
74. Darwin, C.R. (1859) *The Origin of Species*.
75. Demiris, Y., Johnson, M. (2003) Distributed, prediction perception of actions: a biologically inspired architecture for imitation and learning. *Connection Science* 15(4) pp. 231–243.
76. Demiris, Y., Dearden, A. (2005) From motor babbling to hierarchical learning by imitation: a robot developmental pathway. Proceedings of the Fifth International

Workshop on Epigenetic Robotics: Modeling. Cognitive Development in Robotic Systems. Lund University Cognitive Studies.123 pp. 31-37.
77. Demiris, Y., Meltzoff, A. (2008) The Robot in the Crib: A Developmental Analysis of Imitation Skills in Infants and Robots. *Infant and Child Development* 17 pp. 43-53. https://onlinelibrary.wiley.com/doi/epdf/10.1002/icd.543
78. Deng, J., Dong, W., Socher, R., Li, L.-J., Li, K., Li, F.-F. (2009) ImageNet: a Large-Scale Hierarchical Image Database. *IEEE Computer Society Conference on Computer Vision and Pattern Recognition* (CVPR). https://www.researchgate.net/publication/221361415_ImageNet_a_Large-Scale_Hierarchical_Image_Database
79. Devlin, J., Chang, M., Lee, K., Toutanova, K. (2019) BERT: Pre-training of Deep Bidirectional Transformers for Language Understanding. *Proceedings of the 2019 Conference of the North American Chapter of the Association for Computational Linguistics: Human Language Technologies, Volume 1 (Long and Short Papers)*, pages 4171–4186, Minneapolis, Minnesota. Association for Computational Linguistics.
80. Dietterich, T. (2000) The MAXQ Method for Hierarchical Reinforcement Learning. *Journal of Artificial Intelligence Research* 13. pp. 227-303.
81. Dominey, P. F., Yoshida, E. (2009) Real-Time Spoken Language Programming for Cooperative Interaction with a Humanoid Apprentice, *International Journal of Humanoid Robotics* pp. 147-171.
82. Driess, D., Xia, F., Sajjadi, M. S. M., Lynch, C., Chowdhary, A., Ichter, B., Wahid, A., Tompson, J., Vuong, Q., Yu, T., Huang, W., Chebotar, Y., Sermanet, P., Duckworth, D., Levine, S., Vanhoucke, V., Hausman, K., Toussaint, M., Greff, K., Zeng, A., Mordatch, I., Florence, P. (2023) PaLM-E: An Embodied Multimodal Language Model. *arXiv* https://arxiv.org/abs/2303.03378
83. Duan, J., Yu, S., Tan, L, Zhu, H. (2022) A Survey of Embodied AI: From Simulators to Research Tasks. *IEEE Transactions on Emerging Topics in Computational Intelligence*. 6 pp. 230-244.
84. Dubova, M. (2022) Building Human-like Communicative Intelligence: A Grounded Perspective. *Cognitive Systems Research* 72 pp. 63-79. https://www.sciencedirect.com/science/article/abs/pii/S1389041721000966
85. Dutta, S., Chakraborty, T. (2024) Thus Spake ChatGPT *Communications of the ACM*. 68(12) pp. 16-19. https://dl.acm.org/doi/pdf/10.1145/3616863
86. Eckstein, M.K., Guerra-Carrillo, B., Singley A.T.M., Bunge, S.A. (2017) Beyond eye gaze: What else can eyetracking reveal about cognition and cognitive development? *Developmental Cognitive Neuroscience* 25 pp. 69-91.
87. Eimas, P.D., Siqueland, E. R., Jusczyk, P., Virorito, J. (1971) *Science*. 171 (3968), 303-306.
88. Ericsson, K.A., & Simon, H.A. (1993). *Protocol analysis: Verbal reports as Data* (Rev. ed.). Cambridge, MA: MIT Press.
89. Erman, L. D., Hayes-Roth, F., Lesser, V. R., Reddy, D. R. (1980) The Hearsay-II Speech-Understanding System: Integrating Knowledge to Resolve Uncertainty. *Computing Surveys* 12(2) pp. 213-253.
90. Ernst, M.O., Bülthoff, H. H. (2004) Merging the senses into a robust percept. *Science Direct: Trends in Cognitive Sciences*. Vol. 8, Issue 4.
91. Eskenazi, J. (2023) 'No! You Stay!' Cops, firefighters bewildered as driverless cars behave badly. *Mission Local*. (news story) https://missionlocal.org/2023/05/waymo-cruise-fire-department-police-san-francisco/

256. Ravichandar, H., Polydoros, A., Chernova, S., Billard, A. (2020) Recent Advances in Robot Learning from Demonstration, *Annual Review of Control, Robotics, and Autonomous Systems* 3. pp. 297-330. https://www.annualreviews.org/doi/10.1146/annurev-control-100819-063206
257. Redmon, J., Divvala. S., Girshick, R., Farhadi. A. (2016) You Only Look Once: Unified, Real-Time Object Detection. *IEEE Conference on Computer Vision and Pattern Recognition*, https://arXiv.org/pdf/1506.02640.pdf
258. Regier, T. (1991) Learning Perceptually-Grounded Semantics in the $L_0$ Project. 29th Annual Meeting of the Association for Computational Linguistics, pp. 138-145. https://aclanthology.org/P91-1018.pdf
259. Regier, T. (2003) Emergent constraints on word-learning: a computational perspective. *TRENDS in Cognitive Sciences* 17(6). pp. 263-268. https://lclab.berkeley.edu/papers/regier-tics-2003.pdf
260. Root, J. (2020) *Babies* (documentary video series), Netflix, https://www.netflix.com/title/80117833
261. Roy, D. K., Pentland, A. P. (2002) Learning words from sights and sounds: a computational model. *Cognitive Science* 26 pp. 113-146.
262. Roy, D. K. (2002) Learning visually grounded words and syntax for a scene description task. *Computer Speech and Language* 16, pp. 353-385.
263. Rizzolatti, G., Fadiga, L., Gallese, V., Fogassi, L. (1996) Premotor cortex and the recognition of motor actions. *Cognitive Brain Research* 3(2) pp. 131-141.
264. Rizzolatti, G., Arbib, M. A. (1998) Language within our grasp. Viewpoint. *Trends in Neurosciences* 21 pp. 188-194.
265. Rizzolatti, G., Fogassi, L. (2014) The mirror mechanism: recent findings and perspectives. *Philosophical Transactions of the Royal Society B*. 369. https://pubmed.ncbi.nlm.nih.gov/24778385/
266. Rumelhart, D. E.; Hinton, G. E.; Williams, J. (1986). "Learning representations by back-propagating errors". *Nature*. **323** (6088) pp. 533–536.
267. Russell, S. (2019) *Human Compatible: Artificial Intelligence and the Problem of Control.* Penguin Books, Random house. New York, New York. Viking, imprint of Penguin Random House, New York, New York.
268. Russell, S., Norvig, P. (2021) *Artificial Intelligence: A Modern Approach* (4th edition). (Kindle edition) Pearson, Harlow, United Kingdom.
269. Saby, J.N., Meltzoff, A.N., Marshall, P.J. (2013) Infants' Somatotopic Neural Responses to Seeing Human Actions: I've Got You under My Skin. *PLOS ONE*. 8 (10) https://doi.org/10.1371/journal.pone.0077905
270. Sacerdoti, E. (1974) Planning in a Hierarchy of Abstraction Spaces, *Artificial Intelligence* 5(2), pp. 115-135.
271. Saegusa, R., Metta, G., Sandini, G., Sakka, S. (2008) Active Motor Babbling for Sensory-Motor Learning. *2008 IEEE International Conference on Robotics and Biomimetics*. Pp. 794-799.
272. Sandini, G., Metta, G., Konczak, J. (1997) Human Sensori-Motor Development and Artificial Systems. *International Symposium on Artificial Intelligence and intellectual Human Activity Support for Applications*. Wakoshi, Japan. https://www.academia.edu/16672101/Human_SensoriMotor_Development_and_Artificial_Systems
107